\RequirePackage{fix-cm}
\documentclass[twocolumn]{svjour3}          % twocolumn
\smartqed  % flush right qed marks, e.g. at end of proof
\usepackage{graphicx}
\usepackage{amssymb}
\usepackage[english]{babel}
\usepackage[latin1]{inputenc}
\usepackage[T1]{fontenc}
\usepackage{amsmath}
\usepackage{amssymb}
\usepackage{program}

\usepackage{xcolor}
\usepackage{bm}
\usepackage{import}
\usepackage{array,booktabs,calc}
\usepackage{soul}

\usepackage[pagebackref=true,breaklinks=false,letterpaper=true,colorlinks,bookmarks=false]{hyperref}

\newcommand{\ie}{\textit{i}.\textit{e}., }
\newcommand{\eg}{\textit{e}.\textit{g}., }
\journalname{JMIV}
\begin{document}

\title{Variational Reflectance Estimation from Multi-view Images}


\author{Jean \textsc{M\'elou}$^{1,2}$ \and Yvain \textsc{Qu\'eau}$^3$ \and Jean-Denis \textsc{Durou}$^1$ \and \\ Fabien \textsc{Castan}$^2$ \and Daniel \textsc{Cremers}$^3$}

\authorrunning{Jean M\'elou et al.}

\institute{Jean \textsc{M\'elou} \at \email{jeme@mikrosimage.eu}
	\and
Yvain \textsc{Qu{\'e}au} \at \email{yvain.queau@tum.de} 
  \and
Jean-Denis \textsc{Durou} \at \email{durou@irit.fr}  
  \and	
Fabien \textsc{Castan} \at   \email{faca@mikrosimage.eu} 
  \and
Daniel \textsc{Cremers} \at 	\email{cremers@tum.de}
  \newline{}
  \newline{}  
  $^{1}$IRIT, UMR CNRS 5505, Universit\'e de Toulouse, Toulouse, France \at
  $^{2}$Mikros Image, Levallois-Perret, France  \at
  $^{3}$Department of Computer Science, Technical University of Munich, Garching, Germany \at
}
\date{Received: date / Accepted: date}

\maketitle

\begin{abstract}
 We tackle the problem of reflectance estimation from a set of multi-view images, assuming known geometry. The approach we put forward turns the input images into reflectance maps, through a robust variational method. The variational model comprises an image-driven fidelity term and a term which enforces consistency of the reflectance estimates with respect to each view. If illumination is fixed across the views, then reflectance estimation remains under-constrained: a regularization term, which ensures piece\-wi\-se-smooth\-ness of the reflectance, is thus used. Reflectance is parameterized in the image domain, rather than on the surface, which makes the numerical solution much easier, by resorting to an alternating majorization-mini\-mi\-zation approach. Experiments on both synthetic and real-world datasets are carried out to validate the proposed strategy. 
\end{abstract}

\section{Introduction}

Acquiring the shape and the reflectance of a scene is a key issue, \eg for the movie industry, as it allows proper relighting. The current proposed solutions focus on small objects and stand on multiple priors \cite{UnityDelight} or need very controlled environments \cite[Chapter 9]{ReinhardWardPattanaikDebevec}. Well-established shape acquisition techniques such as multi-view stereo exist for accurate 3D-recons\-truction. Nevertheless, they do not aim at recovering the surface reflectance. Hence, the original input images are usually mapped onto the 3D-reconstruction as texture. Since the image graylevel mixes shading information (induced by lighting and geometry) and reflectance (which is characteristic of the surface), relighting based on this approach usually lacks realism. To improve the results, reflectance needs to be separated from shading.

In order to more precisely illustrate our purpose, let us take the example of a Lambertian surface. In a 2D-point (pixel) $\mathbf{p}$ conjugate to a 3D-point $\mathbf{x}$ of a Lambertian surface, the graylevel $I(\mathbf{p})$ is written
\begin{equation}
	I(\mathbf{p}) = \rho(\mathbf{x}) \, \mathbf{s}(\mathbf{x}) \cdot \mathbf{n}(\mathbf{x}).
\label{eq:1}
\end{equation}
In the right-hand side of \eqref{eq:1}, $\rho(\mathbf{x}) \in \mathbb{R}$ is the albedo\footnote{Since the albedo suffices to characterize the reflectance of a Lambertian surface, we will name it ``reflectance'' as well.}, $\mathbf{s}(\mathbf{x}) \in \mathbb{R}^3$ the lighting vector, and $\mathbf{n}(\mathbf{x}) \in \mathbb{S}^2 \subset \mathbb{R}^3$ the outer unit-length normal to the surface. All these elements a priori depend on $\mathbf{x}$ \ie they are defined locally. Whereas $I(\mathbf{x})$ is always supposed to be given, different situations can occur, according to which are also known, among $\rho(\mathbf{x})$, $\mathbf{s}(\mathbf{x})$ and $\mathbf{n}(\mathbf{x})$.

\begin{figure*}[htbp]
\begin{center}
	\begin{tabular}{cccc}
		\includegraphics[height = .19\linewidth]{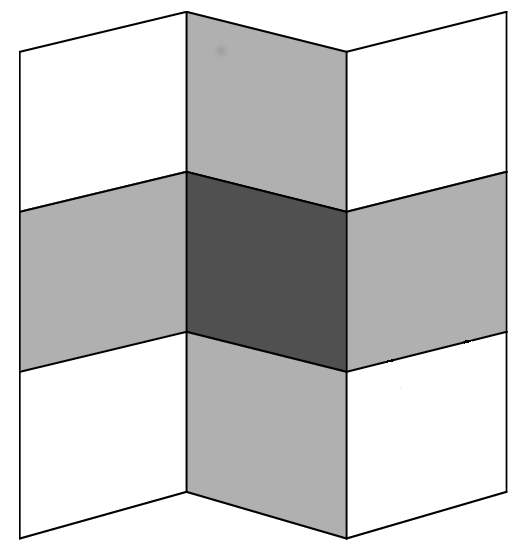} & \quad
		\includegraphics[height = .22\linewidth]{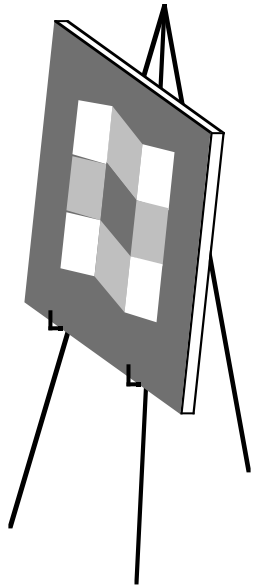} & \quad
		\includegraphics[height = .16\linewidth]{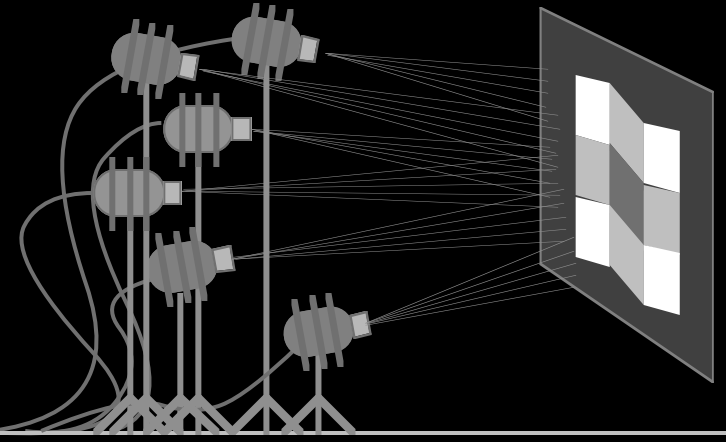} & \quad
		\includegraphics[height = .2\linewidth]{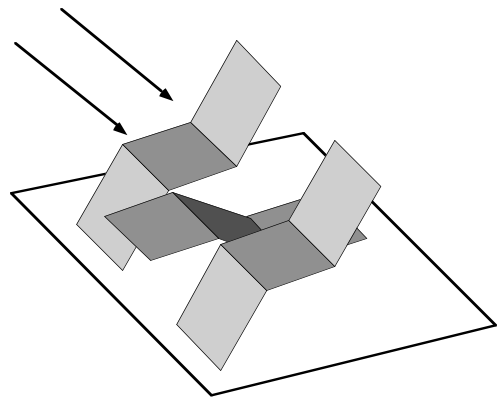} \\
		\small{(a)} &\quad \small{(b)} &\quad \small{(c)} &\quad \small{(d)}
	\end{tabular}
\end{center}
\caption{The ``workshop metaphor'' (extracted from a paper by Adelson and Pentland \cite{Adelson}). Image (a) may be interpreted either by: (b) incorporating all the brightness variations inside the reflectance; (c) modulating the lighting of a white planar surface; (d) designing a uniformly white 3D-shape illuminated by a parallel and uniform light beam. This last interpretation is one of the solutions of the shape-from-shading problem.}
\label{fig:Adelson}
\end{figure*}

One equation \eqref{eq:1} per pixel is not enough to simultaneously estimating the reflectance $\rho(\mathbf{x})$, the lighting $\mathbf{s}(\mathbf{x})$ and the geometry, represented here by $\mathbf{n}(\mathbf{x})$, because there are much more unknowns than equations. Figure \ref{fig:Adelson} illustrates this source of ill-posedness through the so-called ``workshop metaphor'' introduced by Adelson and Pentland in \cite{Adelson}: among three plausible interpretations (b), (c) and (d) of image (a), we are particularly interested in (d), which illustrates the principle of \emph{photometric} 3D-reconstruction. This class of methods usually assume that the lighting $\mathbf{s}(\mathbf{x})$ is known. Still, there remains three scalar unknowns per equation \eqref{eq:1}: $\rho(\mathbf{x})$ and $\mathbf{n}(\mathbf{x})$, which has two degrees of freedom. Assuming moreover that the reflectance $\rho(\mathbf{x})$ is known, the \emph{shape-from-shading} technique \cite{Horn} uses the shading $\mathbf{s}(\mathbf{x}) \cdot \mathbf{n}(\mathbf{x})$ as unique clue to recover the shape $\mathbf{n}(\mathbf{x})$ from Equation~\eqref{eq:1}, but the problem is still ill-posed.

A classical way to make photometric 3D-reconstruc\-tion well-posed is to use $m>1$ images taken using a single camera pose, but under varying known lighting:
\begin{equation}
	I^i(\mathbf{p}) = \rho(\mathbf{x}) \, \mathbf{s}^i(\mathbf{x}) \cdot \mathbf{n}(\mathbf{x}), \quad i \in \{1,\dots,m\}
\label{eq:2}
\end{equation}
In this variant of shape-from-shading called \emph{photometric stereo} \cite{Woodham1980a}, the reflectance $\rho(\mathbf{x})$ and the normal $\mathbf{n}(\mathbf{x})$ can be estimated without any ambiguity, as soon as $m\geq 3$ non-coplanar lighting vectors $\mathbf{s}^i(\mathbf{x})$ are used.

Symmetrically to \eqref{eq:2}, solving the problem:
\begin{equation}
	I^i(\mathbf{p}) = \rho(\mathbf{x}) \, \mathbf{s}(\mathbf{x}) \cdot \mathbf{n}^i(\mathbf{x}), \quad i \in \{1,\dots,m\}
\label{eq:3}
\end{equation}
allows to estimate the lighting $\mathbf{s}(\mathbf{x})$, as soon as the reflectance $\rho(\mathbf{x})$ and $m \geq 3$ non-coplanar normals $\mathbf{n}^i(\mathbf{x})$, $i \in \{1,\dots,m\}$, are known. This can be carried out, for instance, by placing a small calibration pattern with known color and known shape near each 3D-point $\mathbf{x}$~\cite{Queau2017bis}.

\begin{figure*}[!ht]
\centering
	\includegraphics[width = .98\linewidth]{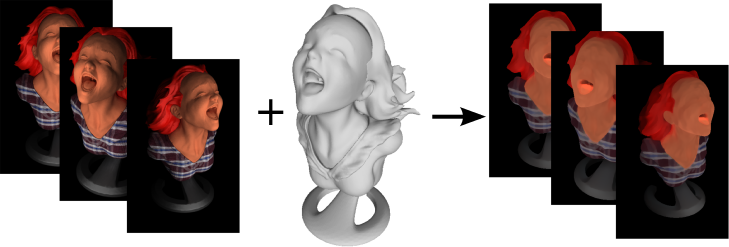}
\caption{Overview of our contribution. From a set of $n$ images of a surface acquired under different angles, and a coarse geometry obtained for instance using multi-view stereo, we estimate a shading-free reflectance map per view.}
\label{fig:catchy}
\end{figure*}

The problem we aim at solving in this paper is slightly different. Suppose we are given a series of $m>1$ images of a scene taken using a single lighting, but $m$ camera poses. According to Lambert's law, this ensures that a 3D-point looks equally bright in all the images where it is visible. Such invariance is the basic clue of multi-view stereo (MVS), which has become a very popular technique for 3D-reconstruction \cite{MVS}. Therefore, since an estimate of the surface shape is available, $\mathbf{n}(\mathbf{x})$ is known. Now, we have to index the pixels by the image number $i$. Fortunately, additional data provided by MVS are the correspondences between the different views, taking the form of $m$-tuples of pixels $(\mathbf{p}^i)_{i \in \{1,\dots,m\}}$ which are conjugate to a common 3D-point $\mathbf{x}$.

Our problem is written\footnote{Even if they look very similar, Problems \eqref{eq:2}, \eqref{eq:3} and \eqref{eq:4} have completely different peculiarities.}:
\begin{equation}
	I^i(\mathbf{p}^i) = \rho(\mathbf{x}) \, \mathbf{s}(\mathbf{x}) \cdot \mathbf{n}(\mathbf{x}), \quad i \in \{1,\dots,m\}
\label{eq:4}
\end{equation}
where $\mathbf{p}^i$ is the projection of $\mathbf{x}$ in the $i$-th image, and $\rho(\mathbf{x})$ and $\mathbf{s}(\mathbf{x})$ are unknown. Obviously, this system reduces to Equation \eqref{eq:1}, since its $m$ equations are the same one: the right-hand side of \eqref{eq:4} does not depend on $i$, not more than the left-hand side $I^i(\mathbf{p}^i)$ since, as already noticed, the lighting $\mathbf{s}(\mathbf{x})$ does not vary from one image to another, and the surface is Lambertian.

Multi-view helps estimating the reflectance, because it provides the 3D-shape via MVS. However, even if $\mathbf{n}(\mathbf{x})$ is known, Equation \eqref{eq:1} remains ill-posed. This is illustrated, in Figure \ref{fig:Adelson}, by the solutions (b) and (c), which correspond to the same image (a) and to a common planar surface. In the absence of any prior, Equation \eqref{eq:1} has an infinity of solutions in $\rho(\mathbf{x}) \, \mathbf{s}(\mathbf{x})$. In addition, determining $\rho(\mathbf{x})$ from each of these solutions would give rise to another ambiguity, since $\mathbf{s}(\mathbf{x})$ is not forced to be unit-length, contrarily to $\mathbf{n}(\mathbf{x})$.

Such a double source of ill-posedness probably explains why various methods for reflectance estimation have been designed, introducing a variety of priors in order to disambiguate the problem. Most of them assume that brightness variations induced by reflectance changes are likely to be strong but sparsely distributed, while the lighting is likely to induce smoother changes~\cite{Land1971}.

This suggests to separate a single image into a piecewise smooth layer and a more oscillating one. In the computer vision literature, this is often referred to as ``intrinsic image decomposition'', while the terminology ``cartoon + texture decomposition'' is more frequently used by the mathematical imaging community (both these problems will be discussed in Section \ref{sec:relatedwork}).

\paragraph{Contributions.}
In this work, we show the relevance of using multi-view images for reflectance estimation. Indeed, this enables a prior shape estimation using MVS, which essentially reduces the decomposition problem to the joint estimation of a set of reflectance maps, as illustrated in Figure \ref{fig:catchy}. We elaborate on the variational approach to multi-view decomposition into reflectance and shading, which we initially presented in \cite{SSVM_2017_Jean}. The latter introduced a robust $l^1$-TV framework for the joint estimation of piecewise-smooth reflectance maps and of spherical harmonics lighting, with an additional term ensuring the consistency of the reflectance maps. The present paper extends this approach by developing the theoretical foundations of this variational model. In this view, our parameterization choices are further discussed and the underlying ambiguities are exhibited. The variational model is motivated by a Bayesian rationale, and the proposed numerical scheme is interpreted in terms of a majorization-minimization algorithm. Finally, we conclude that, besides a preliminary measurement of the incoming lighting, varying the lighting along with the viewing angle, in the spirit of photometric stereo, is the only way to estimate the reflectance without resorting to any prior.

\paragraph{Organization of the Paper.}
After reviewing related approaches in Section \ref{sec:relatedwork}, we formalize in Section \ref{sec:model} the problem of multi-view reflectance estimation. Section~\ref{sec:var} then introduces a Bayesian-to-variational approach to this problem. A simple numerical strategy for solving the resulting variational problem, which is based on alternating majorization-minimization, is presented in Section \ref{sec:numerics}. Experiments on both synthetic and real-world datasets are then conducted in Section \ref{sec:results}, before summarizing our achievements and suggesting future research directions in Section \ref{sec:conclusion}.

\section{Related Works}
\label{sec:relatedwork}

Studied since the 1970s \cite{Land1971}, the problem of decomposing an image (or a set of images) into a piecewise-smooth component and an oscillatory one is a fundamental computer vision problem, which has been addressed in numerous ways.

\paragraph{Cartoon + Texture Decomposition.}
Researchers in the field of mathematical imaging have suggested various variational models for this task, using for instance non-smooth regularization and Fourier-based frequency analysis \cite{Aujol2006}, or $l^1$-TV variational models \cite{IpolCartoon}. However, such techniques do not use an explicit photometric model for justifying the decomposition, whereas photometric analysis, which is another important branch of computer vision, may be a source of inspiration for motivating new variational models.

\paragraph{Photometric Stereo.}
As discussed in the introduction, photometric stereo techniques \cite{Woodham1980a} are able to unambiguously estimate the reflectance and the geometry, by considering several images obtained from the same viewing angle but under calibrated, varying lighting. Photometric stereo has even been extended to the case of uncalibrated, varying lighting \cite{Basri2007}. In the same spirit as uncalibrated photometric stereo, our goal is to estimate reflectance under unknown lighting. However, the problem is less constrained in our case, since we cannot ensure that the lighting is varying. Our hope is that this can be somewhat compensated by the prior knowledge of geometry, and by the resort to appropriate priors. Various priors for reflectance have been discussed in the context of intrinsic image decomposition.

\paragraph{Intrinsic Image Decomposition.}
Separating reflectance from shading in a single image is a challenging problem, often referred to as intrinsic image decomposition. Given the ill-posed nature of this problem, prior information on shape, reflectance and/or lighting must be introduced. Most of the existing works are based on the ``retinex theory'' \cite{Land1971}, which states that most of the slight brightness variations in an image are due to lighting, while reflectance is piecewise-constant (as for instance a Mondrian image). A variety of clustering-based \cite{GMLG12,Shen2011} or sparsity-enhancing methods \cite{Gehler2011,Nadian-Ghomsheh2016,Shen2011,Song2017} have been developed based on this theory. Among others, the work of Baron and Malik \cite{Barron}, which presents interesting results, stands on multiple priors to solve the fundamental ambiguity of shape-from-shading, that we aim at revoking in the multi-view context. Some other methods disambiguate the problem by requiring the user to ``brush'' uniform reflectance parts \cite{BPD09,Nadian-Ghomsheh2016}, or by resorting to a crowdsourced database \cite{bell14intrinsic}. Still, these works require user interactions, which may not be desirable in certain cases.

\paragraph{Multi-view 3D-reconstruction.}
Instead of introducing possibly unverifiable priors, or relying on user interactions, ambiguities can be reduced by assuming that the geometry of the scene is known. Intrinsic image decomposition has for instance been addressed using an RGB-D camera \cite{Chen2013} or, closer to our proposal, multiple views of the same scene under different angles \cite{Laffont2013,Laffont2012}. In the latter works, the geometry is first extracted from the multi-view images, before the problem of reflectance estimation is addressed. Geometry computation can be achieved using multi-view stereo (MVS). MVS techniques \cite{Seitz} have seen significant growth over the last decade, an expansion which goes hand in hand with the development of structure-from-motion (SfM) solutions \cite{Moulon}. Indeed, MVS requires the parameters of the cameras, outputs of the SfM algorithm. Nowadays, these mature methods are commonly used in uncontrolled environments, or even with large-scale Internet data \cite{RomeInADay}. For the sake of completeness, let us also mention that some efforts in the direction of multi-view and photometrically consistent 3D-reconstruction have been devoted recently \cite{Jin-et-al-08,Kim,Langguth,Robert,Maurer}. Similar to these methods, we will resort to a compact representation of lighting, namely the spherical harmonics model.

\paragraph{Spherical Harmonics Lighting Model.}
Let us consider a point $\mathbf{x}$ lying on the surface $\mathcal{S} \subset \mathbb{R}^3$ of the observed scene, and let $\mathbf{n}(\mathbf{x})$ be the outer unit-length normal vector to $\mathcal{S}$ in $\mathbf{x}$. Let $\mathcal{H}(\mathbf{x})$ be the hemisphere centered in $\mathbf{x}$, having as basis plane the tangent plane to $\mathcal{S}$ in $\mathbf{x}$. Each light source visible from $\mathbf{x}$ can be associated to a point $\omega$ on $\mathcal{H}(\mathbf{x})$. If we describe by the vector $\mathbf{s}(\mathbf{x},\omega)$ the corresponding elementary light beam (oriented towards the source), then by definition of the \emph{reflectance} (or BRDF) of the surface, denoted $r$, the \emph{luminance} of $\mathbf{x}$ in the direction $\mathbf{v}$ is given by
\begin{equation}
	L(\mathbf{x},\mathbf{v}) = \!\!\displaystyle\int_{\mathcal{H}(\mathbf{x})} \!\!\!\!\!\!\!\!\!r(\mathbf{x},\mathbf{n}(\mathbf{x}),\frac{{\mathbf{s}}(\mathbf{x},\omega)}{\|{\mathbf{s}}(\mathbf{x},\omega)\|},\mathbf{v}) \,[\mathbf{s}(\mathbf{x},\omega) \cdot \mathbf{n}(\mathbf{x})] \,\mathrm{d}\omega,
\label{eq:luminance_integral}
\end{equation}
where $[\mathbf{s}(\mathbf{x},\omega) \cdot \mathbf{n}(\mathbf{x})]$ is the surface \emph{illuminance}. In general, $r$ depends both on the direction of the light ${\mathbf{s}}(\mathbf{x},\omega)$, and on the viewing direction $\mathbf{v}$, relatively to $\mathbf{n}(\mathbf{x})$.

This expression of the luminance is intractable in the general case. However, if we restrict our attention to Lambertian surfaces, the reflectance reduces to the albedo $\rho(\mathbf{x})$, which is independent of any direction, and $L(\mathbf{x},\mathbf{v})$ does not depend on the viewing direction $\mathbf{v}$ anymore. If the light sources are further assumed to be distant enough from the object, then $\mathbf{s}(\mathbf{x},\omega)$ is independent of $\mathbf{x}$ \ie the light beams are the same for the whole (supposedly convex) object, and thus the lighting is completely defined on the unit sphere. Therefore, the integral \eqref{eq:luminance_integral} acts as a convolution on $\mathcal{H}(\mathbf{x})$, having as kernel $\mathbf{s}(\omega) \cdot \mathbf{n}(\mathbf{x})$. Spherical harmonics, which can be considered as the analogue to the Fourier series on the unit sphere, have been shown to be an efficient low-dimensional representation of this convolution \cite{Basri,Ramamoorthi}. Many vision applications \cite{Kim,Wu} use second order spherical harmonics, which can capture over $99\%$ of the natural lighting \cite{Frolova} using only nine coefficients. This yields an approximation of the luminance of the form
\begin{equation}
	L = \frac{\rho}{\pi} \, {\bm \sigma} \cdot {\bm \nu},
\label{eq:luminance_1}
\end{equation}
where $\rho \in \mathbb{R}$ is the albedo (reflectance), ${\bm \sigma} \in \mathbb{R}^9$ is a compact lighting representation, and ${\bm \nu} \in \mathbb{R}^9$ stores the local geometric information. The latter is deduced from the normal according to:
\begin{equation}
	{\bm \nu} = \begin{bmatrix}
		\mathbf{n} \\ 1 \\ n_1\,n_2 \\ n_1\,n_3 \\ n_2\,n_3 \\ n_1^2-n_2^2 \\ 3n_3^2-1
	\end{bmatrix}.
\label{eq:m}
\end{equation}

In \eqref{eq:luminance_1}, the lighting vector ${\bm \sigma}$ is the same in all the points of the surface, but the reflectance $\rho$ and the geometric vector ${\bm \nu}$ vary along the surface $\mathcal{S}$ of the observed scene. Hence we will write \eqref{eq:luminance_1} as:
\begin{equation}
	L(\mathbf{x}) = \frac{\rho(\mathbf{x})}{\pi} \, {\bm \sigma} \cdot {\bm \nu}(\mathbf{x}),\quad \forall \mathbf{x} \in \mathcal{S}.
\label{eq:luminance}
\end{equation}
Our aim in this paper is to estimate the reflectance $\rho(\mathbf{x})$ in each point $\mathbf{x} \in \mathcal{S}$, as well as the lighting vector ${\bm \sigma}$, given a set of multi-view images and the geometric vector ${\bm \nu}(\mathbf{x})$. We formalize this problem in the next section.

\section{Multi-view Reflectance Estimation}
\label{sec:model}

In this section, we describe with more care the problem of reflectance estimation from a set of multi-view images. First, we need to explicit the relationship between graylevel, reflectance, lighting and geometry.

\subsection{Image Formation Model}

Let $\mathbf{x} \in \mathcal{S}$ be a point on the surface of the scene. Assume that it is observed by a graylevel camera with linear response function and let $I:\,\Omega \subset \mathbb{R}^2 \to \mathbb{R}$ be the image, where $\Omega$ is the projection of $\mathcal{S}$ onto the image plane. Then, the graylevel in the pixel $\mathbf{p} \in \Omega$ conjugate to $\mathbf{x}$ is proportional to the luminance of $\mathbf{x}$ in the direction of observation $\mathbf{v}$:
\begin{equation}
	I(\mathbf{p}) = \gamma \, L(\mathbf{v},\mathbf{x}),
\label{eq:IL}
\end{equation}
where the coefficient $\gamma > 0$, referred to in the following as the ``camera coefficient'', is unknown\footnote{This coefficient depends on several factors such as the lens aperture, the magnification, the exposure time, etc.}. By assuming Lambertian reflectance and the light sources distant enough from the object, Equations \eqref{eq:luminance} and \eqref{eq:IL} yield:
\begin{equation}
	I(\mathbf{p}) = \gamma \, \frac{\rho(\mathbf{x})}{\pi} \, {\bm \sigma} \cdot {\bm \nu}(\mathbf{x}).
\label{eq:I}
\end{equation}

Now, let us assume that $m$ images $I^i$ of the surface, $i \in \{1,\dots,m\}$, obtained while moving a single camera, are available, and discuss how to adapt \eqref{eq:I}.

\paragraph{Case 1: unknown, yet fixed lighting and camera coefficient.} If all the automatic settings of the camera are disabled, then the camera coefficient is independent from the view. We can thus incorporate this coefficient and the denominator $\pi$ into the lighting vector: ${\bm \sigma}:= \frac{\gamma}{\pi} \, {\bm \sigma}$. Moreover, if the illumination is fixed, the lighting vector~${\bm \sigma}$ is independent from the view. In any point $\mathbf{x}$ which is visible in the $i$-th view, Equation \eqref{eq:I} becomes:
\begin{equation}
	I^i(\pi^i(\mathbf{x})) = \rho(\mathbf{x}) \, {\bm \sigma} \cdot {\bm \nu}(\mathbf{x}),
\label{eq:Ii_fixed_lighting}
\end{equation}
where we denote by $\pi^i$ the 3D-to-2D projection associated to the $i$-th view. In \eqref{eq:Ii_fixed_lighting}, the unknowns are the reflectance $\rho(\mathbf{x})$ and the lighting vector ${\bm \sigma}$. Equations~\eqref{eq:Ii_fixed_lighting}, $i \in \{1,\dots,m\}$, constitute a generalization of \eqref{eq:4} to more complex illumination scenarios. For the whole scene, this is a problem with $n+9$ unknowns and up to $nm$ equations, where $n$ is the number of 3D-points $\mathbf{x}$ which have been estimated by multi-view stereo. However, as for System \eqref{eq:4}, only $n$ equations are linearly independent, hence the problem of reflectance and lighting estimation is under-constrained.

\paragraph{Case 2: unknown and varying lighting and camera coefficient.}
If lighting is varying, then we have to make the lighting vector view-dependent. If it is also assumed to vary, the camera coefficient can be integrated into the lighting vector with the denominator $\pi$ \ie ${\bm \sigma^i}:= \frac{\gamma^i}{\pi} \, {\bm \sigma^i}$, since the estimation of each ${\bm \sigma^i}$ will include that of $\gamma^i$. Equation \eqref{eq:I} then becomes:
\begin{equation}
	I^i(\pi^i(\mathbf{x})) = \rho(\mathbf{x}) \, {\bm \sigma^i} \cdot {\bm \nu}(\mathbf{x}).
\label{eq:Ii_varying_lighting}
\end{equation}
There are even more unknowns ($n+9m$), but this time the $nm$ equations are linearly independent, at least as long as the ${\bm \sigma^i}$ are not proportional \ie if not only the camera coefficient or the lighting intensity vary across the views, but also the lighting direction\footnote{Another case, which we do not study here, is when the lighting and camera coefficient are both varying, yet only lighting is calibrated. This is known as ``semi-calibrated'' photometric stereo \cite{Cho2016}.}. Typically, $n$ is of the order of $[10^3,10^6]$, hence the problem is over-constrained as soon as at least two out of the $m$ lighting vectors are non-collinear. This is a situation similar to uncalibrated photometric stereo \cite{Basri2007}, but much more favorable: the geometry is known, hence the ambiguities arising in uncalibrated photometric stereo are likely to be reduced. However, contrarily to uncalibrated photometric stereo, lighting is not actively controlled in our case. Lighting variations are likely to happen \eg in outdoor scenarios, yet they will be limited. The $m$ lighting vectors ${\bm \sigma}^i$, $i \in \{1,\ldots,m\}$, will thus be close to each other: lighting variations will not be sufficient in practice for disambiguation (ill-conditioning).

Since \eqref{eq:Ii_fixed_lighting} is under-constrained and \eqref{eq:Ii_varying_lighting} is ill-condi\-tioned, additional information will have to be introduced either ways, and we can restrict our attention to the varying lighting case \eqref{eq:Ii_varying_lighting}.

So far, we have assumed that graylevel images were available. To extend our study to RGB images, we abusively assume channel separation, and apply the framework independently in each channel $\star \in \{R,G,B\}$. We then consider the expression:
\begin{equation}
	I^i_\star(\pi^i(\mathbf{x})) = \rho_\star(\mathbf{x}) \, {\bm \sigma}_\star^i \cdot {\bm \nu}(\mathbf{x})
\label{eq:Ii_colored}
\end{equation}
where $\rho_\star(\mathbf{x})$ and ${\bm \sigma}_\star^i$ denote, respectively, the colored reflectance and the $i$-th colored lighting vector, relatively to the response of the camera in channel $\star$. A more complete study of Model \eqref{eq:Ii_colored} is presented in \cite{JMIV2017_LEDS}.

Since we will apply the same framework independently in each color channel, we consider hereafter the graylevel case only \ie we consider the image formation model \eqref{eq:Ii_varying_lighting} instead of \eqref{eq:Ii_colored}. The question which arises now is how to estimate the reflectance $\rho(\mathbf{x})$ from a set of equations such as \eqref{eq:Ii_varying_lighting}, when the geometry ${\bm \nu}(\mathbf{x})$ is known but the lighting ${\bm \sigma}^i$ is unknown.

\subsection{Reflectance Estimation on the Surface}

We place ourselves at the end of the multi-view 3D-reconstruction pipeline. Thus, the projections $\pi^i$ are known (in practice, they are estimated using SfM techniques), as well as the geometry, represented by a set of $n$ 3D-points $\mathbf{x}_j \in \mathbb{R}^3$, $j \in \{1,\dots,n\}$, and the corresponding normals $\mathbf{n}(\mathbf{x}_j)$ (obtained for instance using SFM techniques), from which the $n$ geometric vectors ${\bm \nu}_j:={\bm \nu}(\mathbf{x}_j)$ are easily deduced according to \eqref{eq:m}.

The unknowns are then the $n$ reflectance values $\rho_j := \rho(\mathbf{x}_j) \in \mathbb{R}$ and the $m$ lighting vectors ${\bm \sigma}^i \in \mathbb{R}^9$, which are independent from the 3D-point number $j$ due to the distant light assumption. At first glance, one may think that their estimation can be carried out by simultaneously solving \eqref{eq:Ii_varying_lighting} in all the 3D-points $\mathbf{x}_j$, in a purely data-driven manner, using some fitting function $F:\,\mathbb{R} \to \mathbb{R}$:
\begin{equation}
	\min_{\substack{\{\rho_j \in \mathbb{R}\}_j\\ \{ {\bm \sigma^i} \in \mathbb{R}^9\}_i}} \sum_{i=1}^m \sum_{j=1}^n v^i_j \, F\left( \rho_j \, {\bm \sigma}^i \cdot {\bm \nu}_j - I^i_j \right),
\label{eq:LS_naive}
\end{equation}
where we denote $I^i_j = I^i(\pi^i(\mathbf{x}_j))$, and $v^i_j$ is a visibility boolean such that $v^i_j = 1$ if $\mathbf{x}_j$ is visible in the $i$-th image, and $v^i_j = 0$ otherwise.

Let us consider, for the sake of pedagogy, the simplest case of least-squares fitting ($F(x) = x^2$) and perfect visibility ($v^i_j \equiv 1$). Then, Problem \eqref{eq:LS_naive} is rewritten in matrix form:
\begin{equation}
	\min_{\substack{ {\bm \rho} \in \mathbb{R}^n\\ \mathbf{S} \in \mathbb{R}^{9 \times m} }} \left\| \mathbf{N} \left( {\bm \rho} \otimes \mathbf{S}\right) - \mathbf{I} \right\|_F^2,
\label{eq:kron1}
\end{equation}
where the Kronecker product ${\bm \rho} \otimes \mathbf{S}$ is a matrix of $\mathbb{R}^{9n \times m}$, ${\bm \rho}$ being a vector of $\mathbb{R}^n$ which stores the $n$ unknown reflectance values, and $\mathbf{S}$ a matrix of $\mathbb{R}^{9 \times m}$ which stores the $m$ unknown lighting vectors ${\bm \sigma}^i \in \mathbb{R}^{9}$, column-wise, $\mathbf{N} \in \mathbb{R}^{n \times 9n}$ is a block-diagonal matrix whose $j$-th block, $j \in \{1,\dots,n\}$, is the row vector ${\bm \nu}_j^\top$, matrix $\mathbf{I} \in \mathbb{R}^{n \times m}$ stores the graylevels, and $\|\cdot\|_F$ is the Frobenius norm.

Using the pseudo-inverse $\mathbf{N}^\dagger$ of $\mathbf{N}$, \eqref{eq:kron1} is rewritten:
\begin{equation}
	\min_{\substack{ {\bm \rho} \in \mathbb{R}^n\\ \mathbf{S} \in \mathbb{R}^{9 \times m} }} \left\| {\bm \rho} \otimes \mathbf{S} - \mathbf{N}^\dagger \, \mathbf{I} \right\|_F^2.
\label{eq:kron2}
\end{equation}
Problem \eqref{eq:kron2} is a nearest Kronecker product problem, which can be solved by singular value decomposition (SVD) \cite[Theorem 12.3.1]{GolubV4}.

However, this matrix factorization approach suffers from three shortcomings:
\begin{itemize}
	\item[1)] It is valid only if all 3D-points are visible under all the viewing angles, which is rather unrealistic. In practice, \eqref{eq:kron1} should be replaced by
	\begin{equation}
		\min_{\substack{ {\bm \rho} \in \mathbb{R}^n\\ \mathbf{S} \in \mathbb{R}^{9 \times m} }} \left\| \mathbf{V}
			\circ \left[ \mathbf{N} \left( {\bm \rho} \otimes \mathbf{S}\right) - \mathbf{I} \right] \right\|_F^2,
	\label{eq:kron3}
	\end{equation}
	where $\mathbf{V} \in \{0,1\}^{n \times m}$ is a visibility matrix containing the values $v^i_j$, and $\circ$ is the Hadamard product. This yields a Kronecker product problem with missing data, which is much more arduous to solve.

	\item[2)] It is adapted only to least-squares estimation. Considering a more robust fitting function would prevent a direct SVD solution.

	\item[3)] If lighting is not varying (${\bm \sigma}^i = {\bm \sigma} , \forall i \in \{1,\dots,m\}$), then it can be verified that \eqref{eq:kron1} is ill-posed. Among its many solutions, the following trivial one can be exhibited:
	\begin{align}
		& \mathbf{S}_{\text{trivial}} = {\bm \sigma}_{\text{diffuse}} \, \mathbf{1}_{1 \times m}, \\
		& {\bm \rho}_{\text{trivial}} = \left[ E_i[I^i_1],\dots,E_i[I^i_n] \right]^\top,
	\end{align}
	where:
	\begin{align}
		& {\bm \sigma}_{\text{diffuse}} = \left[0,0,0,1,0,0,0,0,0\right]^\top
	\label{eq:sigma_naive}
	\end{align}
	and $E_i$ is the mean over the view indices $i$. This trivial solution means that the lighting is assumed to be completely diffuse\footnote{In the computer graphics community, this is referred to as ``ambient lighting''.}, and that the reflectance is equal to the image graylevel, up to noise only. Obviously, this is not an acceptable interpretation. As discussed in the previous subsection, in real-world scenarios we will be very close to this degenerate case, hence additional regularization will have to be introduced, which makes things even harder.
\end{itemize}

\noindent{Overall, the optimization problem which needs to be addressed is not as easy as \eqref{eq:kron2}. It is a non-quadratic regularized problem of the form:}
\begin{equation}
	\min_{\substack{\{\rho_j \in \mathbb{R}\}_j\\ \{ {\bm \sigma^i} \in \mathbb{R}^9\}_i }}
	\! \sum_{i=1}^p \sum_{j=1}^n \! v^i_j \, F\!\left( \rho_j \, {\bm \sigma}^i \!\cdot\! {\bm \nu}_j \!-\! I^i_j \right) + \sum_{j=1}^n \sum_{k \vert \mathbf{x}_k \in \mathcal{V}(\mathbf{x}_j)} \!\!\!\!\!\!\! R(\rho_j,\rho_k),
\label{eq:LS_reg}
\end{equation}
where $\mathcal{V}(\mathbf{x}_j)$ is the set of neigbors of $\mathbf{x}_j$ on surface $\mathcal{S}$, and the regularization function $R$ needs to be chosen appropriately to ensure piecewise-smoothness.

However, the sampling of the points $\mathbf{x}_j$ on surface~$\mathcal{S}$ is usually non-uniform, because the shape of $\mathcal{S}$ is potentially complex. It may thus be difficult to design appropriate fidelity and regularization functions $F$ and $R$, and to design an appropriate numerical solving. In addition, some thin brightness variations may be missed if the sampling is not dense enough. Overall, direct estimation of reflectance on the surface looks promising at first sight, but rather tricky in practice. Therefore, we leave this as an interesting future research direction and follow in this paper a simpler approach, which consists in estimating reflectance in the image domain.

\subsection{Reflectance Estimation in the Image Domain}

Instead of trying to colorize the $n$ 3D-points estimated by MVS \ie of parameterizing the reflectance over the (3D) surface $\mathcal{S}$, we can also formulate the reflectance estimation problem in the (2D) image domain.

Equation \eqref{eq:Ii_varying_lighting} is equivalently written, in each pixel $\mathbf{p} := \pi^i(\mathbf{x}) \in \Omega^i:= \pi^i(\mathcal{S})$:
\begin{equation}
	I^i(\mathbf{p}) = \rho^i(\mathbf{p}) \, {\bm \sigma}^i \cdot {\bm \nu}^i(\mathbf{p}),
\label{eq:Ii}
\end{equation}
where we denote $\rho^i(\mathbf{p}) := \rho({\pi^i}^{-1}(\mathbf{p}))$ and ${\bm \nu}^i(\mathbf{p}) := {\bm \nu}({\pi^i}^{-1}(\mathbf{p}))$. Instead of estimating one reflectance value $\rho(\mathbf{x})$ per estimated 3D-point, the reflectance estimation problem is thus turned into the estimation of $m$ ``reflectance maps''
\begin{equation}
	\rho^i:\,\Omega^i \subset \mathbb{R}^2 \to \mathbb{R}.
\label{eq:rhoi}
\end{equation}

On the one hand, the 2D-parameterization \eqref{eq:rhoi} does not enforce the consistency of the reflectance maps. This will have to be explicitly enforced later on. Besides, the surface will not be directly colorized, but the estimated reflectance maps have to be back-projected and fused over the surface in a final step.

On the other hand, the question of occlusions (visibility) does not arise, and the domains $\Omega^i$ are subsets of a uniform square 2D-grid. Therefore, it will be much easier to design appropriate fidelity and regularization terms. Besides, there will be as many reflectance estimates as pixels in those sets: with modern HD cameras, this number is much larger than the number of 3D-points estimated by multi-view stereo. Estimation will thus be much denser.

With such a parameterization choice, the regularized problem \eqref{eq:LS_reg} will be turned into:
\begin{align}
	& \min_{\substack{\{\rho^i:\,\Omega^i \to \mathbb{R}\}_i\\ \{ {\bm \sigma^i} \in \mathbb{R}^9\}_i }} 
	\! \sum_{i=1}^p \sum_{\mathbf{p} \in \Omega^i} F\!\left( \rho^i(\mathbf{p}) \, {\bm \sigma}^i \!\cdot\! {\bm \nu}^i(\mathbf{p}) \!-\! I^i(\mathbf{p}) \right) \nonumber \\
	& \qquad\qquad + \sum_{i=1}^p \sum_{\mathbf{p} \in \Omega^i} \sum_{ \mathbf{q} \in \mathcal{V}^i(\mathbf{p})} R(\rho^i(\mathbf{p}),\rho^i(\mathbf{q})) \nonumber \\
	& \qquad \text{s.t. } C(\{\rho^i\}_i) = 0,
\label{eq:LS_reg2}
\end{align}
with $C$ some function to ensure multi-view consistency, and where $\mathcal{V}^i(\mathbf{p})$ is the set of neighbors of pixel $\mathbf{p}$ which lie inside $\Omega^i$. Note that, since $\Omega^i$ is a subset of a square, regular 2D-grid, this neighborhood is much easier to handle than that appearing in \eqref{eq:LS_reg}.

In the next section, we discuss appropriate choices for $F$, $R$ and $C$ in \eqref{eq:LS_reg2}, by resorting to a Bayesian rationale.

\section{A Bayesian-to-variational Framework for Multi-view Reflectance Estimation}
\label{sec:var}

Following Mumford's Bayesian rationale for the variational formulation~\cite{Mumford1994}, let us now introduce a Bayesian-to-variational framework for estimating reflectance and lighting from multi-view images.

\subsection{Bayesian Inference}

Our problem consists in estimating the $m$ reflectance maps $\rho^i:\,\Omega^i \to \mathbb{R}$ and the $m$ lighting vectors ${\bm\sigma}^i \in \mathbb{R}^9$, given the $m$ images $I^i:\,\Omega^i \to \mathbb{R}$, $i \in \{1,\dots,m\}$. As we already stated, a maximum likelihood approach is hopeless, because a trivial solution arises. We rather resort to Bayesian inference, estimating $(\{\rho^i\}_i,\{\bm\sigma^i\}_i)$ as the maximum a posteriori (MAP) of the distribution
\begin{align}
	& \mathcal{P}(\{\rho^i\}_i,\{\bm\sigma^i\}_i \vert \{I^i\}_i) \nonumber \\
	& \qquad = \frac{\mathcal{P}(\{I^i\}_i \vert \{\rho^i\}_i,\{\bm\sigma^i\}_i ) \, \mathcal{P}(\{\rho^i\}_i,\{\bm\sigma^i\}_i)}{\mathcal{P}(\{I^i\}_i)},
\label{eq:MAP}
\end{align}
where the denominator is the evidence, which can be discarded since it depends neither on the reflectance nor on the lighting, and the factors in the numerator are the likelihood and the prior, respectively.

\paragraph{Likelihood.} The image formation model~\eqref{eq:Ii} is never strictly satisfied in practice, due to noise, cast-shadows and possibly slightly specular surfaces. We assume that such deviations from the model can be represented as independent (with respect to pixels and views) Laplace laws\footnote{We consider the Laplace law here because: i) since it has higher tails than the Gaussian, it allows for sparse outliers to the Lambertian model such as cast-shadows or specularities; ii) it yields convex optimization problems, unlike other heavy-tailed distributions such as Cauchy or $t$ distributions.} with zero mean and scale parameter $\alpha$:
\begin{align}
	& \mathcal{P}(\{I^i\}_i \vert \{\rho^i\}_i,\{\bm\sigma^i\}_i ) \nonumber \\
	& \quad = \prod_{i=1}^m
\left(\frac{1}{2 \alpha}\right)^{\lvert\Omega^i\rvert} \exp\left\{- \frac{1}{\alpha} \left\| \rho^i \, {\bm \sigma}^i \cdot {\bm \nu}^i - I^i \right\|_{i,1}
 \right\} \nonumber \\
	& \quad = \left(\frac{1}{2 \alpha}\right)^{\sum_{i=1}^m \lvert\Omega^i\rvert} \!\!\!\exp\left\{- 
 \frac{1}{\alpha} \sum_{i=1}^m \left\| \rho^i \, {\bm \sigma}^i \cdot {\bm \nu}^i - I^i \right\|_{i,1}
 \right\}
\end{align}
where $\|\cdot\|_{i,p}$, $p \geq 0$, is the $\ell^p$-norm over $\Omega^i$ and $\lvert\Omega^i\rvert$ is the cardinality of $\Omega^i$.

\paragraph{Prior.} Since the reflectance maps $\{\rho^i\}_i$ are independent from the lighting vectors $\{\bm\sigma^i\}_i$, the prior can be factorized to $ \mathcal{P}(\{\rho^i\}_i,\{\bm\sigma^i\}_i) = \mathcal{P}(\{\rho^i\}_i) \mathcal{P}(\{\bm\sigma^i\}_i)$. Since the lighting vectors are independent from each other, the prior distribution of the lighting vectors factorizes to $\mathcal{P}(\{\bm\sigma^i\}_i) = \prod_{i=1}^m \mathcal{P}(\bm\sigma^i)$. As each lighting vector is unconstrained, we can consider the same uniform distribution \ie $\mathcal{P}(\bm\sigma^i) = \tau$, independently from the view index $i$. This distribution being independent from the unknowns, we can discard the lighting prior from the inference process. Regarding the reflectance maps, we follow the retinex theory \cite{Land1971}, and consider each of them as piecewise-constant. The natural prior for each such map is thus the Potts model:
\begin{equation}
	\mathcal{P}(\rho^i) = K^i \exp\left\{ - \frac{1}{\beta^i} \left\| \nabla \rho^i \right\|_{i,0} \right\}
\label{eq:Potts}
\end{equation}
where $\nabla \rho^i(\mathbf{p}) = \left[\partial_x \rho^i(\mathbf{p}),\partial_y \rho^i(\mathbf{p}) \right]^\top$ represents the gradient of $\rho^i$ at pixel $\mathbf{p}$ (approximated, in practice, using first-order forward stencils with a Neumann boundary condition), and with $K^i$ a normalization coefficient and $\beta^i$ a scale parameter. Note that we use the abusive $\ell^0$-norm notation $\| \nabla \rho^i\|_{i,0}$ to denote:
\begin{equation}
	\left\| \nabla \rho^i \right\|_{i,0} = \sum_{\mathbf{p} \in \Omega^i} \sum_{\mathbf{q} \in \mathcal{V}^i(\mathbf{p})} f\left( \rho^i(\mathbf{p}) - \rho^i(\mathbf{q}) \right)
\end{equation}
with $f(x) = 1$ if $x\neq 0$, and $f(x) = 0$ otherwise.

The $m$ reflectance maps are obviously not independent: the reflectance, which characterizes the surface, should be independent from the view. It follows that the parameters $(K^i,\beta^i)$ are the same for each Potts model~\eqref{eq:Potts}, and that the reflectance prior $\mathcal{P}(\{\rho^i\}_i)$ can be taken as the product of $m$ independent distributions with the same parameters $(K,\beta)$:
\begin{equation}
	\mathcal{P}(\{\rho^i\}_i) = K^m \exp\left\{ - \frac{1}{\beta} \sum_{i=1}^m \left\| \nabla \rho^i \right\|_{i,0} \right\}
\end{equation}
but only if the coupling between the reflectance maps is enforced by the following linear constraint:
\begin{equation}
	C^{i,j}(\rho^i-\rho^j) = 0,~\forall (i,j) \in \{1,\dots,m\}^2,
\end{equation}
where $C^{i,j}$ is a $\Omega^i \times \Omega^j \to \{0,1\}$ ``correspondence function'', which is easily created from the (known) projection functions $\{\pi^i\}_i$ and the geometry, and which is defined as follows:
\begin{equation}
	C^{i,j}(\mathbf{p}^i,\!\mathbf{p}^j) \!=\! \begin{cases}
		1 & \text{if pixels $\mathbf{p}^i$ and $\mathbf{p}^j$ correspond} \\
		& \text{\quad to the same surface point}, \\
		0 & \text{otherwise}.
	\end{cases}
\label{eq:hard_constraint}
\end{equation}

Since maximizing the MAP probability~\eqref{eq:MAP} is equivalent to minimizing its negative logarithm, we eventually obtain the following constrained variational problem, which explicits the functions $F$, $R$ and $C$ in~\eqref{eq:LS_reg2}:
\begin{align}
	& \min_{\substack{\{\rho^i:\,\Omega^i \to \mathbb{R} \}_i \\ \{{\bm \sigma^i} \in \mathbb{R}^9 \}_i }} \sum_{i=1}^m \left\| \rho^i \, {\bm \sigma}^i \cdot {\bm \nu}^i - I^i \right\|_{i,1} + \lambda \sum_{i=1}^m \left\| \nabla \rho^i \right\|_{i,0} \nonumber \\
	& \text{s.t.}\quad C^{i,j}(\rho^i-\rho^j) = 0,~\forall (i,j) \in \{1,\dots,m\}^2,
\label{eq:var_1}
\end{align}
where $\lambda = \alpha / \beta$ and where we neglect all the normalization coefficients.

\subsection{Relationship with Cartoon + Texture Decomposition}

Applying a logarithm transformation to both sides of~\eqref{eq:Ii}, we obtain:
\begin{equation}
	\tilde{I}^i(\mathbf{p}) = \tilde{\rho}^i(\mathbf{p}) + \log\left({\bm \sigma}^i \cdot {\bm \nu}^i (\mathbf{p})\right),
\end{equation}
where the tilde notation is used as a shortcut for the logarithm.

By applying the exact same Bayesian-to-variational rationale, we would end up with the following variational problem:
\begin{align}
	& \min_{\substack{\{\tilde{\rho}^i:\,\Omega^i \to \mathbb{R} \}_i \\ \{{\bm \sigma^i} \in \mathbb{R}^9 \}_i }} \sum_{i=1}^m \left\| \tilde{\rho}^i +\log\left({\bm \sigma}^i \cdot {\bm \nu}^i\right) - \tilde{I}^i \right\|_{i,1} + \lambda \sum_{i=1}^m \left\| \nabla \tilde{\rho}^i \right\|_{i,0} \nonumber \\
	& \text{s.t.}\quad C^{i,j}(\tilde{\rho}^i-\tilde{\rho}^j) = 0,~\forall (i,j) \in \{1,\dots,m\}^2,
\label{eq:CT}
\end{align}

The variational problem~\eqref{eq:CT} can be interpreted as a multi-view cartoon + texture decomposition problem, where each log-image $\tilde{I}$ is decomposed into a component $C^i := \tilde{\rho}^i$ which is piecewise-smooth (``cartoon'', here the log-reflectance), and a component $T^i := \log\left({\bm \sigma}^i \cdot {\bm \nu}^i\right)$ which contains higher-frequency details (``texture'', here the log-shading). In contrast with conventional methods for such a task, the present one uses an explicit shading model for the texture term.

Note however that such a decomposition is justified only if the $\log$-images $\tilde{I}^i$ are considered. If using the original images $I^i$, our framework should rather be considered as a multi-view cartoon ``${\bm \times}$'' texture decomposition framework.

\subsection{Bi-convex Relaxation of the Variational Model~\eqref{eq:var_1}}
\label{sec:simplification}

Problem~\eqref{eq:var_1} is a non-convex (due to the $\ell^0$-regularizers), non-smooth (due to the $\ell^0$-regularizers and to the $\ell^1$-fidelity term). Although some efforts have recently been devoted to the resolution of optimization problems involving $\ell^0$-regularizers~\cite{Storath2014}, we prefer to keep the optimization simple, and approximate these by (convex, but non-smooth) anisotropic total variation terms:
\begin{equation}
	\sum_{i=1}^m \left\| \nabla \rho^i \right\|_{i,0} \approx \sum_{i=1}^m \left\| \nabla \rho^i \right\|_{i,1}.
\end{equation}
Besides, the correspondence function may be slightly inaccurate in practice, due to errors in the prior geometry estimation obtained via multi-view stereo. Therefore, we turn the linear constraint in~\eqref{eq:var_1} into an additional term. Eventually, we replace the non-differentia\-ble absolute values arising from the $\ell^1$-norms by the (differentiable) Moreau envelope \ie the Huber loss\footnote{We use $\delta = 10^{-4}$, in the experiments.}:
\begin{equation}
	\lvert x \rvert \approx \phi_\delta(x) :=
	\begin{cases}
		\dfrac{x^2}{2 \, \delta}, & \lvert x \rvert \leq \delta \\
		\lvert x \rvert - \dfrac{\delta}{2}, & \lvert x \rvert > \delta
	\end{cases}
\label{eq:def_phi_delta}
\end{equation}

Altogether, this yields the following smooth, bi-convex variational problem:
\begin{align}
	& \min_{\substack{\rho := \{\rho^i:\,\Omega^i \to \mathbb{R} \}_i \\ \vec{\sigma}:=\{{\bm \sigma^i} \in \mathbb{R}^9 \}_i }} \!\!\!
\varepsilon(\rho,{\bm \sigma}) := \sum_{i=1}^m \! \sum_{\mathbf{p} \in \Omega^i} \!\! \phi_\delta \!\left( \rho^i(\mathbf{p}) \, {\bm\sigma}^i \!\cdot\! {\bm \nu}^i(\mathbf{p}) - I^i(\mathbf{p}) \right) \nonumber \\
	& + \lambda \sum_{i=1}^m \sum_{\mathbf{p} \in \Omega^i} \left[ \phi_\delta \! \left(\partial_x \rho^i(\mathbf{p}) \right) + \phi_\delta \! \left( \partial_y \rho^i(\mathbf{p}) \right) \right] \nonumber \\
	& + \mu \mathop{\sum\sum}_{1 \leq i<j \leq m} \sum_{\mathbf{p}^i \in \Omega^i} \sum_{\mathbf{p}^j \in \Omega^j} \!\! C_{i,j}(\mathbf{p}^i,\mathbf{p}^j) \, \phi_\delta \!\left( \rho^i(\mathbf{p}^i) - \rho^j(\mathbf{p}^j) \right).
\label{eq:var}
\end{align}
In Equation~\eqref{eq:var}, the first term ensures photometric consistency (in the sense of the Huber loss function), the second one ensures reflectance smoothness (smoothed anisotropic total variation), and the third term ensures multi-view consistency of the reflectance estimates (again, in the sense of the Huber loss function). At last, $\lambda$ and $\mu$ are tunable hyper-parameters controlling the reflectance smoothness and the multi-view consistency, respectively.

\section{Alternating Majorization-minimization for Solving~\eqref{eq:var}}
\label{sec:numerics}

\begin{figure}[!ht]
\centering
	\def\svgwidth{0.85\linewidth}
		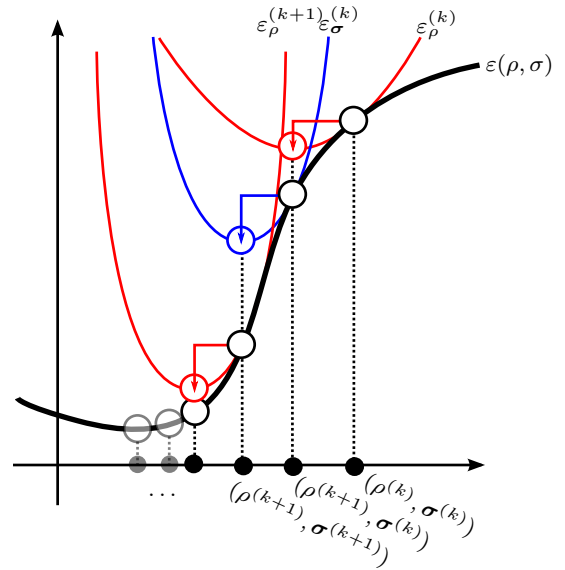
\caption{Sketch of the proposed alternating majorization-minimization solution. The partially freezed energies $\varepsilon(\cdot,\vec{\sigma})$ and $\varepsilon(\rho,\cdot)$ are locally majorized by the quadratic functions $\varepsilon_{\rho}$ (in red) and $\varepsilon_{\vec{\sigma}}$ (in blue). Then, these quadratic majorants are (globally) minimized and the process is repeated until convergence is reached.}
\label{fig:sketch_optim}
\end{figure}

To solve~\eqref{eq:var}, we propose an alternating ma\-jo\-ri\-za\-tion-minimization method, which combines alternating and majorization-minimization optimization techniques. As sketched in Figure~\ref{fig:sketch_optim}, this algorithm works as follows. Given an estimate $(\rho^{(k)},{\bm \sigma}^{(k)})$ of the solution at iteration $(k)$, the lighting vectors and the reflectance maps are successively updated according to:
\begin{align}
	\rho^{(k+1)} & = \underset{\rho}{\operatorname{argmin~}} \varepsilon_\rho^{(k)}(\rho), \label{eq:11} \\
	\vec{\sigma}^{(k+1)} & = \underset{\vec{\sigma}}{\operatorname{argmin~}} \varepsilon_{\vec{\sigma}}^{(k)}({\bm \sigma}), \label{eq:12}
\end{align}
where $\varepsilon_\rho^{(k)}$ and $\varepsilon_{\vec{\sigma}}^{(k)}$ are local quadratic majorants of $\varepsilon(\cdot,{\bm \sigma^{(k)}})$ and $\varepsilon(\rho^{(k+1)},\cdot)$ around, respectively, $\rho^{(k)}$ and $\vec{\sigma}^{(k)}$. Then, the process is repeated until convergence.

To this end, let us first remark that the function
\begin{equation}
	\psi_\delta(x;x_0) =
	\begin{cases}
		\dfrac{ x^2}{2 \, \delta}, & \quad \lvert x_0 \rvert \leq \delta, \\
		\dfrac{ x^2}{2 \, \lvert x_0 \rvert} + \dfrac{\lvert x_0 \rvert}{2} - \dfrac{\delta}{2}, & \quad \lvert x_0 \rvert > \delta,
	\end{cases}
\label{eq:psi_delta}
\end{equation}
is such that $\psi_\delta(x_0;x_0) = \phi_\delta(x_0)$, and is a proper local quadratic majorant of $\phi_\delta$ around $x_0$, $\forall x_0 \in \mathbb{R}$. This is easily verified if $\lvert x_0 \rvert \leq \delta$, from the definition \eqref{eq:def_phi_delta} of $\phi_\delta$. If $\lvert x_0 \rvert > \delta$, the difference $\psi_\delta(x;x_0) - \phi_\delta(x)$ writes:
\begin{equation}
	 \begin{cases}
		\dfrac{\left( \lvert x_0 \rvert-\delta \right) \left(\lvert x_0 \rvert \, \delta - x^2 \right)}{2 \, \lvert x_0 \rvert \, \delta}, & \quad \lvert x \rvert \leq \delta, \\
		\dfrac{(\lvert x \rvert-\lvert x_0 \rvert)^2}{2 \, \lvert x_0 \rvert}, & \quad \lvert x \rvert > \delta,
	\end{cases}
\end{equation}
which is positive in any case.

Therefore, the function
\begin{align}
	& \varepsilon_\rho^{(k)}(\rho) := \! \sum_{i=1}^m \sum_{\mathbf{p} \in \Omega^i} \!\! \psi_\delta \!\left( \! \rho^{i}(\mathbf{p}) \, {\bm\sigma}^{i,(k)} \!\!\cdot\! {\bm \nu}^i(\mathbf{p}) \!-\! I^i(\mathbf{p}); r^{i,(k),(k)} \!\right) \nonumber \\
	& \quad + \lambda \sum_{i=1}^m \sum_{\mathbf{p} \in \Omega^i} \left[ \psi_\delta \left(\partial_x \rho^i(\mathbf{p}) ; \partial_x \rho^{i,(k)}(\mathbf{p}) \right) \right. \nonumber \\
	& \quad\qquad\quad\qquad\,\, \left. + \psi_\delta \left( \partial_y \rho^i(\mathbf{p}); \partial_y \rho^{i,(k)}(\mathbf{p}) \right) \right] \nonumber \\
	& \quad+ \mu \mathop{\sum\sum}_{1\leq i<j \leq m} \, \sum_{\mathbf{p}^i \in \Omega^i} \, \sum_{\mathbf{p}^j \in \Omega^j} C_{i,j}(\mathbf{p}^i,\mathbf{p}^j) \nonumber \\
	& \qquad \psi_\delta \left( \rho^i(\mathbf{p}^i) - \rho^j(\mathbf{p}^j); \rho^{i,(k)}(\mathbf{p}^i) - \rho^{j,(k)}(\mathbf{p}^j)\right),\label{eq:major_rho}
\end{align}
with
\begin{equation}
	r^{i,(k_1),(k_2)} = \rho^{i,(k_1)}(\mathbf{p}) \, {\bm\sigma}^{i,(k_2)} \cdot {\bm \nu}^i(\mathbf{p}) - I^i(\mathbf{p}),
\end{equation}
is a local quadratic majorant of $\varepsilon(\cdot,{\bm \sigma^{(k)}})$ around $\rho^{(k)}$ which is suitable for the update~\eqref{eq:11}.

Similarly, the function
\begin{align}
	& \varepsilon_\sigma^{(k)}({\bm \sigma}) \!:=\!\!\sum_{i=1}^m \! \sum_{\mathbf{p} \in \Omega^i} \!\!\! \psi_\delta \!\left( \! \rho^{i,(k+1)}(\mathbf{p}) {\bm\sigma}^i \!\!\cdot\! {\bm \nu}^i(\mathbf{p}) \!-\!I^i(\mathbf{p}) ; \!\! r^{i,(k+1),(k)} \! \right) \nonumber \\
	& + \lambda \sum_{i=1}^m
	 \sum_{\mathbf{p} \in \Omega^i} \left[ \phi_\delta \! \left(\partial_x \rho^{i,(k+1)}(\mathbf{p}) \right) + \phi_\delta \! \left( \partial_y \rho^{i,(k+1)}(\mathbf{p}) \right) \right] \nonumber \\
	& + \mu\! \mathop{\sum\sum}_{1\leq i<j \leq m} \sum_{\mathbf{p}^i \in \Omega^i} \sum_{\mathbf{p}^j \in \Omega^j} \Big[ C_{i,j}(\mathbf{p}^i,\mathbf{p}^j) \nonumber \\
	& \qquad\qquad\qquad\quad \phi_\delta \left( \rho^{i,(k+1)}(\mathbf{p}^i) - \rho^{j,(k+1)} (\mathbf{p}^j) \right) \Big]
\label{eq:major_sigma}
\end{align}
is a local quadratic majorant of $\varepsilon(\rho^{(k+1)},\cdot)$ around $\vec{\sigma}^{(k)}$ which is suitable for the update~\eqref{eq:12}.

The update~\eqref{eq:11} then comes down to solving a large sparse linear least-squares problem, which we achieve by applying conjugate gradient iterations to the associated normal equations. Regarding~\eqref{eq:12}, it comes down to solving a series of $m$ independent small-scale linear least-squares problems, for instance by resorting to the pseudo-inverse.

We iterate the optimisation steps~\eqref{eq:11} and~\eqref{eq:12} until convergence or a maximum iteration number is reached, starting from the trivial solution of the non-regularized ($\lambda = \mu = 0$) problem. This non-regularized solution is attained by considering diffuse lighting (see~\eqref{eq:sigma_naive}) and using the input images as reflectance maps. In our experiments, we found $50$ iterations were always sufficient to reach a stable solution ($10^{-3}$ relative residual between two consecutive energy values $\varepsilon(\rho^{(k)},{\bm \sigma}^{(k)})$ and $\varepsilon(\rho^{(k+1)},{\bm \sigma}^{(k+1)})$).

\begin{figure}[!ht]
\centering
	\begin{tabular}{c}
		\includegraphics[width = 0.7\linewidth]{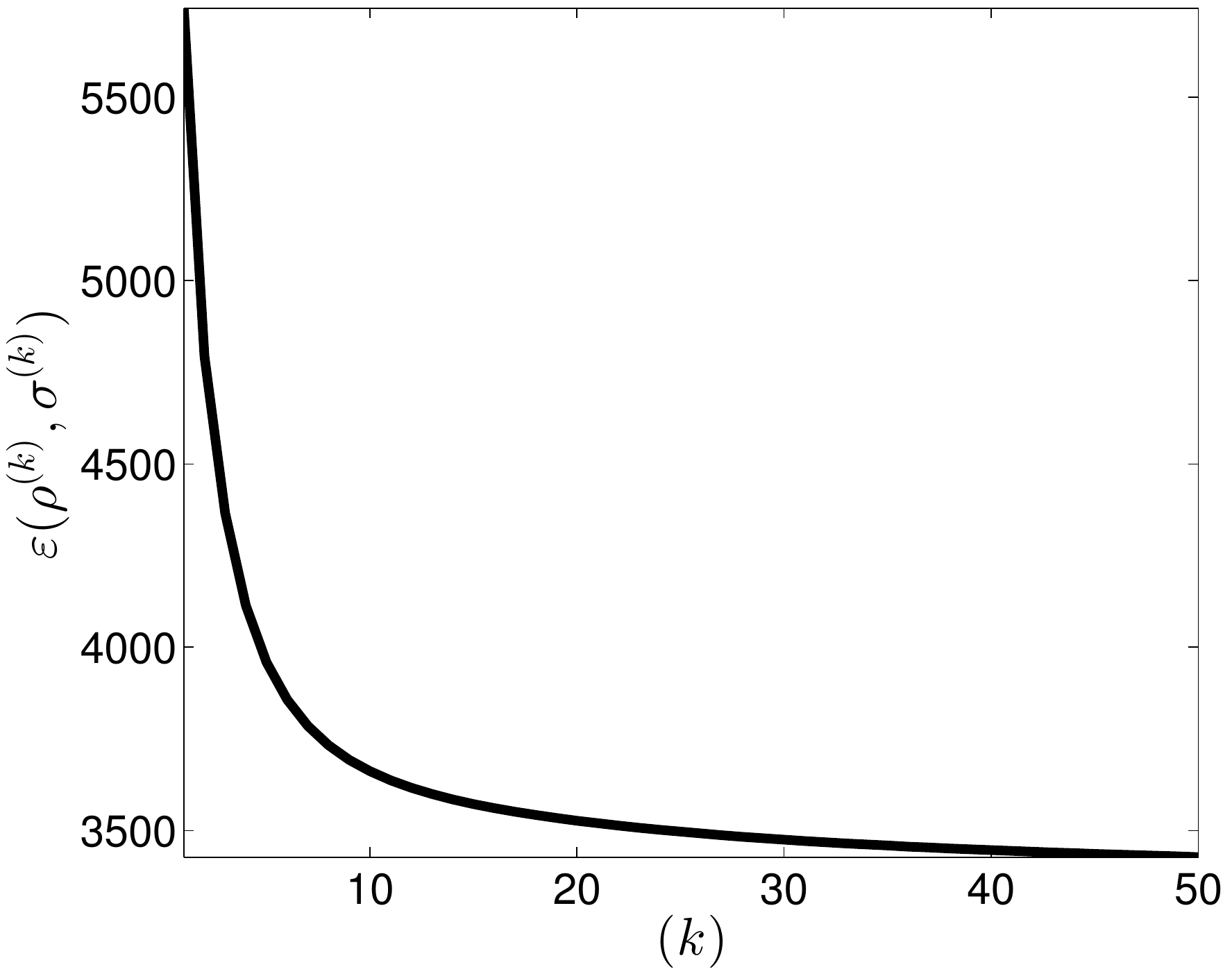} \\
		\includegraphics[width = 0.75\linewidth]{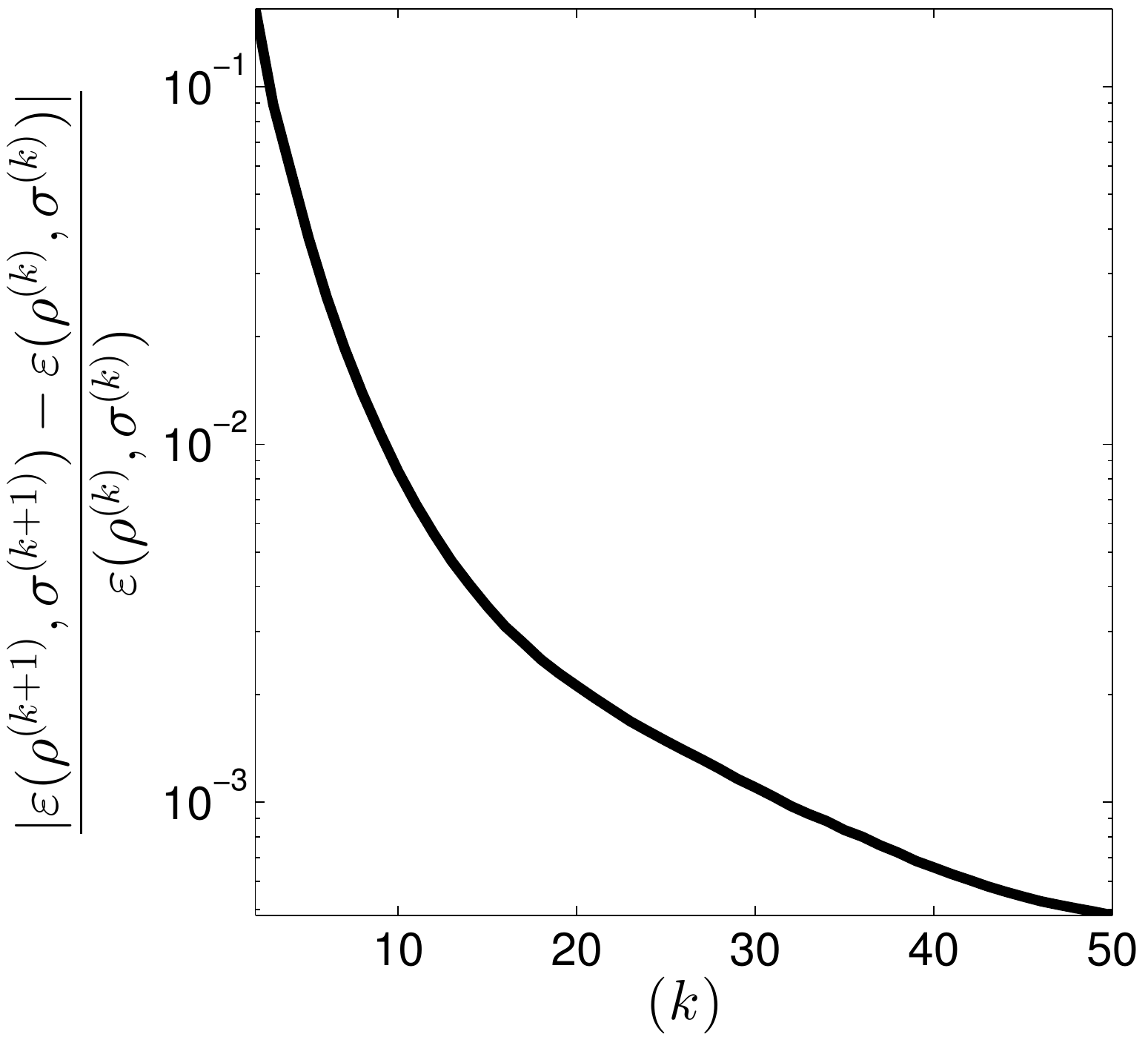}
	\end{tabular}
\caption{Top: evolution of the energy $\varepsilon(\rho^{(k)},{\bm \sigma}^{(k)})$ defined in~\eqref{eq:var}, in function of iterations $(k)$, concerning the test presented in Figure~\ref{fig:5}. Bottom: absolute value of the relative variation between two successive energy values. Our algorithm stops when this value is less than $10^{-3}$, which happens in less than 50 iterations and takes around 3 minutes on a recent i7 processor, with non-optimized Matlab codes for $m=13$ images of size $540 \times 960$.}
\label{fig:graphs}
\end{figure}

Proving convergence of our scheme is beyond the scope of this paper, but the proof could certainly be derived from that in~\cite{JMIV2017_LEDS}, where a similar alternating majorization-minimization called ``alternating reweigh\-ted least-squares'' is used. Note, however, that the convergence rate seems to be sublinear (see Figure~\ref{fig:graphs}), hence possibly faster numerical strategies could be explored in the future.

\section{Results}
\label{sec:results}

In this section, we evaluate the proposed variational method for multi-view reflectance estimation, on a variety of synthetic and real-world datasets. We start by a quantitative comparison of our results with two single-view methods, namely, the cartoon + texture decomposition method from~\cite{IpolCartoon} and the intrinsic image decomposition method from~\cite{Gehler2011}.

\subsection{Quantitative Evaluation on a Synthetic Dataset}

\begin{figure*}[!ht]
\begin{center}
	\begin{tabular}{cccc}
		\includegraphics[width = 0.21\linewidth]{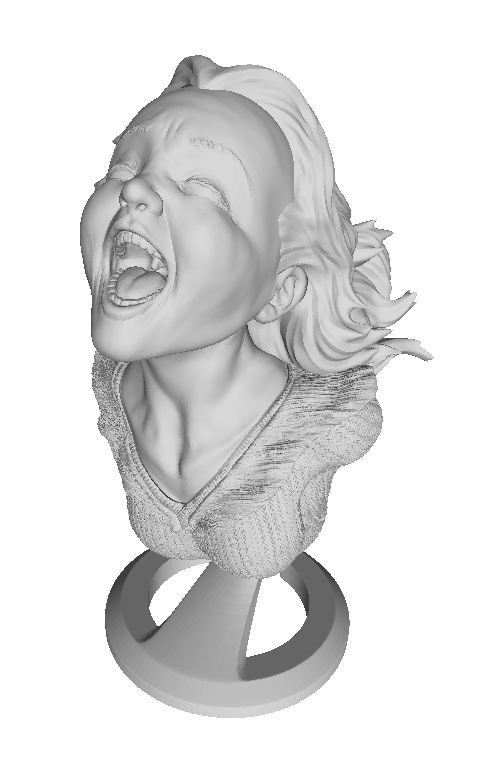} &
		\includegraphics[width = 0.21\linewidth]{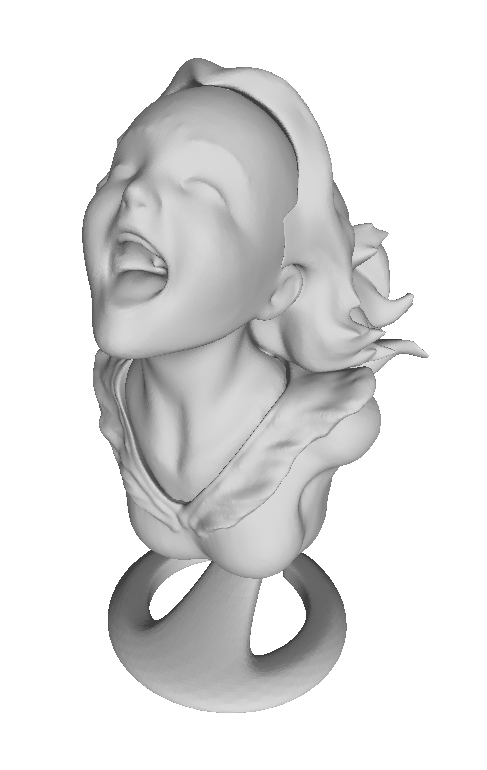} & \quad
		\includegraphics[width = 0.21\linewidth]{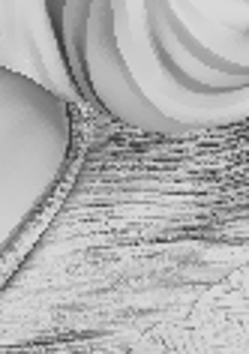} & \quad\quad
		\includegraphics[width = 0.215\linewidth]{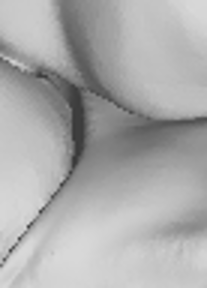} \\
		\small{(a)} & \small{(b)} & \quad \small{(c)} & \quad\quad \small{(d)}
	\end{tabular}
\end{center}
\caption{(a) 3D-shape used in the tests (the well-known ``Joyful Yell'' 3D-model), which will be imaged under two scenarios (see Figures~\ref{fig:1} and~\ref{fig:2}). (b) Same, after smoothing, thus less accurate. (c)-(d) Zooms of (a) and (b), respectively, near the neck.}
\label{fig:0}
\end{figure*}

\begin{figure*}[!ht]
\begin{center}
	\begin{tabular}{m{.22\textwidth} m{.23\textwidth} m{.23\textwidth} m{.23\textwidth}}
		\qquad Input images &
		\includegraphics[width = \linewidth, trim={10cm 2cm 6cm 0},clip]{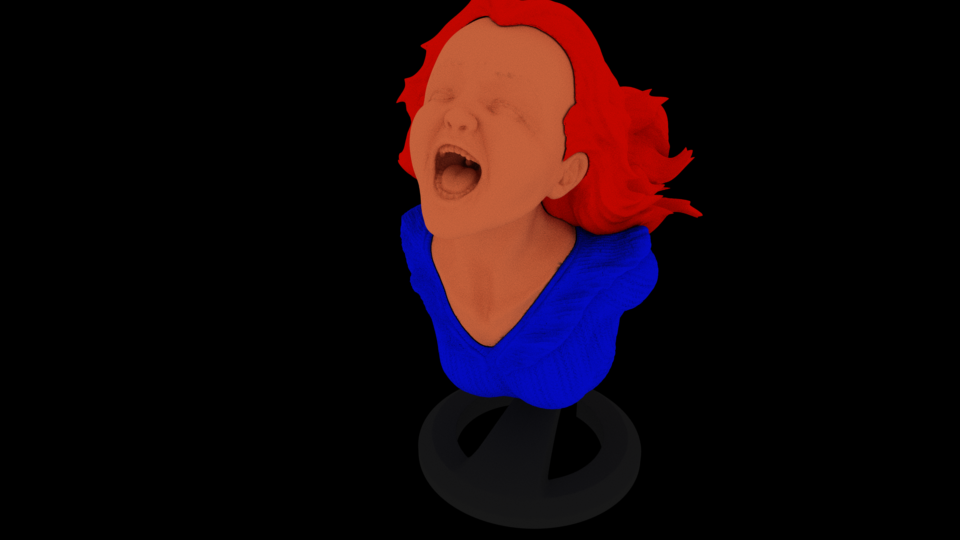} & \,
		\includegraphics[width = \linewidth, trim={8cm 2cm 8cm 0},clip]{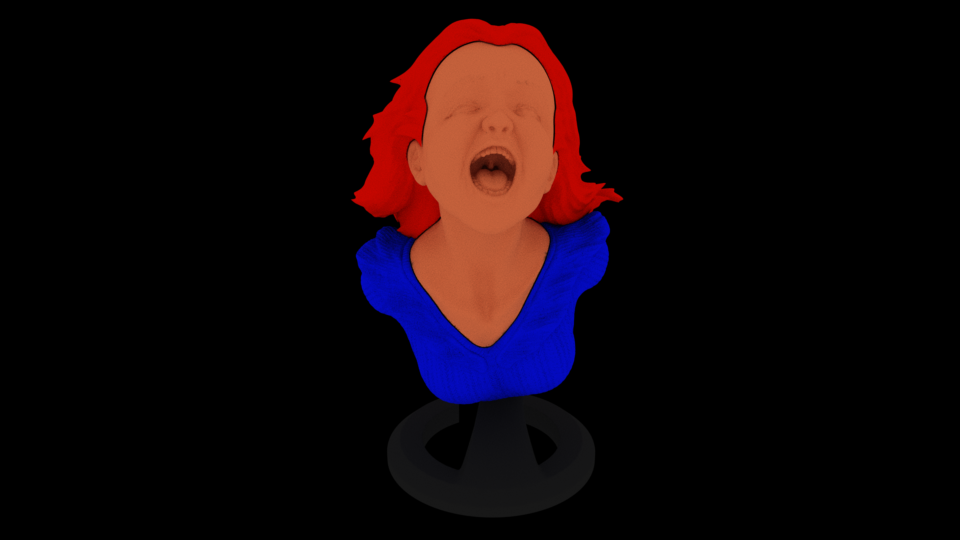} & \,
		\includegraphics[width = \linewidth, trim={10cm 2cm 6cm 0},clip]{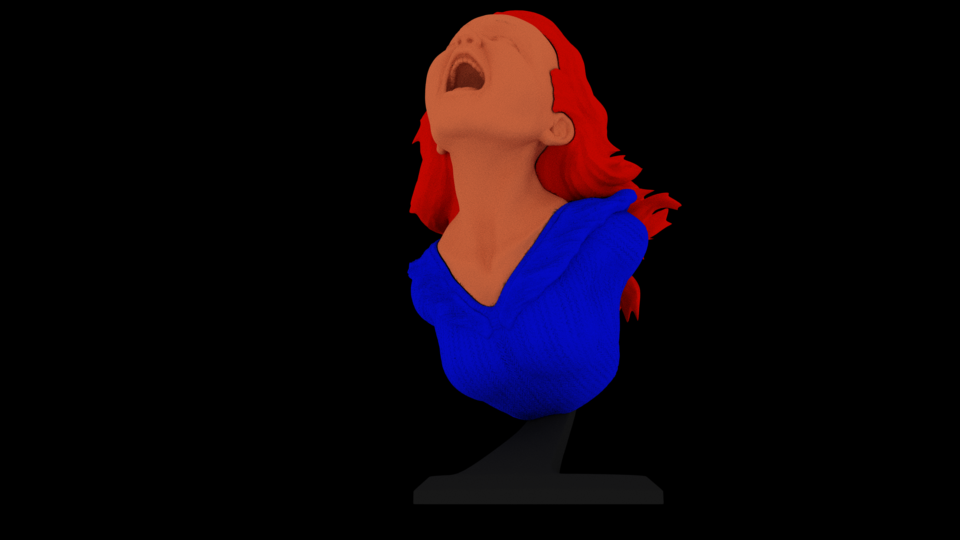} \\
		Cartoon + texture \cite{IpolCartoon} &
		\includegraphics[width = \linewidth, trim={10cm 2cm 6cm 0},clip]{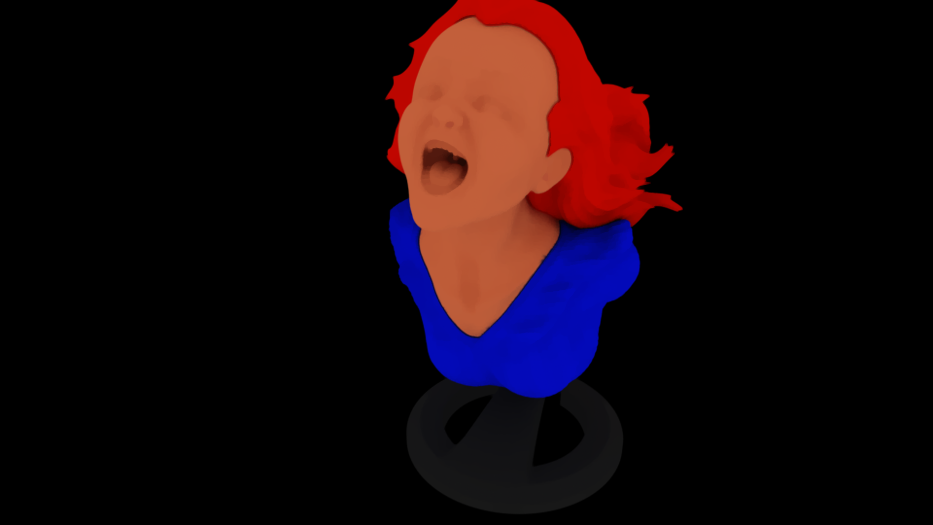} & \,
		\includegraphics[width = \linewidth, trim={8cm 2cm 8cm 0},clip]{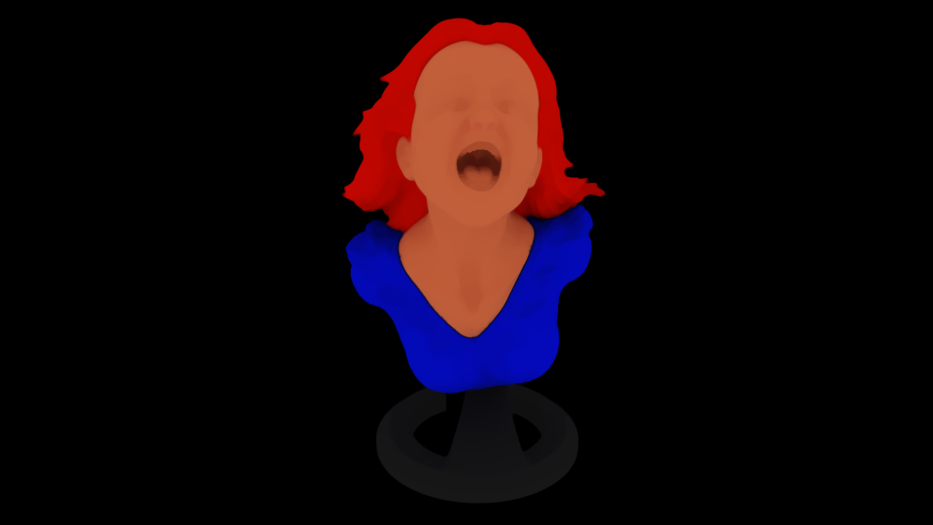} & \,
		\includegraphics[width = \linewidth, trim={10cm 2cm 6cm 0},clip]{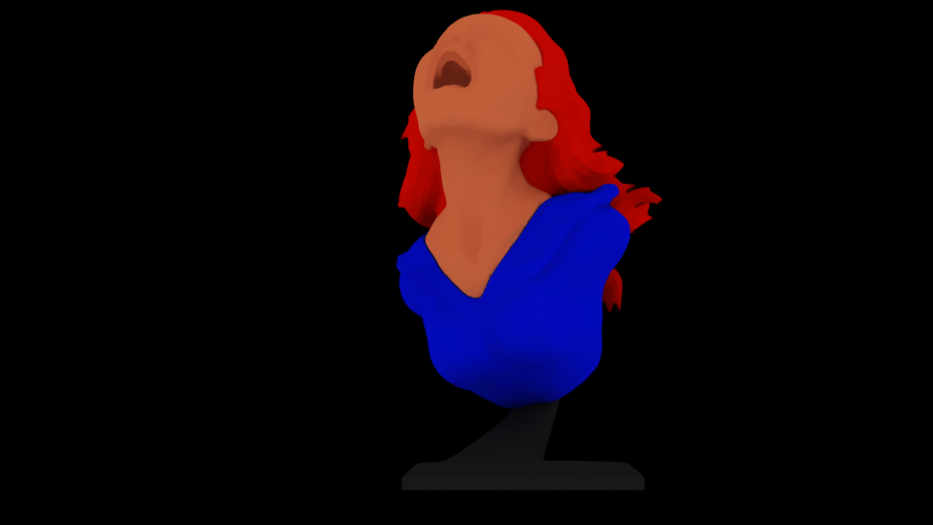} \\
		Intrinsic decomposition \cite{Gehler2011} &
		\includegraphics[width = \linewidth, trim={10cm 2cm 6cm 0},clip]{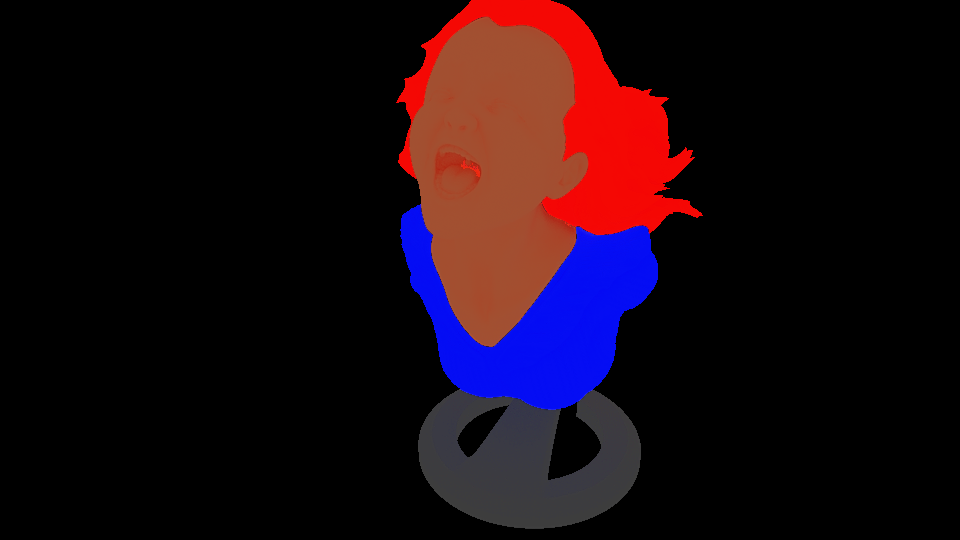} & \,
		\includegraphics[width = \linewidth, trim={8cm 2cm 8cm 0},clip]{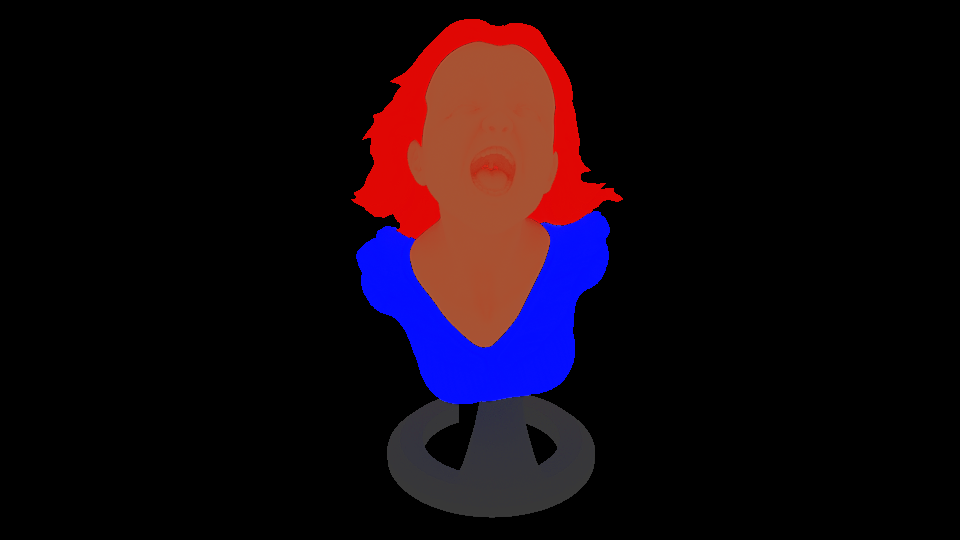} & \,
		\includegraphics[width = \linewidth, trim={10cm 2cm 6cm 0},clip]{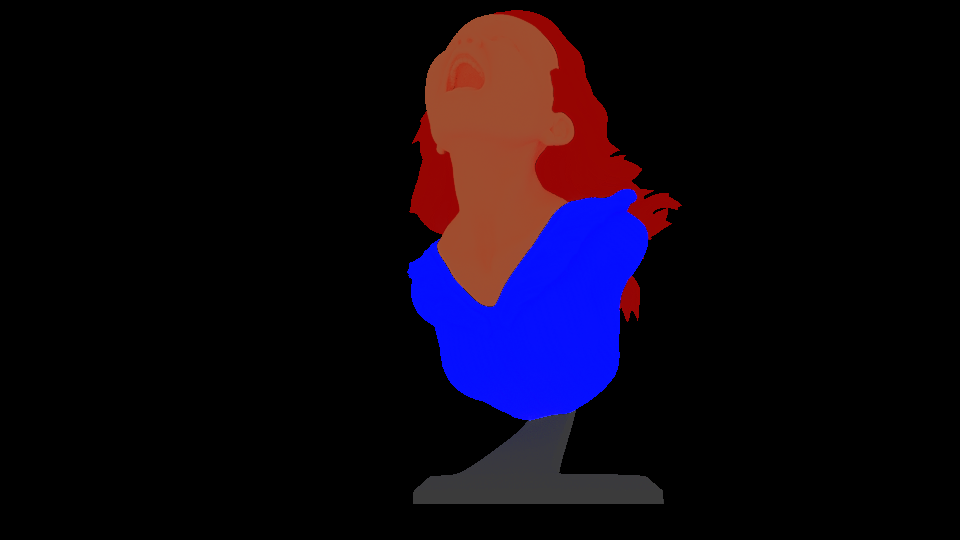} \\
		\qquad\qquad Ours &
		\includegraphics[width = \linewidth, trim={10cm 2cm 6cm 0},clip]{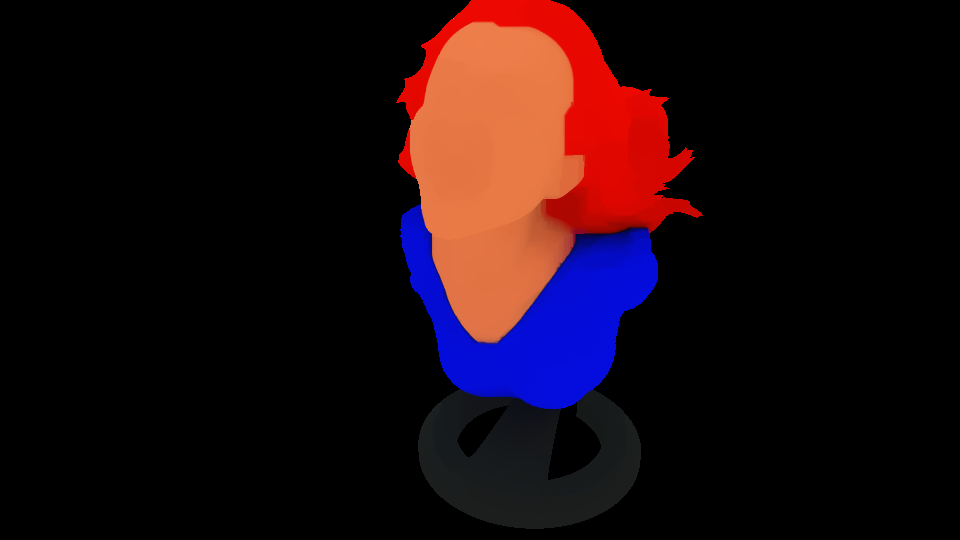} & \,
		\includegraphics[width = \linewidth, trim={8cm 2cm 8cm 0},clip]{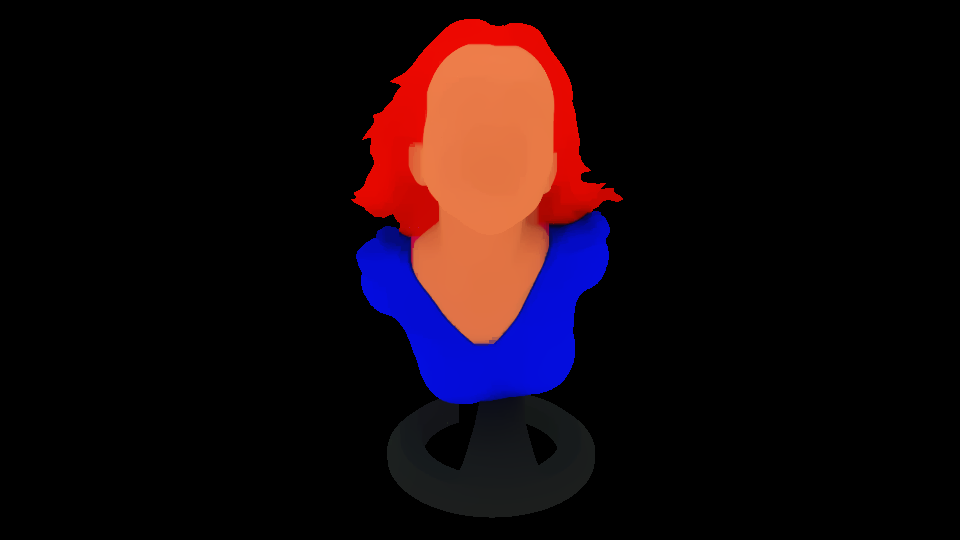} & \,
		\includegraphics[width = \linewidth, trim={10cm 2cm 6cm 0},clip]{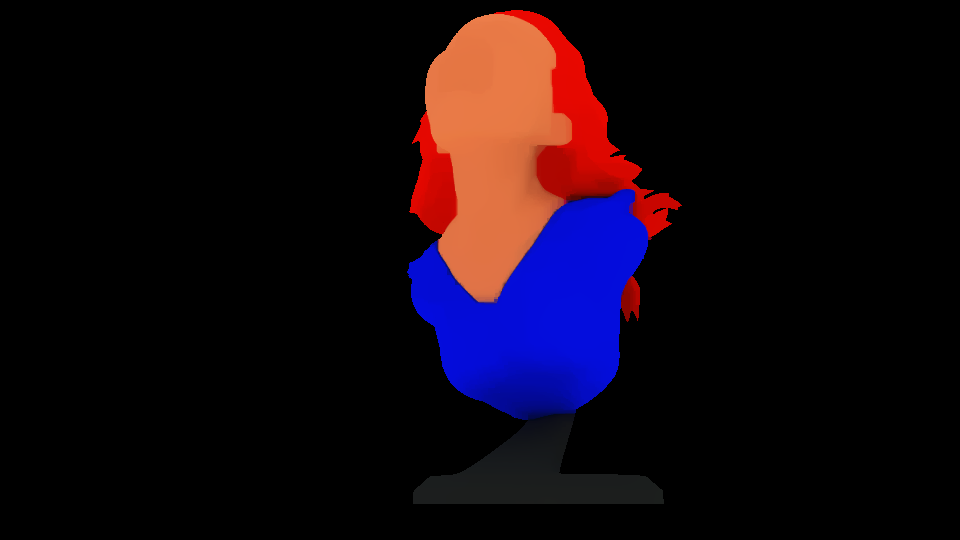} \\
		\qquad Ground truth &
		\includegraphics[width = \linewidth, trim={10cm 2cm 6cm 0},clip]{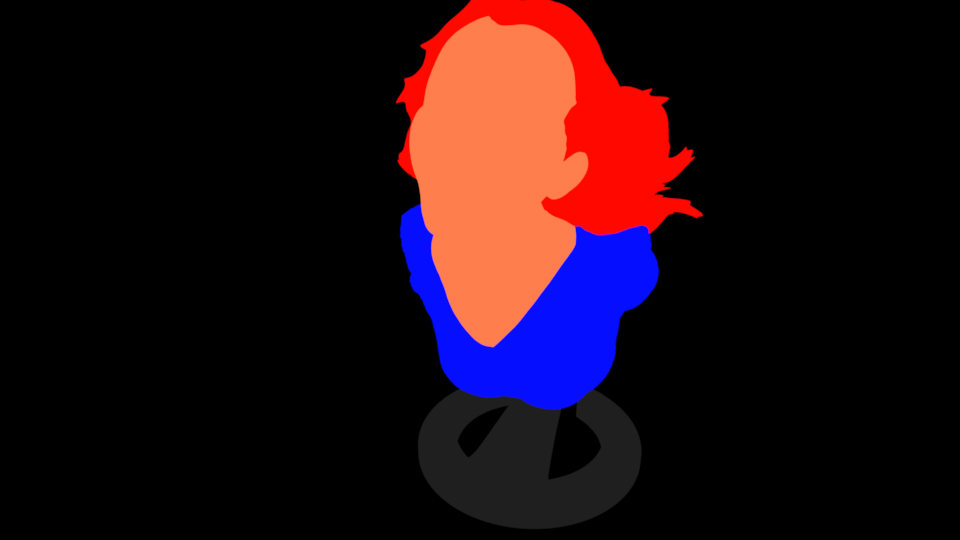} & \,
		\includegraphics[width = \linewidth, trim={8cm 2cm 8cm 0},clip]{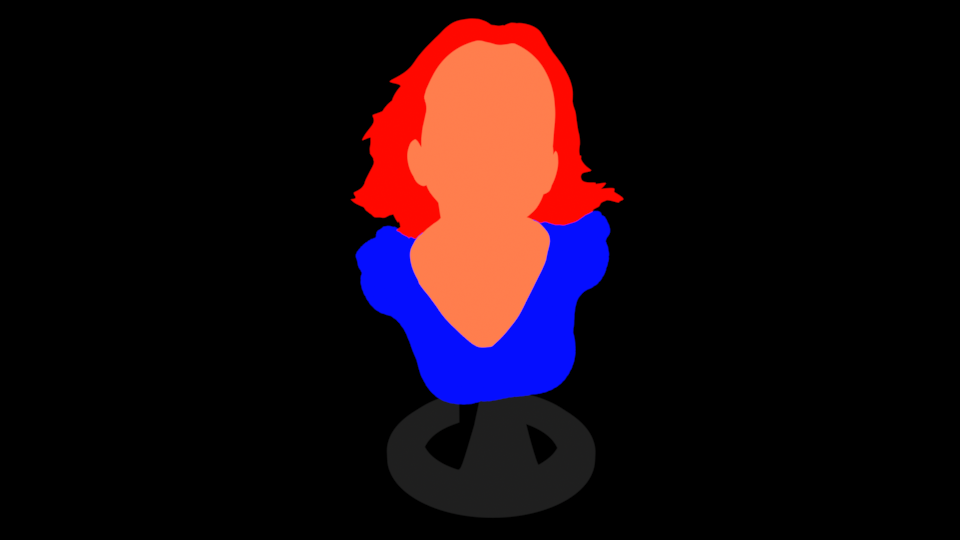} & \,
		\includegraphics[width = \linewidth, trim={10cm 2cm 6cm 0},clip]{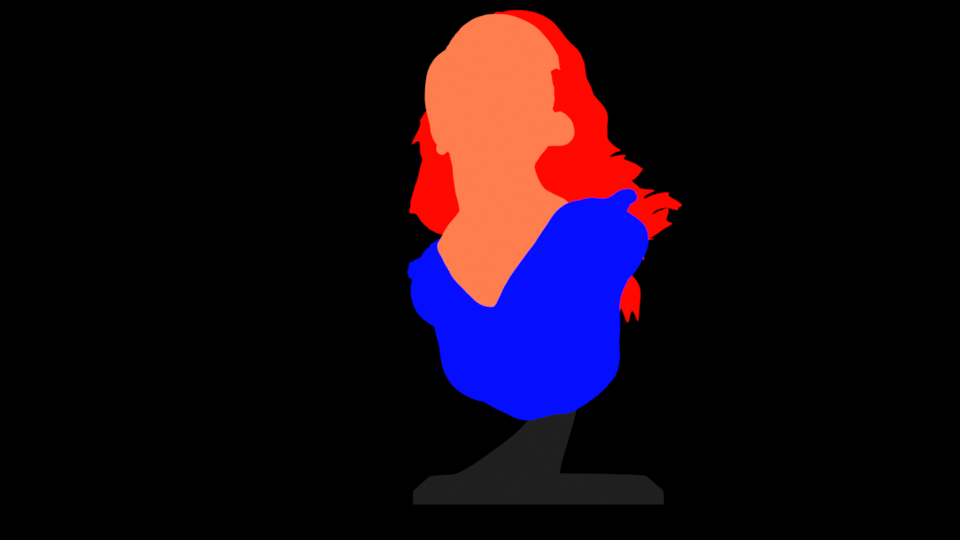}
	\end{tabular}
\end{center}
\caption{First row: three (out of $m=13$) synthetic views of the object of Figure \ref{fig:0}-a, computed with a purely-Lambertian reflectance taking only four different values (hair, face, shirt and plinth), illuminated by a ``skydome''. Second row: estimation of the reflectance using the cartoon + texture decomposition described in \cite{IpolCartoon} (with its parameter fixed to $0.4$). Third row: estimation of the reflectance using the method proposed in \cite{Gehler2011} (with 4 clusters). Forth row: estimation of the reflectance using the proposed approach (with $\lambda = 8$ and $\mu = 1000$). Fifth row: ground truth.}
\label{fig:1}
\end{figure*}

\begin{figure*}[!ht]
\begin{center}
	\begin{tabular}{ m{.22\textwidth} m{.23\textwidth} m{.23\textwidth} m{.23\textwidth}}
		\qquad Input images &
		\includegraphics[width = \linewidth, trim={10cm 2cm 6cm 0},clip]{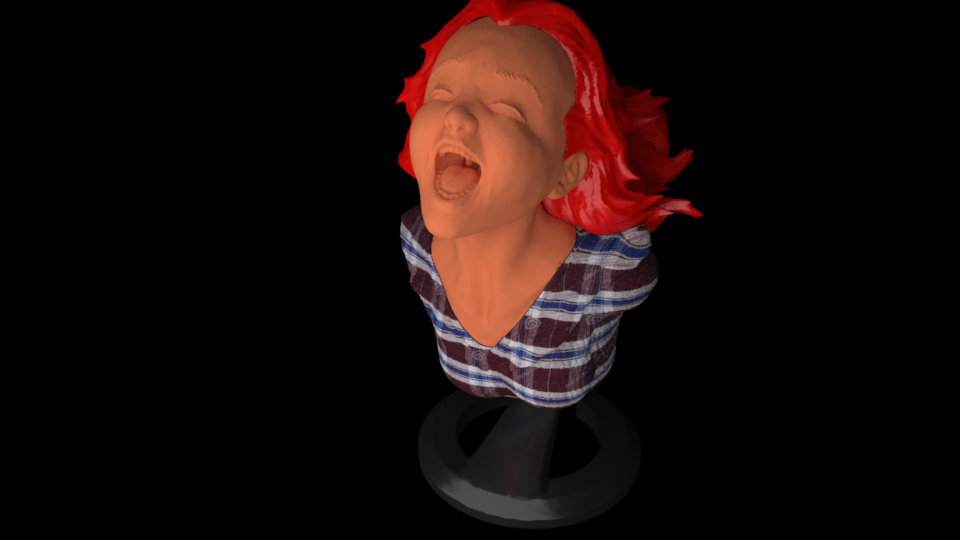} & \,
		\includegraphics[width = \linewidth, trim={8cm 2cm 8cm 0},clip]{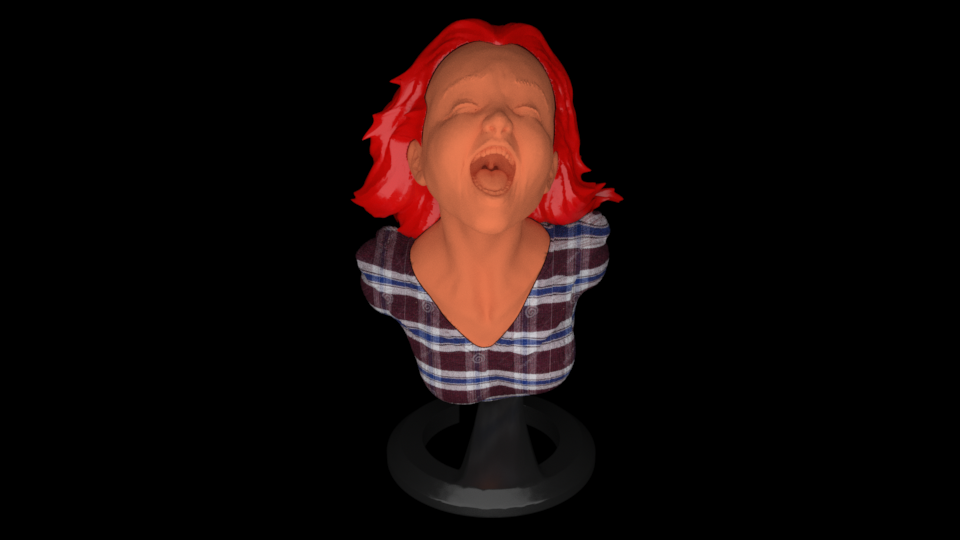} & \,
		\includegraphics[width = \linewidth, trim={10cm 2cm 6cm 0},clip]{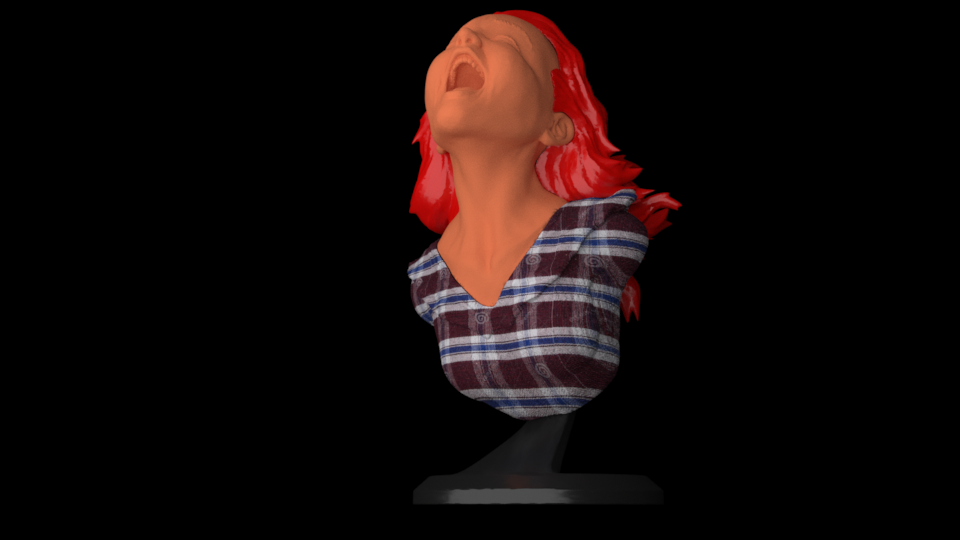} \\
		Cartoon + texture \cite{IpolCartoon} &
		\includegraphics[width = \linewidth, trim={10cm 2cm 6cm 0},clip]{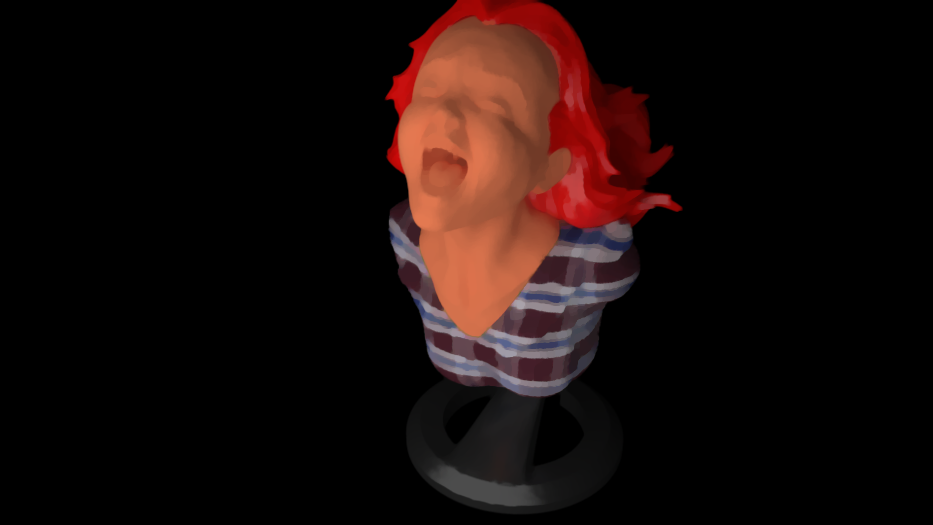} & \,
		\includegraphics[width = \linewidth, trim={8cm 2cm 8cm 0},clip]{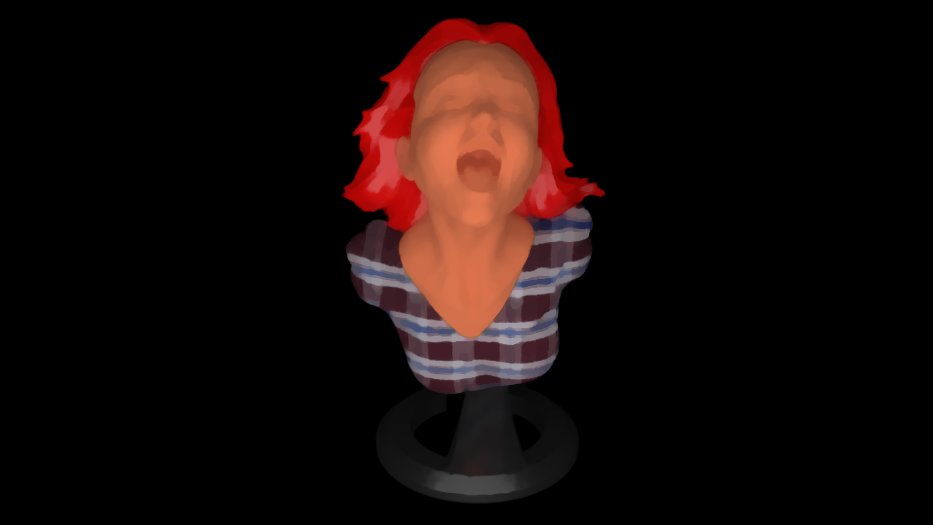} & \,
		\includegraphics[width = \linewidth, trim={10cm 2cm 6cm 0},clip]{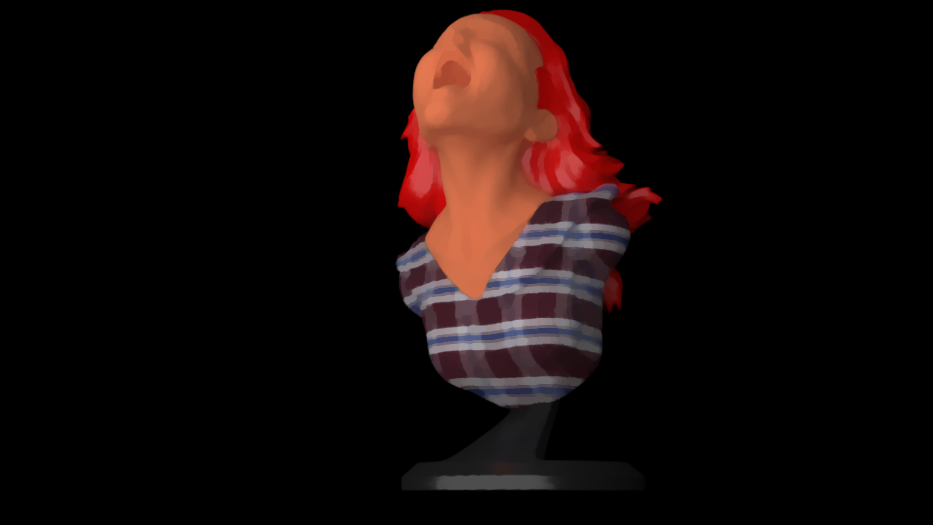} \\
		Intrinsic decomposition \cite{Gehler2011} &
		\includegraphics[width = \linewidth, trim={10cm 2cm 6cm 0},clip]{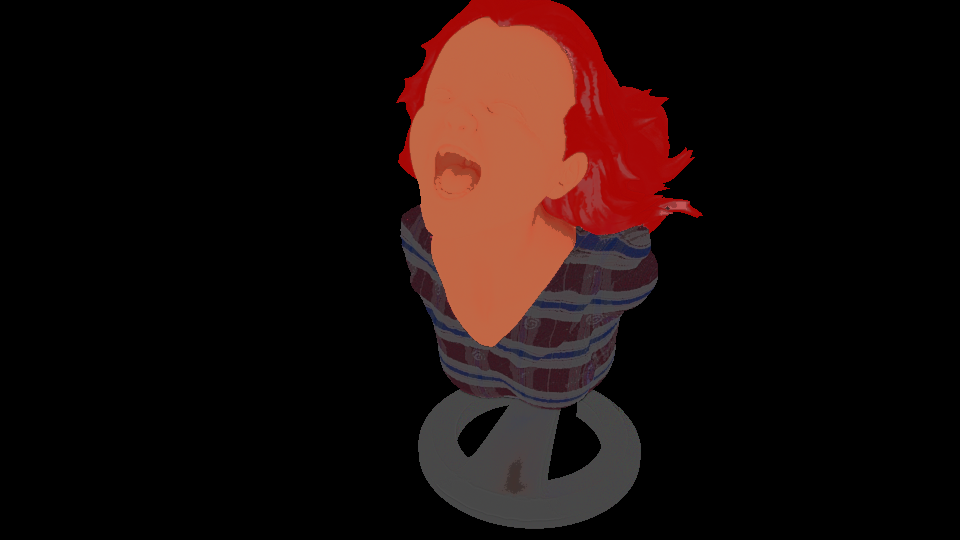} & \,
		\includegraphics[width = \linewidth, trim={8cm 2cm 8cm 0},clip]{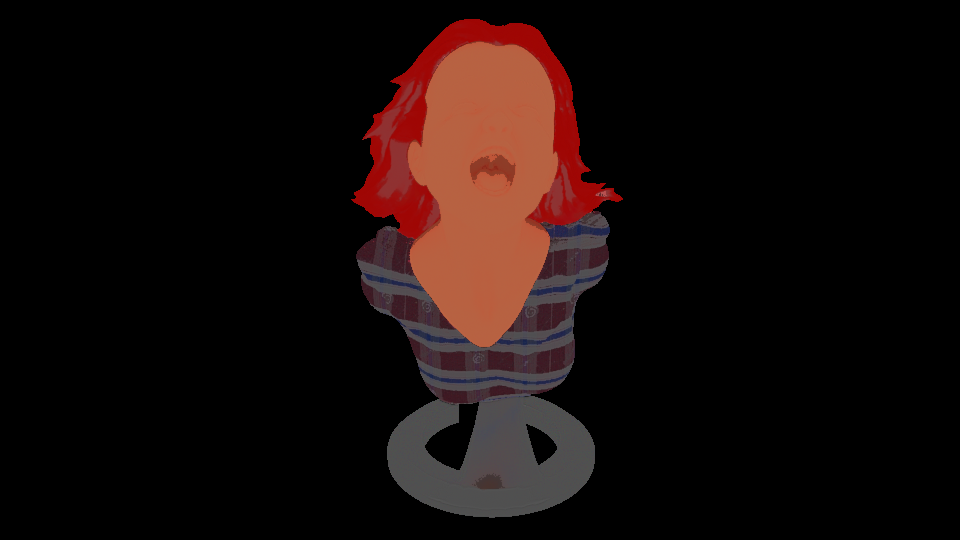} & \,
		\includegraphics[width = \linewidth, trim={10cm 2cm 6cm 0},clip]{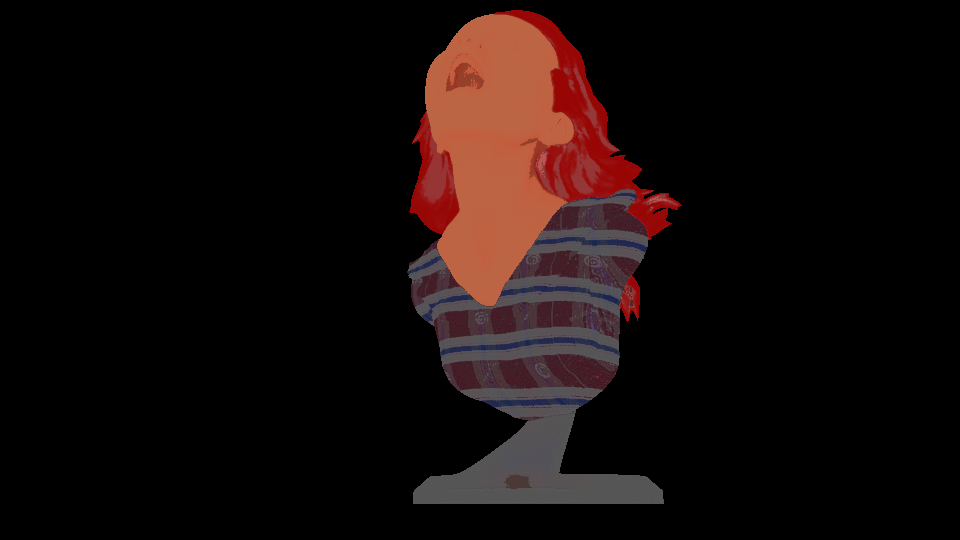} \\
		\qquad\qquad Ours &
		\includegraphics[width = \linewidth, trim={10cm 2cm 6cm 0},clip]{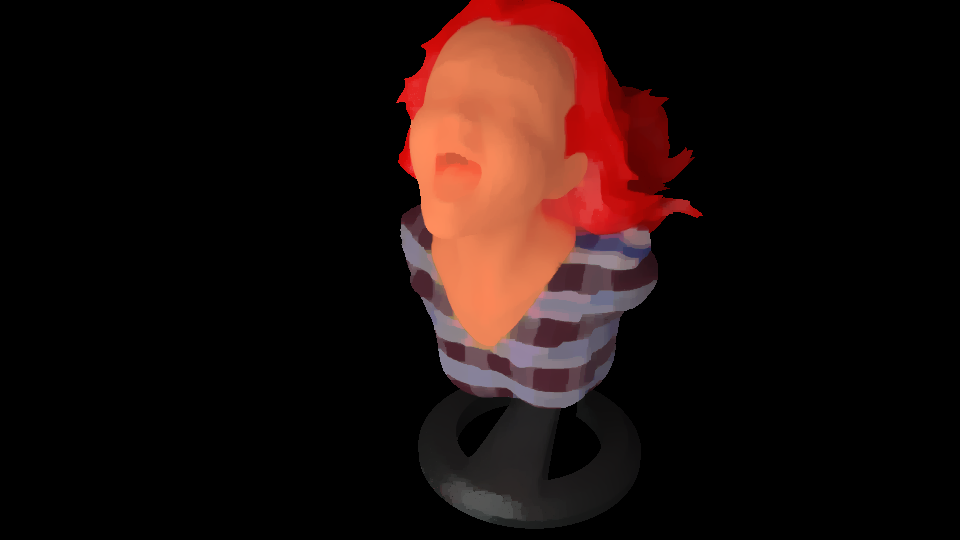} & \,
		\includegraphics[width = \linewidth, trim={8cm 2cm 8cm 0},clip]{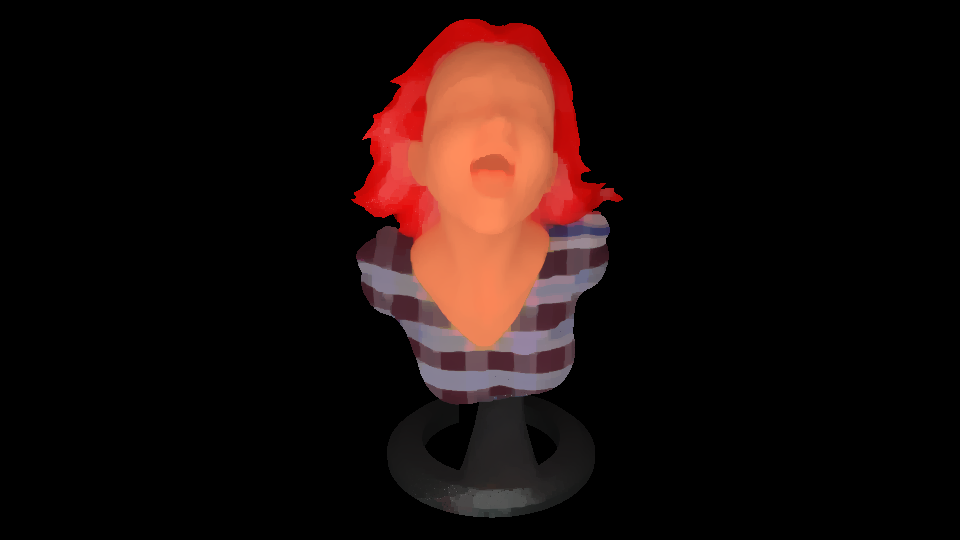} & \,
		\includegraphics[width = \linewidth, trim={10cm 2cm 6cm 0},clip]{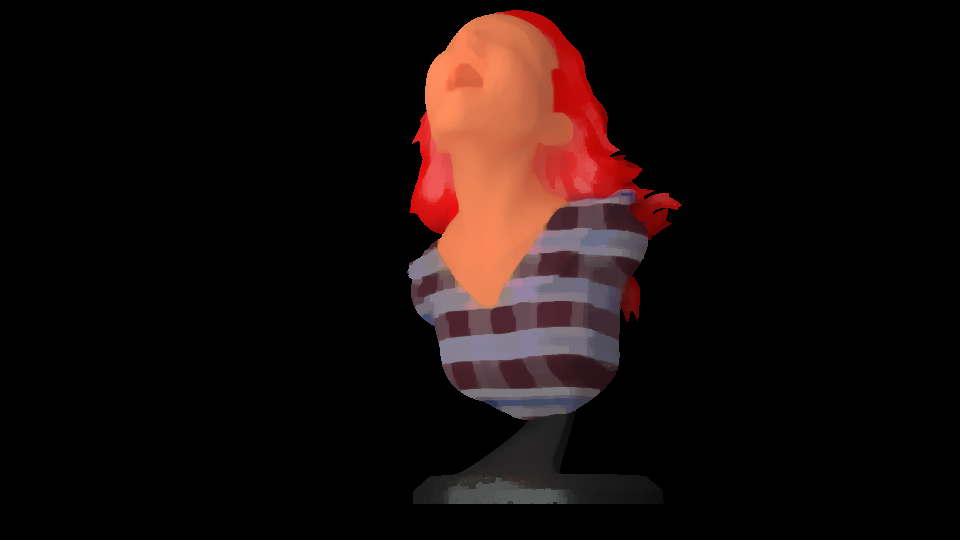} \\
		\qquad Ground truth &
		\includegraphics[width = \linewidth, trim={10cm 2cm 6cm 0},clip]{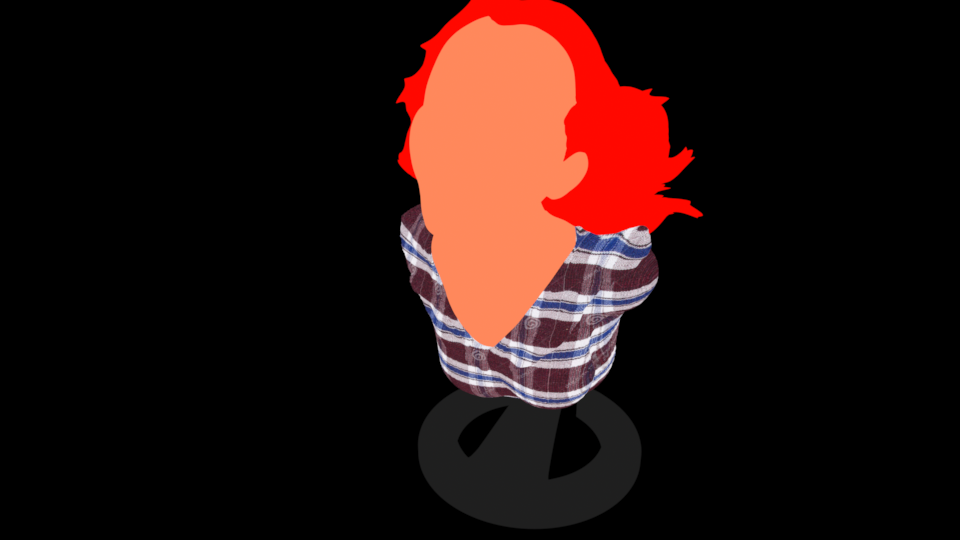} & \,
		\includegraphics[width = \linewidth, trim={8cm 2cm 8cm 0},clip]{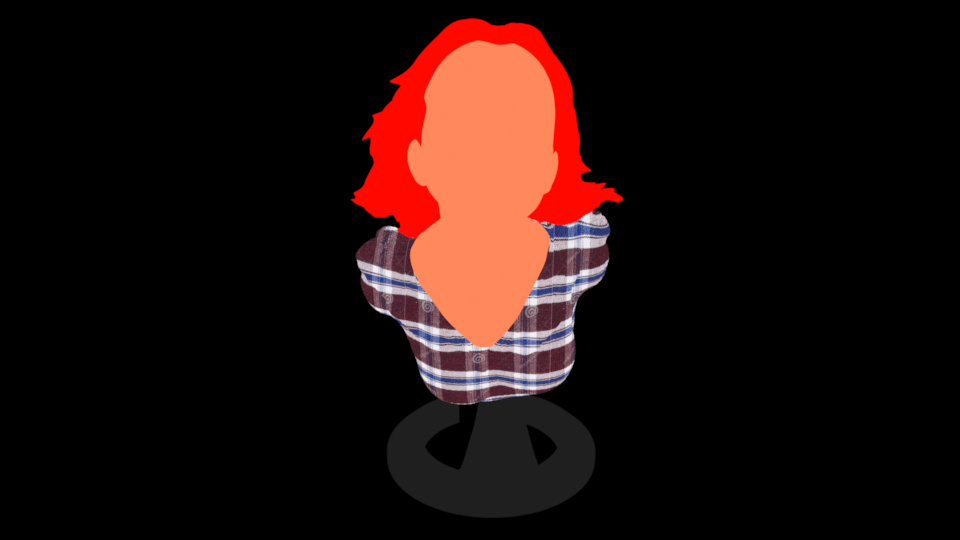} & \,
		\includegraphics[width = \linewidth, trim={10cm 2cm 6cm 0},clip]{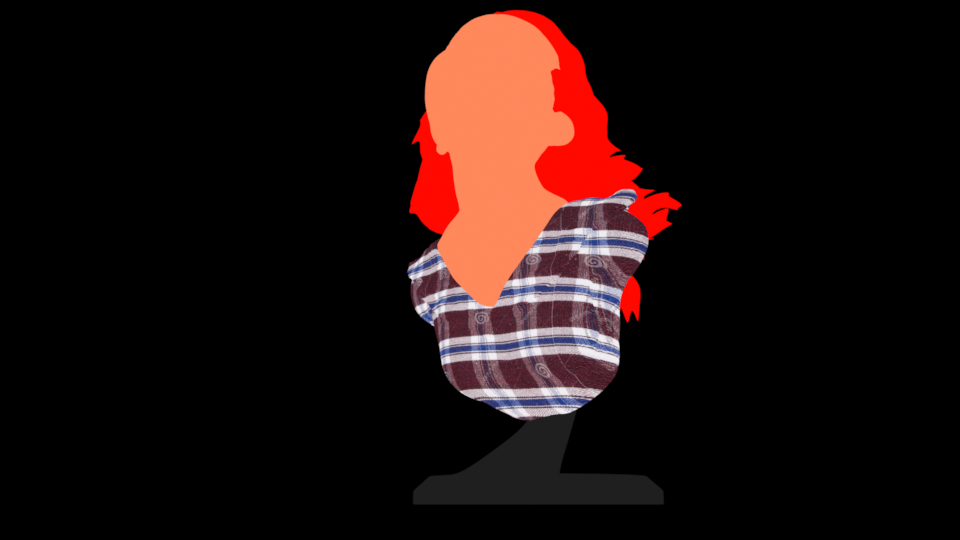}
	\end{tabular}
\end{center}
\caption{First row: three (out of $m=13$) synthetic views of the object of Figure \ref{fig:0}-a, computed with a non-uniform shirt reflectance, a uniform, but partly specular hair reflectance, illuminated by a single extended light source. Second row: estimation of the reflectance using the cartoon + texture decomposition described in \cite{IpolCartoon} (with its parameter fixed to $0.4$). Third row: estimation of the reflectance using the method proposed in \cite{Gehler2011} (with 6 clusters). Forth row: estimation of the reflectance using the proposed approach (with $\lambda = 2.5$ and $\mu = 1000$). Fifth row: ground truth.}
\label{fig:2}
\end{figure*}

\begin{table*}[!ht]
\caption[]{RMSE on the reflectance estimates (the estimated and ground truth reflectance maps are scaled to $[0,1]$), with respect to each channel and to the whole set of images, for our method and two single-view approaches. Our method overcomes the latter on the two considered datasets. See text for details.}
\label{tab:1}
\begin{center}
	\begin{tabular}{ccccc}
		Test & Channel & Cartoon + texture~\cite{IpolCartoon} & Intrinsic decomposition~\cite{Gehler2011} & Ours \\
		\hline
		\begin{tabular}{c} Purely-Lambertian surface \\ + Piecewise-constant reflectance\\ + Skydome lighting \\ (see Figure~\ref{fig:1})\end{tabular} &
		\begin{tabular}{c} R \\ G \\ B \end{tabular} &
		\begin{tabular}{c} 0.62 \\ 0.23 \\ 0.38 \\ \end{tabular} &
		\begin{tabular}{c} 0.26 \\ 0.14 \\ 0.24 \\ \end{tabular} &
		\begin{tabular}{c} \textbf{0.07} \\ \textbf{0.04} \\ \textbf{0.07} \\ \end{tabular} \\
		\hline
		\begin{tabular}{c} Non-uniform shirt reflectance \\ + Partly specular hair reflectance \\ + Single extended light source \\ (see Figure~\ref{fig:2}) \end{tabular} &
		\begin{tabular}{c} R \\ G \\ B \end{tabular} &
		\begin{tabular}{c} 0.60 \\ 0.32 \\ 0.24 \\ \end{tabular} &
		\begin{tabular}{c} 0.29 \\ 0.22 \\ 0.21 \\ \end{tabular} &
		\begin{tabular}{c} \textbf{0.22} \\ \textbf{0.13} \\ \textbf{0.12} \\ \end{tabular} \\
	\end{tabular}
\end{center}
\end{table*}

We first test our reflectance estimation method using $m=13$ synthetic images, of size $540\times960$, of an object whose geometry is perfectly known (see Figure \ref{fig:0}-a). Two scenarios are considered:
\begin{itemize}
	\item[$\bullet$] In Figure~\ref{fig:1}, a purely-Lambertian, piecewise-constant reflectance is mapped onto the surface of the object, which is then illuminated by a ``skydome'' \ie an almost diffuse lighting. Shading effects are thus rather limited, hence applying to each image an estimation method which does not use an explicit reflectance model \eg the cartoon + texture decomposition method from~\cite{IpolCartoon}, should already provide satisfactory results. The reflectance being perfectly piecewise constant, applying sparsity-based intrinsic image decomposition methods such as~\cite{Gehler2011} to each image should also work well.

	\item[$\bullet$] In Figure~\ref{fig:2}, a more complicated (non-uniform) reflectance is mapped onto the shirt, the hair is made partly specular, and the diffuse lighting is replaced by a single extended light source, which induces much stronger shading effects. It will thus be much harder to remove shading without an explicit reflectance model (cartoon + texture approach), while the single-view image decomposition approach should be non-robust to specularities.
\end{itemize}

In both cases, the competing methods~\cite{IpolCartoon} and~\cite{Gehler2011} are applied independently to each of the $m=13$ images. The estimates are thus not expected to be consistent, which may be problematic if the reflectance maps should be further mapped onto the surface for, \eg relighting applications. On the contrary, our approach simultaneously, and consistently, estimates the $m$ reflectance maps.

As we dispose of the reflectance ground truth, we can numerically evaluate these results by estimating the root mean square error (RMSE) for each method, over the whole set of $m=13$ images. The values are presented in Table \ref{tab:1}. In order to compare comparable things, the reflectance estimated by each method is scaled, in each channel, by a factor common to the $m=13$ reflectance maps, so as to minimize the RMSE. This should thus highlight inconsistencies between the reflectance maps.

\begin{figure*}[!ht]
\begin{center}
	\begin{tabular}{ccc}
		\includegraphics[width = 0.31\linewidth, trim={10cm 2cm 6cm 0},clip]{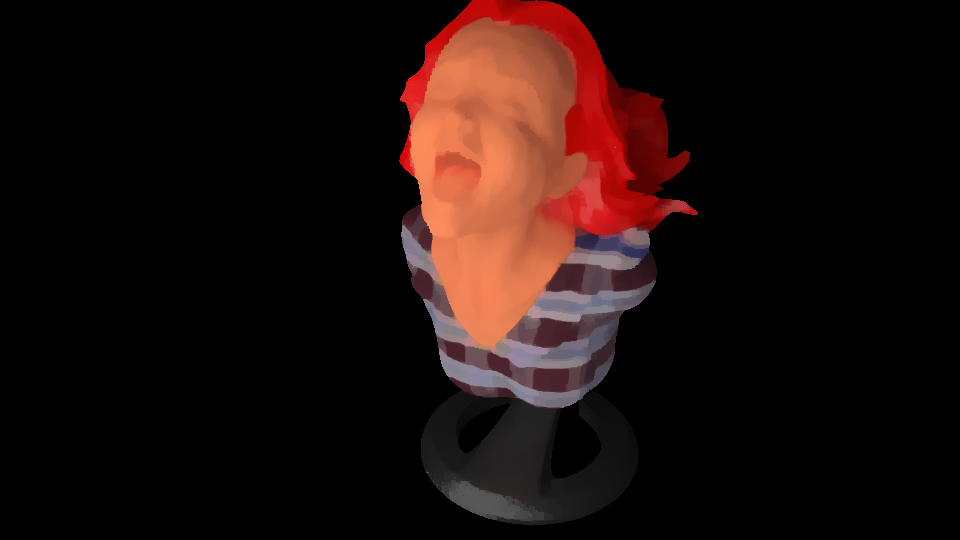} &
		\includegraphics[width = 0.31\linewidth, trim={8cm 2cm 8cm 0},clip]{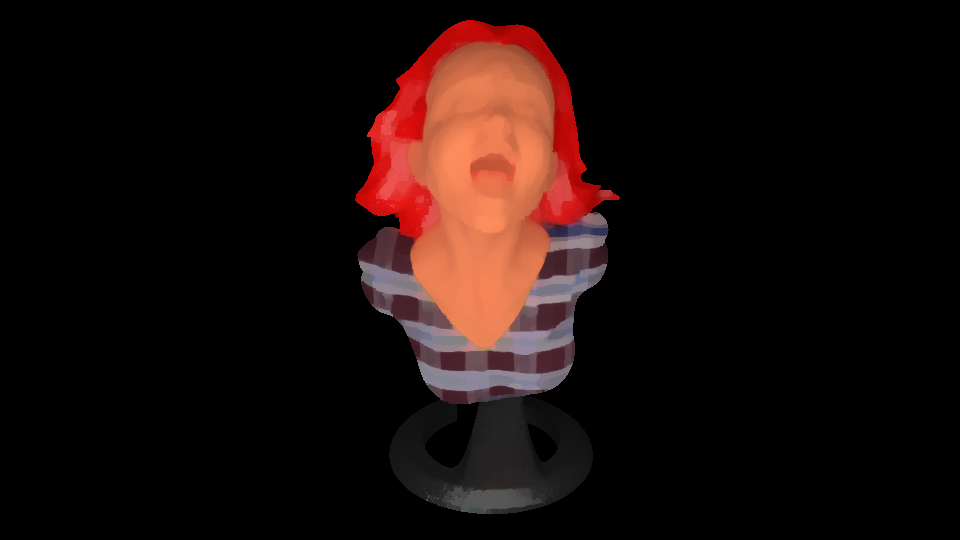} &
		\includegraphics[width = 0.31\linewidth, trim={10cm 2cm 6cm 0},clip]{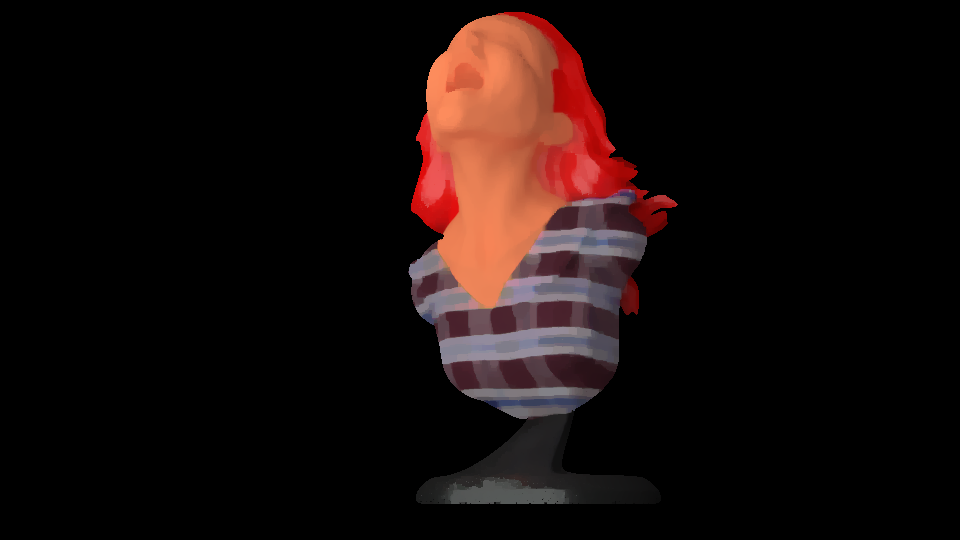}
	\end{tabular}
\end{center}
\caption{Same test as in Figure \ref{fig:2}, using a coarse version of the 3D-shape (see Figures \ref{fig:0}-b and \ref{fig:0}-d), with $\lambda = 2.5$ and $\mu = 1000$. Results are qualitatively similar to those shown in Figure~\ref{fig:2}, obtained with perfect geometry. The RMSE in the RGB channels are, respectively: 0.24, 0.14 and 0.13, which are only slightly higher than those attained with perfect geometry (see Table~\ref{tab:1}).}
\label{fig:5}
\end{figure*}

Based on the qualitative results from Figures~\ref{fig:1} and~\ref{fig:2}, and the quantitative evaluations shown in Table~\ref{tab:1}, we can make the following three observations:

\paragraph{1) Considering an explicit image formation model improves cartoon + texture decomposition.} Actually, the cartoon part from the cartoon + texture decomposition is far less uniform than the reflectance estimated using both other methods. Shading is only blurred, and not really removed. This could be improved by augmenting the regularization weight, but the price to pay would be a loss of detail in the parts containing thinner details (as the shirt, in the example of Figure \ref{fig:2}).

\paragraph{2) Simultaneously estimating the multi-view reflectance maps makes them consistent and improves robustness to specularities.} When estimating each reflectance map individually, inconsistencies arise, which is obvious for the hair in the third line of Figure~\ref{fig:1}, and explains the RMSE values in Table~\ref{tab:1}. In contrast, our results confirm our basic idea \ie that reflectance estimation benefits in two ways from the multi-view framework: this allows us not only to estimate the 3D-shape, but also to constrain the reflectance of each surface point to be the same in all the pictures where it is visible. In addition, since the location of bright spots due to specularity depends on the viewing angle, they usually occur in some places on the surface only under certain viewing angles. Considering multi-view data should thus improve robustness to specularities. This is confirmed in Figure~\ref{fig:2} by the reflectance estimates in the hair, where the specularities are slightly better removed than with single-view methods.

\paragraph{3) A sparsity-based prior for the reflectance should be preferred over total variation.} As we use a TV-smoothing term, which favors piece\-wise-smooth reflectance, the satisfactory results of Figure \ref{fig:1} were predictable. However, some penumbra remains visible around the neck. Since we also know the object geometry, it seems that we could compensate for penumbra. However, this would require that the lighting is known as well, which is not the case in the framework of the targeted usecase, since an outdoors lighting is uncontrolled. Moreover, we would have to consider not only the primary lighting, but also the successive bounces of light on the different parts of the scene (these were taken into account by the ray-tracing algorithm, when synthesizing the images). In contrast, the sparsity-based approach~\cite{Gehler2011} is able to eliminate penumbra rather well, without modeling secundary reflections. It is also able to more appropriately remove shading on the face in the example of Figure~\ref{fig:2}, while not degrading as much as total variation the thin structures of the shirt. Hence, the relative simplicity of the numerical solution, which is a consequence of the choice of replacing the Potts prior by a total variation one (see Section~\ref{sec:simplification}), comes with a price. In future works, it may be important to design a numerical strategy handling the original non-smooth, non-convex problem~\eqref{eq:var_1}.

\subsection{Handling Inaccurate Geometry}

\begin{figure*}[!ht]
\centering
	\begin{tabular}{ccc}
		\includegraphics[width = 0.31\linewidth]{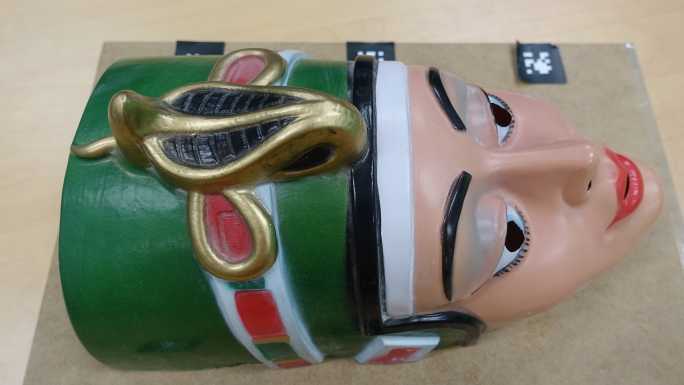} &
		\includegraphics[width = 0.31\linewidth]{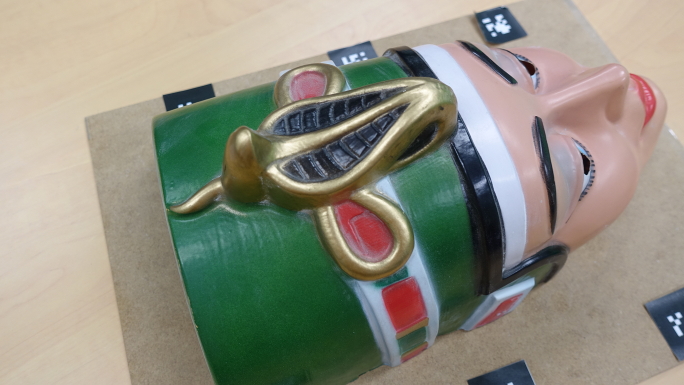} & 
		\includegraphics[width = 0.31\linewidth]{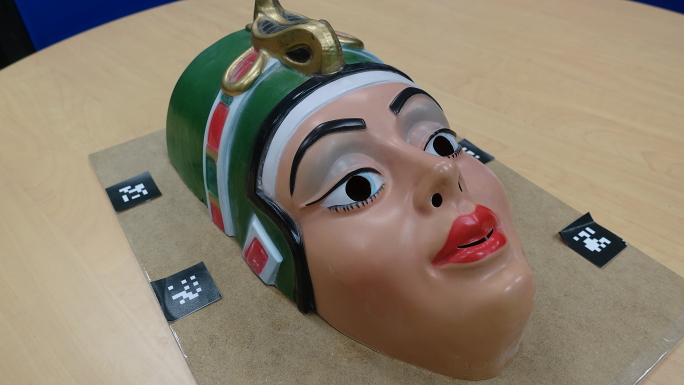} \\ [1em]
		\includegraphics[width = 0.31\linewidth]{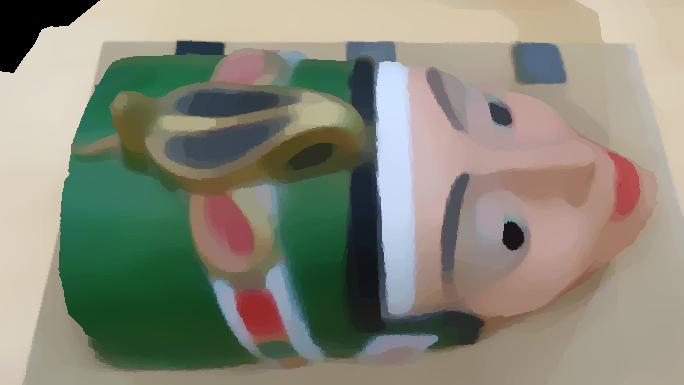} &
		\includegraphics[width = 0.31\linewidth]{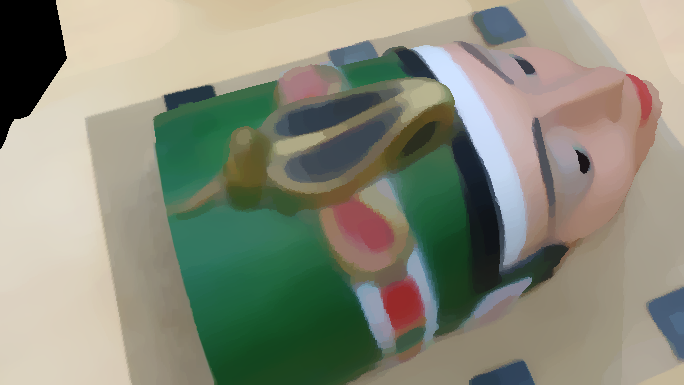} &
		\includegraphics[width = 0.31\linewidth]{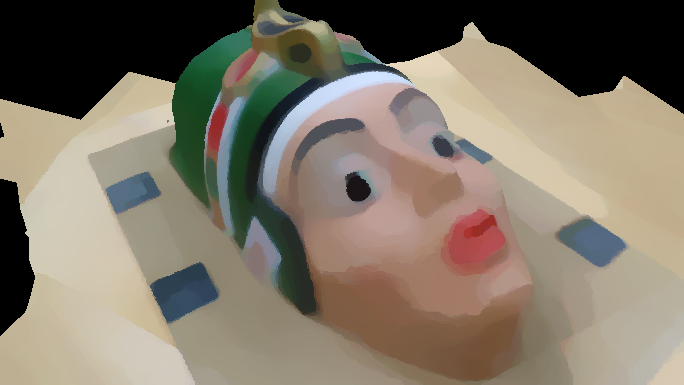}	
	\end{tabular}
\caption{Test on a real-world dataset. First row: three (out of $m=8$) views of the scene. Second row: estimated reflectance maps using the proposed approach (with $\lambda = 2$ and $\mu = 1000$). Geometry and camera parameters were estimated using an SfM/MVS pipeline.}
\label{fig:real_2}
\end{figure*}

In the previous experiments, the geometry was perfectly known. In real-world scenarios, errors in the 3D-shape estimation using SfM and MVS are unavoidable. Therefore, it is necessary to evaluate the ability of our method to handle inaccurate geometry.

Thus, we use for the next experiment the surface shown in Figure \ref{fig:0}-b (zoomed in Figure \ref{fig:0}-d), which is obtained by smoothing the original 3D-shape of Figure \ref{fig:0}-a (zoomed in Figure \ref{fig:0}-c), using a tool from the \verb!meshlab! software. The results provided in Figure \ref{fig:5} show that our method seems robust to such small inaccuracies in the object geometry, and is thus relevant for the intended application.

In Figure~\ref{fig:real_2}, we qualitatively evaluate our method on the outputs of an SfM/MVS pipeline applied to a real-world dataset, which provides estimates of the camera parameters and a rough geometry of the scene. These experiments confirm that small inaccuracies in the geometry input can be handled. The specularities are also appropriately removed, and the reflectance maps present the expected cartoon-like aspect. However, the reflectance is under-estimated in the sides of the nose and around the chin. Indeed, since lighting is fixed, these areas are self-shadowed in all the images. Two workarounds could be used: forcing the regularization term (and, possibly, losing fine-scale details), or actively controlling the lighting in order to be sure that no point on the surface is shadowed in all the views. This is further discussed in the next subsection.

\subsection{Tuning the Hyper-parameters $\lambda$ and $\mu$}

\begin{figure*}[!ht]
\centering
	\begin{tabular}{ccc}
		\includegraphics[width = 0.3\linewidth]{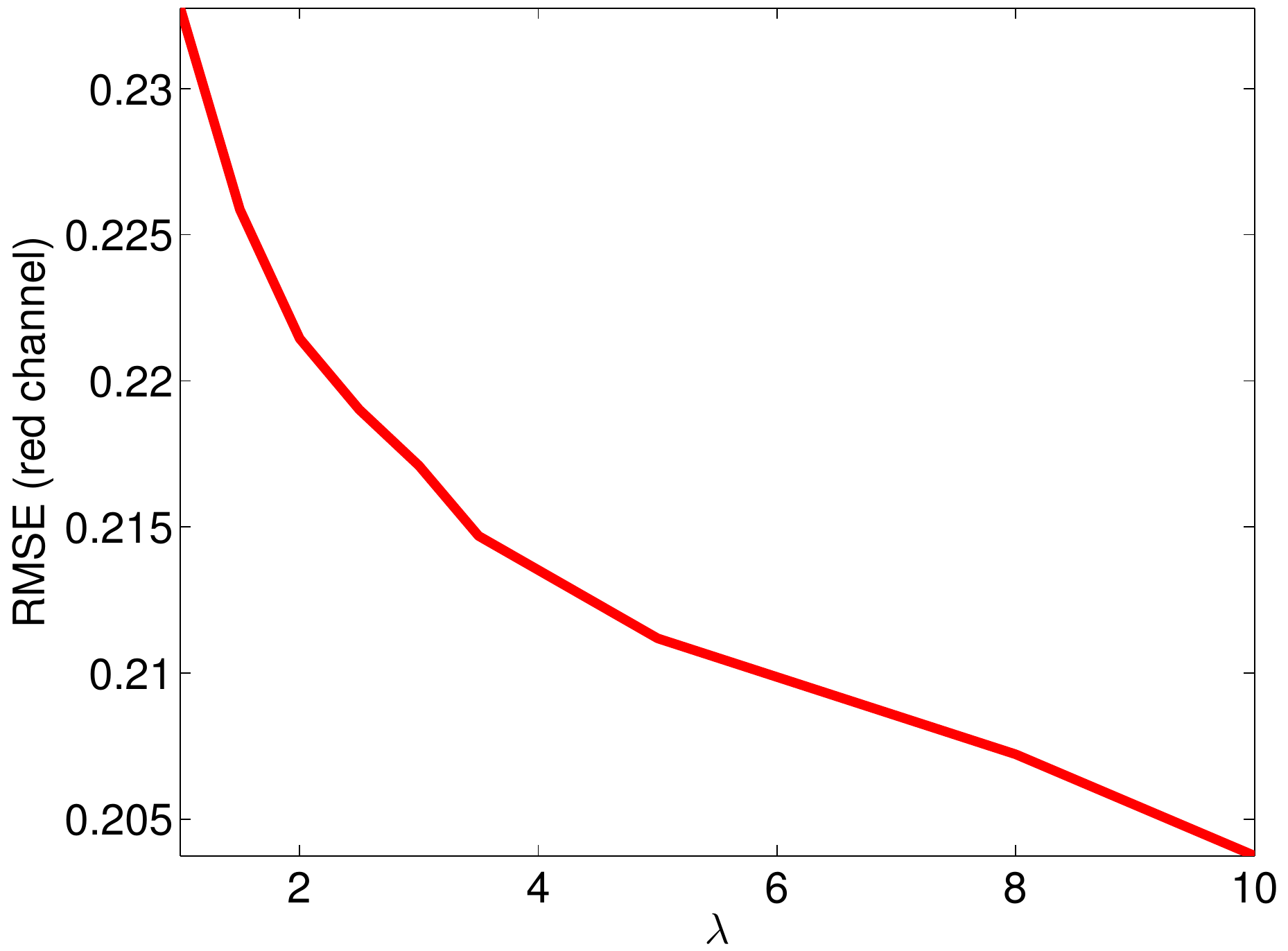} &
		\includegraphics[width = 0.3\linewidth]{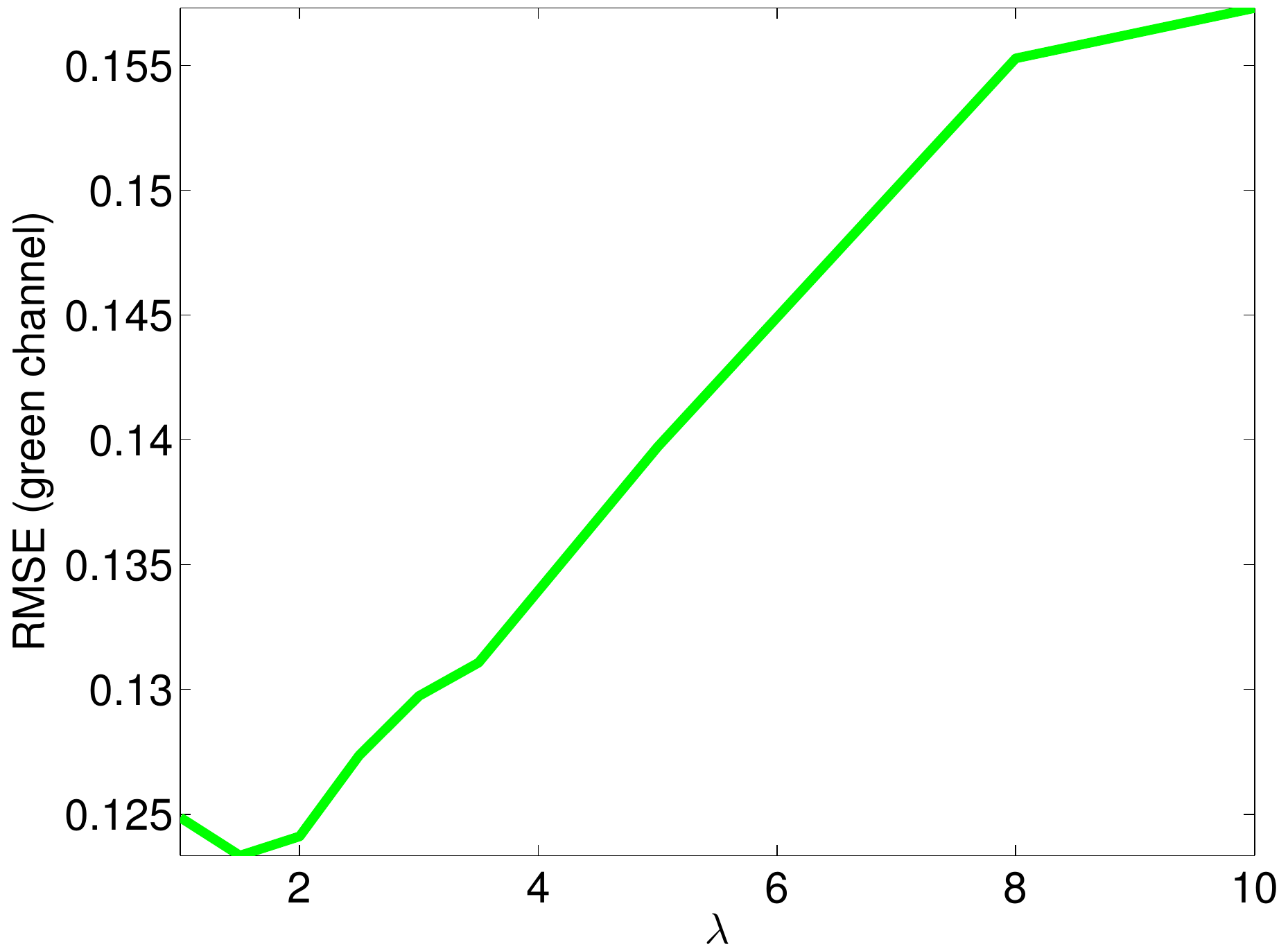} &
		\includegraphics[width = 0.3\linewidth]{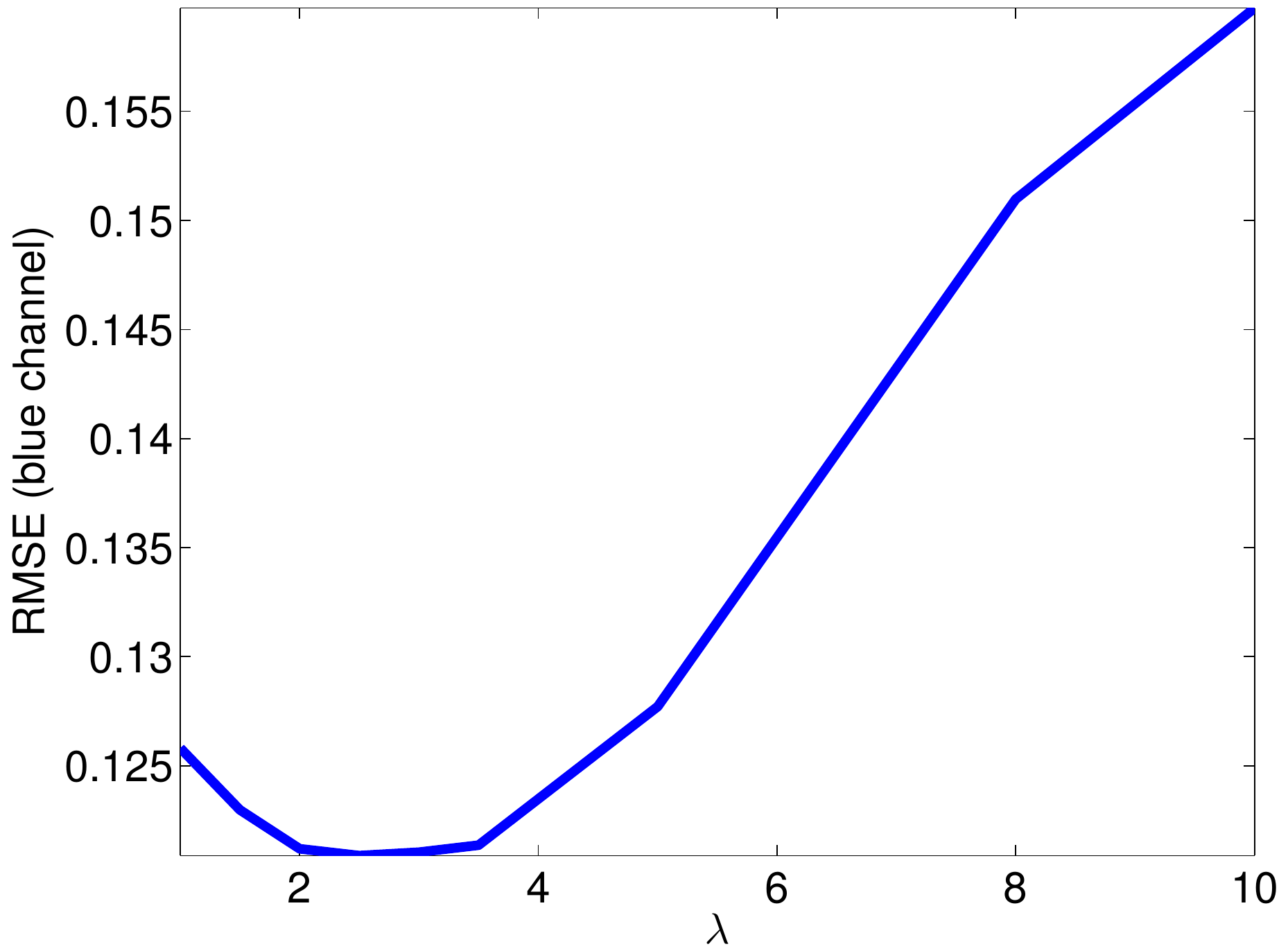}
	\end{tabular}
\caption{Quantitative influence of parameter $\lambda$, using images from the same dataset as that of Figure~\ref{fig:2}, with $\mu = 1000$.}
\label{fig:9}
\end{figure*}

In the previous experiments, we arbitrarily chose the values of parameters $\lambda$ and $\mu$ which provided the ``best'' results. Of course, such a tuning, which may be tedious, must be discussed.

In order to highlight the influence of these parameters, let us first question what would happen without neither regularization nor multi-view consistency \ie when $\lambda = \mu = 0$. In that case, only the photometric term would be optimised, which corresponds to the maximum likelihood case. If lighting is not varying, then we are in a degenerate case which may result in estimating diffuse lighting (see Equation~\eqref{eq:sigma_naive}) and replacing the reflectance maps by the images. Lighting will thus be ``baked in'' the reflectance maps, which is precisely what we pretend not to do.

To avoid this effect, the smoothness term must be activated by setting $\lambda >0$. If we still consider $\mu = 0$, then the variational problem~\eqref{eq:var} comes down to $m$ independent image restoration problems. In fact, these problems are similar to $\ell^1$-TV denoising problems, except that a physically plausible fidelity term is used to help removing the illumination artifacts not only from the total variation regularization, but also by incorporating prior knowledge of the surface geometry. However, because the photometric term is invariant by the transformation $(\rho^i,{\bm \sigma}^i) := (\kappa^i \rho^i, {\bm \sigma}^i/\kappa^i)$, $\kappa^i >0$, each reflectance map $\rho^i$ is estimated only up to a scale factor, hence the $m$ maps will not be consistent, as this is the case for the competing single-view methods.
\newpage

The latter issue is solved by activating the multi-view consistency term \ie by setting $\mu >0$. In that case, there is still an ambiguity $\{\rho^i,{\bm \sigma}^i\}_i := \{\kappa \rho^i, {\bm \sigma}^i/\kappa \}$, $\kappa>0$, but it is now global \ie independent from $i$. To solve this ambiguity, it is enough in practice to set one reflectance value arbitrarily, or to normalize the reflectance values.

Overall, it is necessary to ensure that both $\lambda$ and $\mu$ are strictly positive. The choice of $\mu$ is not really critical. Indeed, the multi-view consistency regularizer which is controlled by $\mu$ arises from relaxing a hard constraint (compare \eqref{eq:var_1} and \eqref{eq:var}). Hence, $\mu$ only needs to be chosen ``high enough'' so that the regularizer approximates fairly well a hard constraint. In all the experiments, we used $\mu = 1000$ and did not face any particular problem. Obviously, if the correspondences were not appropriately computed by SfM, then this value should be reduced, but SfM solutions such as~\cite{Moulon} are now mature enough to provide accurate correspondences.

\begin{figure*}[!ht]
\centering
	\begin{tabular}{m{.06\textwidth} m{.26\textwidth} m{.26\textwidth} m{.26\textwidth}}
		$\lambda = 1$ &\includegraphics[width = \linewidth, trim={10cm 2cm 6cm 0},clip]{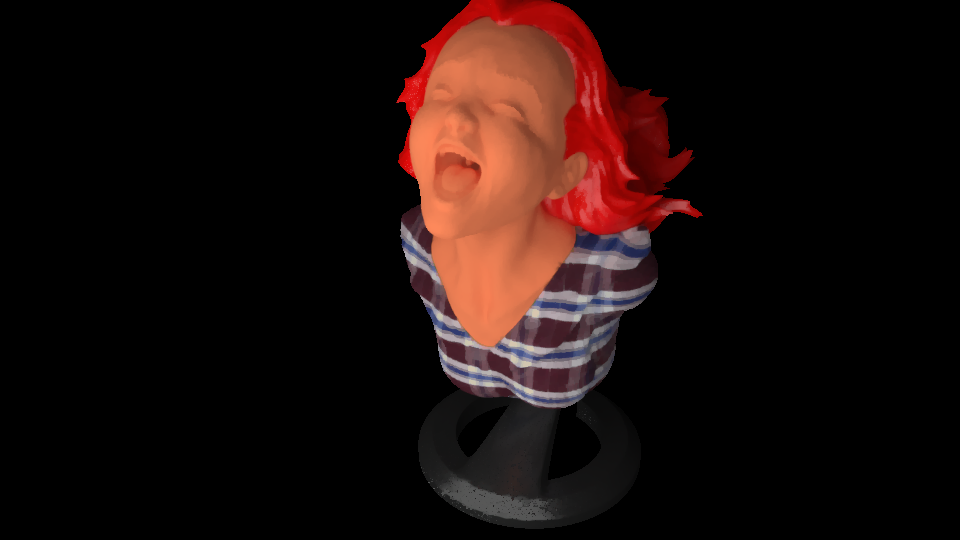} &
		\includegraphics[width = \linewidth, trim={8cm 2cm 8cm 0},clip]{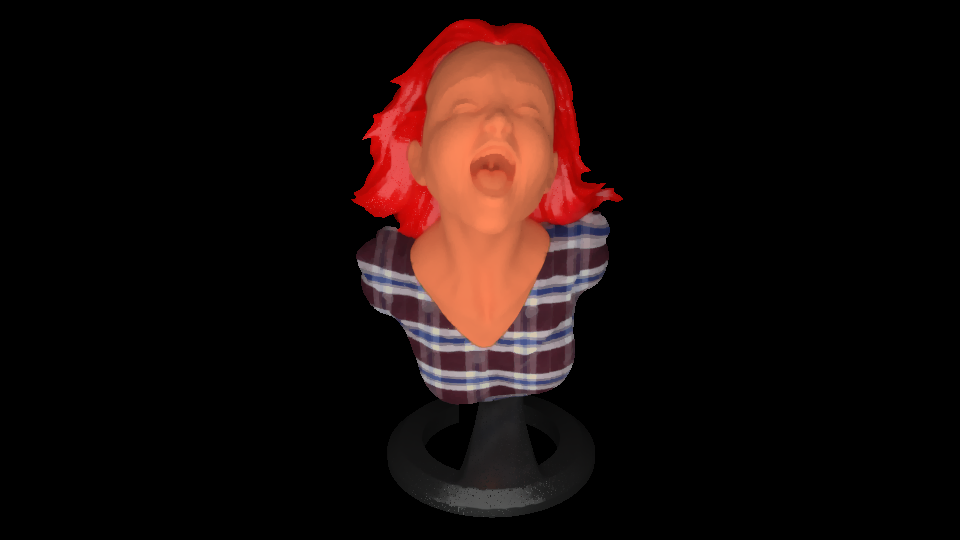} &
		\includegraphics[width = \linewidth, trim={10cm 2cm 6cm 0},clip]{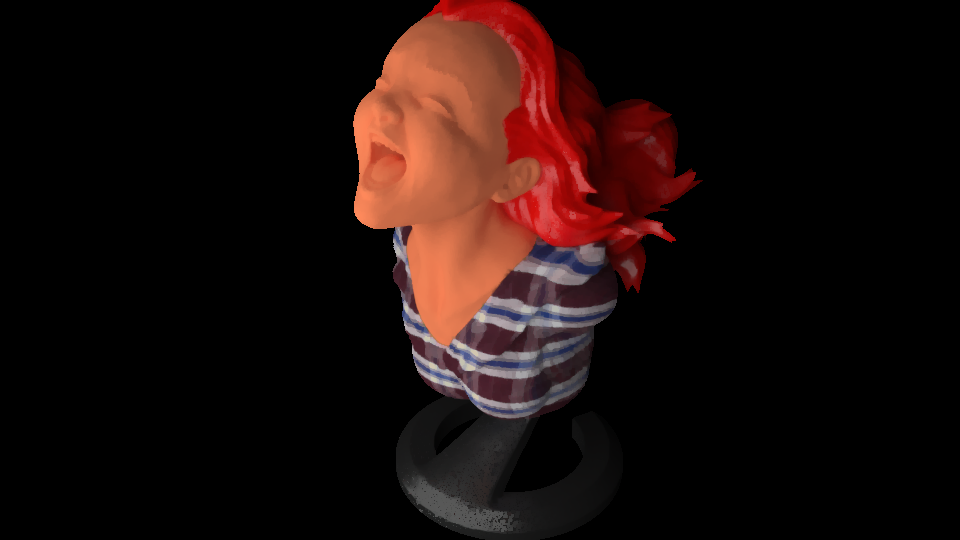} \\[0.2em]
		$\lambda = 2$ & \includegraphics[width = \linewidth, trim={10cm 2cm 6cm 0},clip]{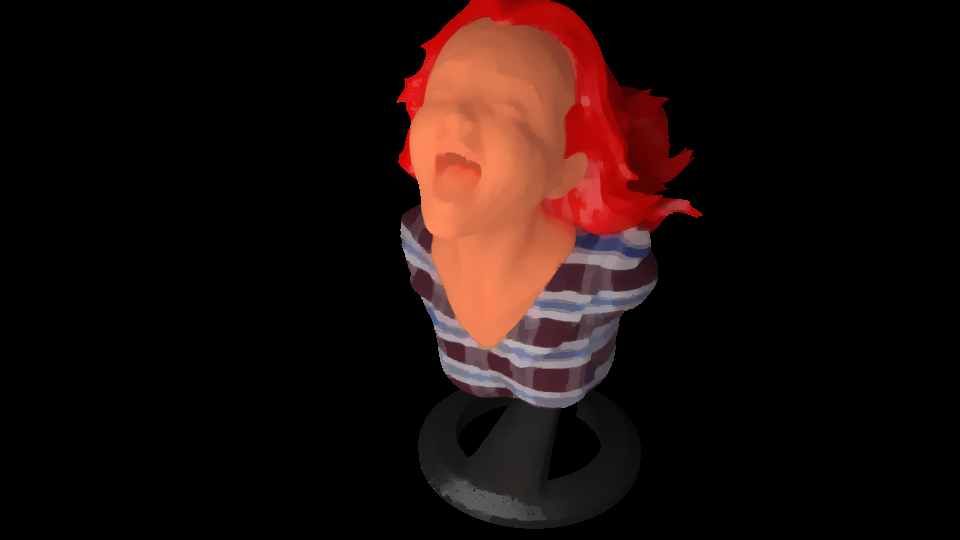} &
		\includegraphics[width = \linewidth, trim={8cm 2cm 8cm 0},clip]{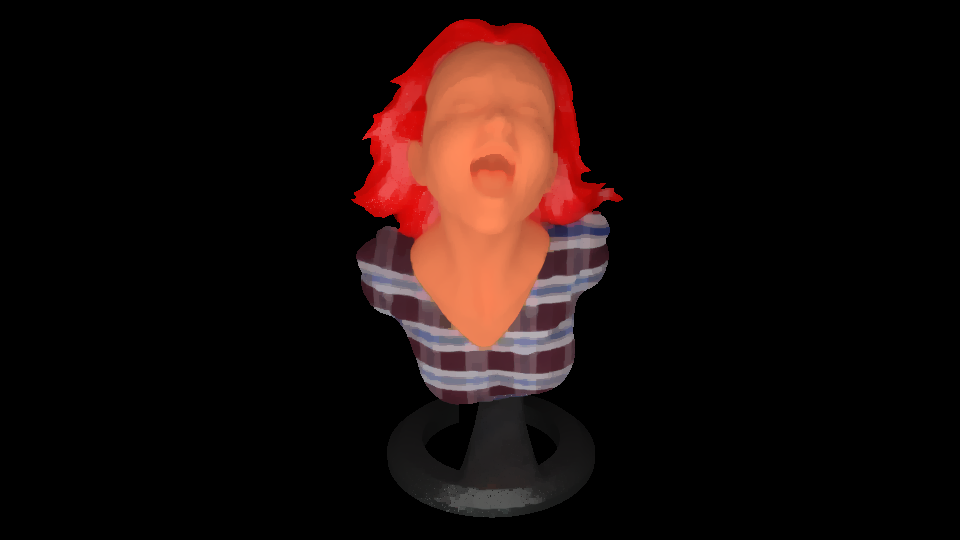} &
		\includegraphics[width = \linewidth, trim={10cm 2cm 6cm 0},clip]{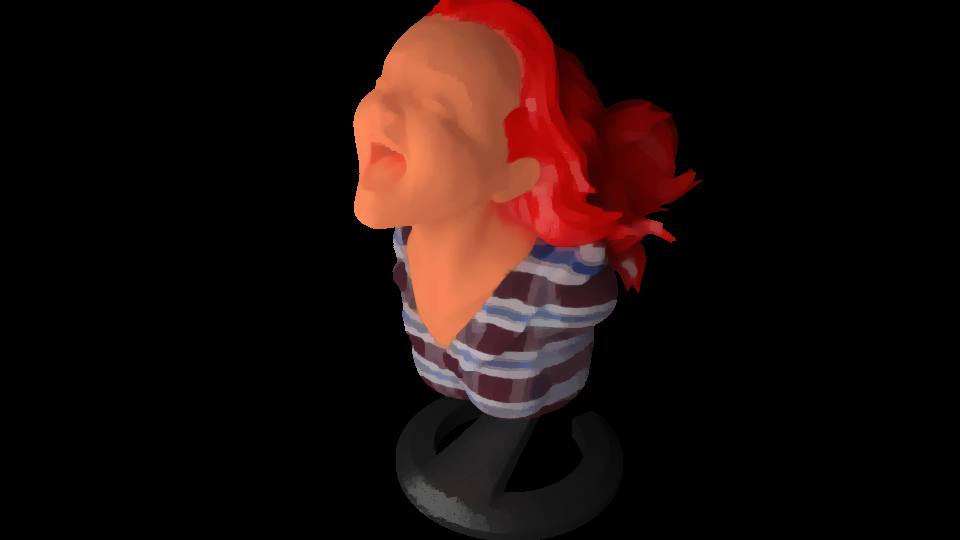} \\[0.2em]
		$\lambda = 3$ & \includegraphics[width = \linewidth, trim={10cm 2cm 6cm 0},clip]{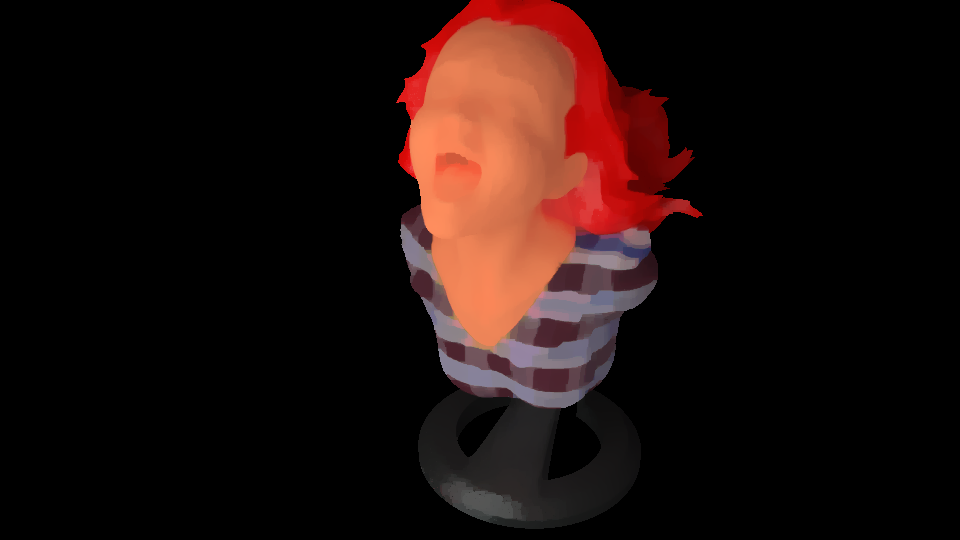} & \includegraphics[width = \linewidth, trim={8cm 2cm 8cm 0},clip]{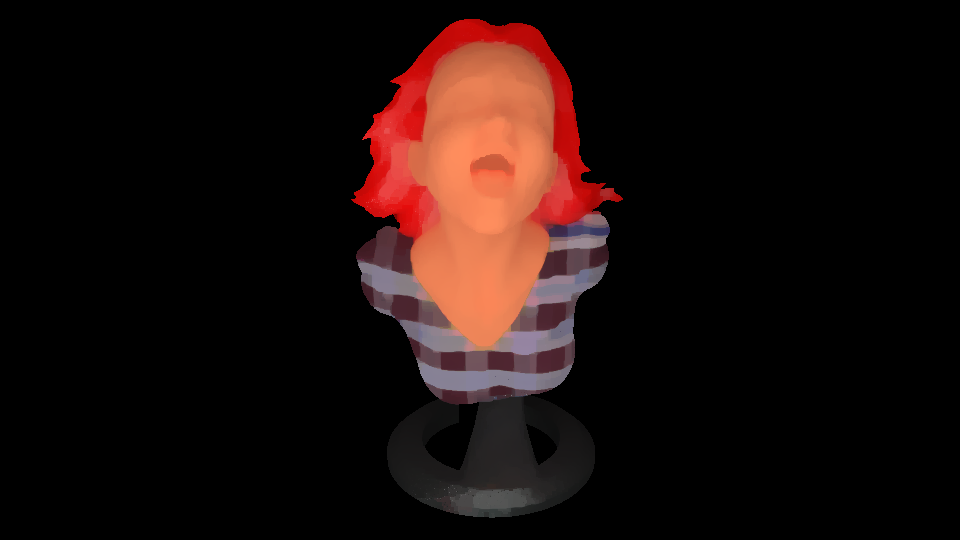} &
		\includegraphics[width = \linewidth, trim={10cm 2cm 6cm 0},clip]{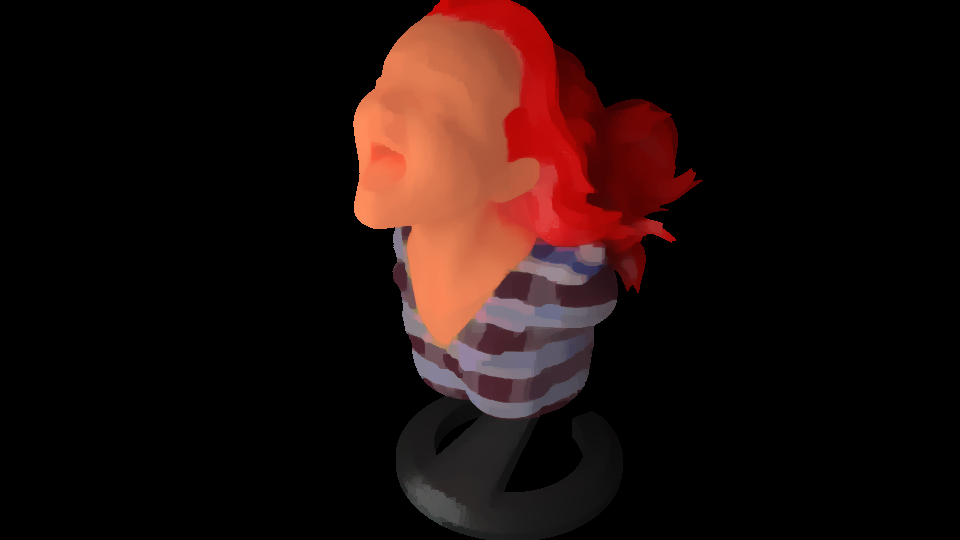} \\[0.2em]
		$\lambda = 5$ & \includegraphics[width = \linewidth, trim={10cm 2cm 6cm 0},clip]{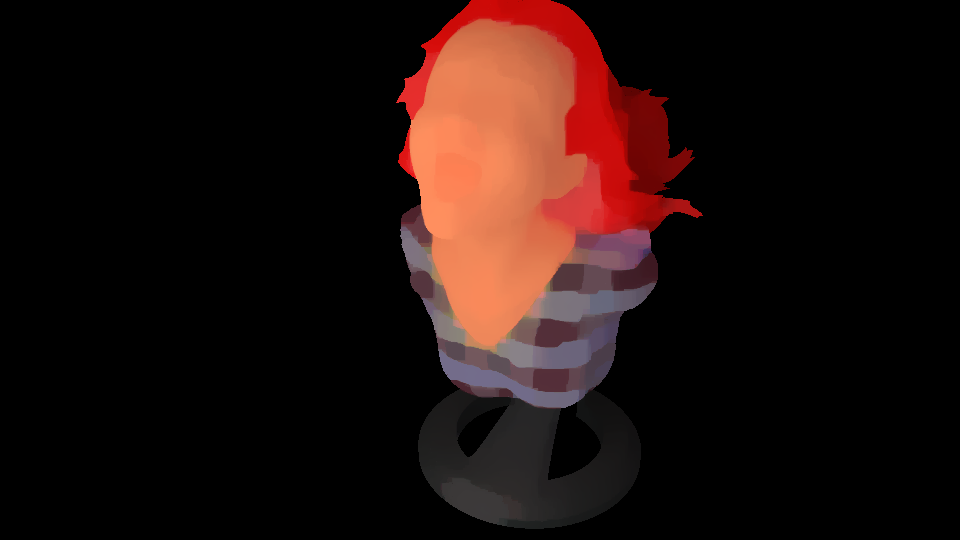} &
		\includegraphics[width = \linewidth, trim={8cm 2cm 8cm 0},clip]{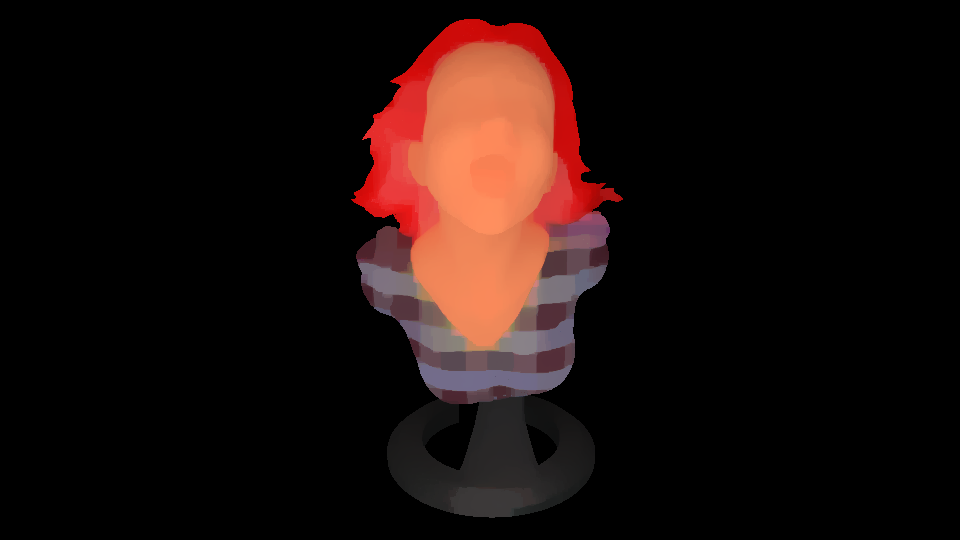} &
		\includegraphics[width = \linewidth, trim={10cm 2cm 6cm 0},clip]{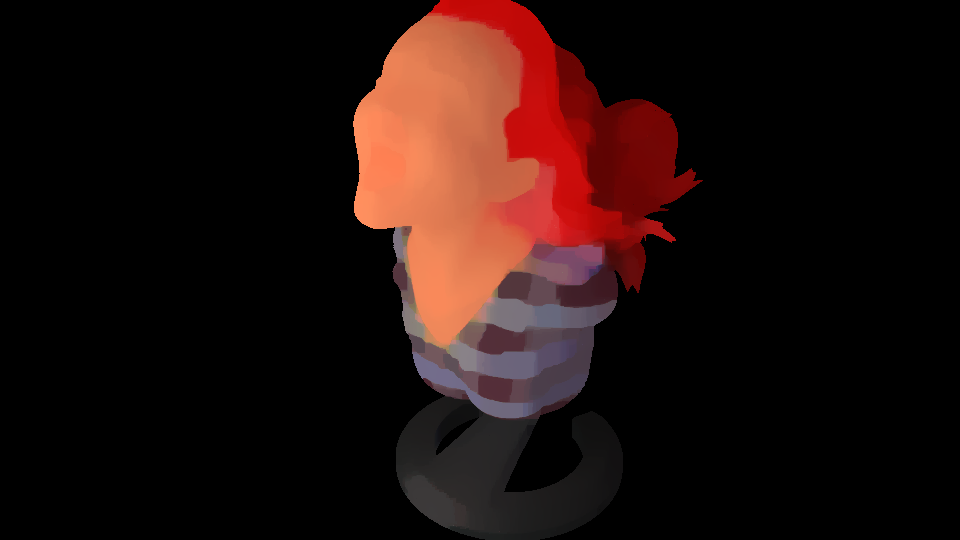}\\[0.2em]
		$\lambda = 10$ & \includegraphics[width = \linewidth, trim={10cm 2cm 6cm 0},clip]{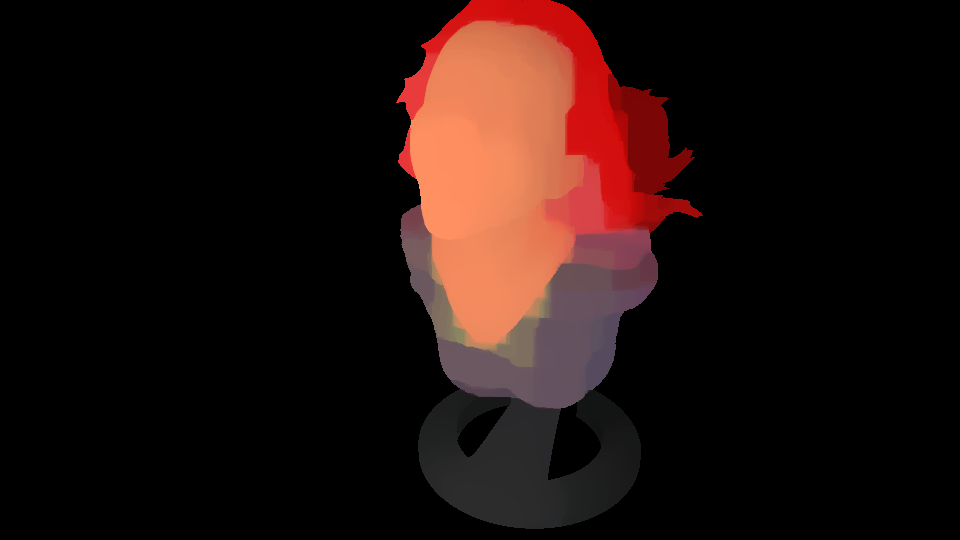} & \includegraphics[width = \linewidth, trim={8cm 2cm 8cm 0},clip]{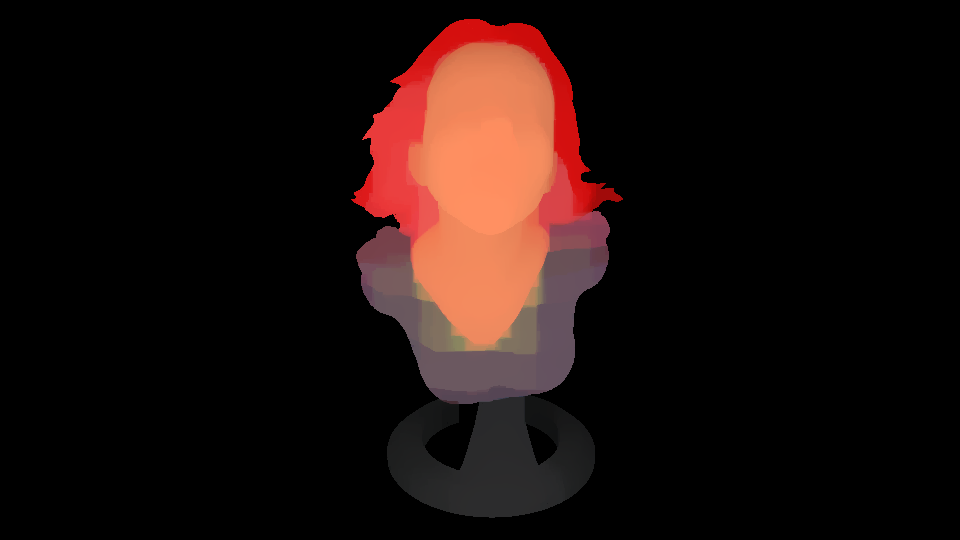} &
		\includegraphics[width = \linewidth, trim={10cm 2cm 6cm 0},clip]{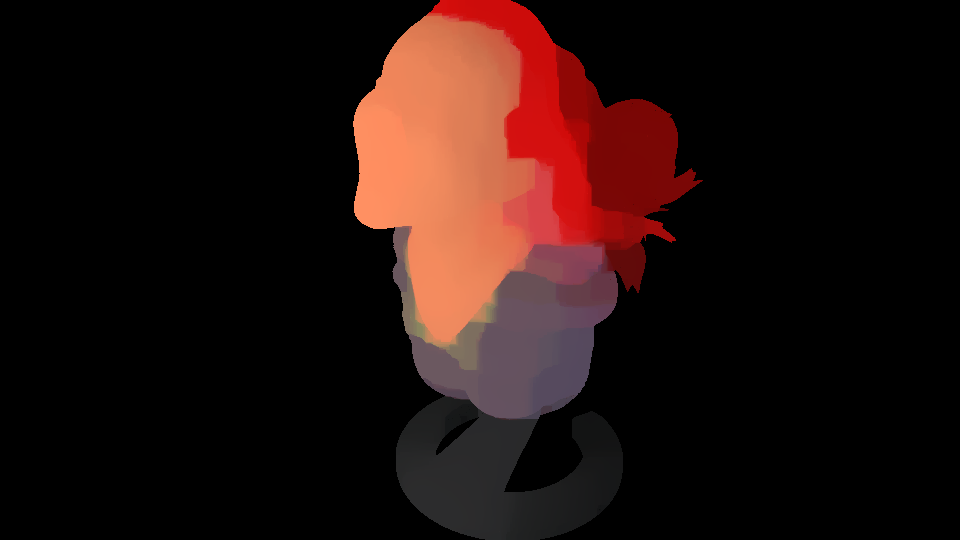}
	\end{tabular}
\caption{Qualitative influence of parameter $\lambda$, using images from the same dataset as that of Figure~\ref{fig:2}, with $\mu = 1000$.}
\label{fig:7}
\end{figure*}

The choice of $\lambda$ is much more critical. This is illustrated in Figure~\ref{fig:9}, which shows the RMSE in each channel, using images from the same dataset as that of Figure \ref{fig:2}, at convergence of our algorithm, as a function of $\lambda$. This graph shows that the ``optimal'' value of $\lambda$ is very hard to find: in this example, a high value of $\lambda$ would diminish the RMSE in the face and the hair (which are mostly red), because this would make them uniform as expected (see Figure~\ref{fig:7}, last rows). However, a much lower value of $\lambda$ is required in order to preserve the thin shirt details, which mostly contain green and blue components (see Figure~\ref{fig:7}, first rows).

\begin{figure*}[!ht]
\centering
	\begin{tabular}{ccc}
		\includegraphics[width = 0.28\linewidth, trim={10cm 2cm 6cm 0},clip]{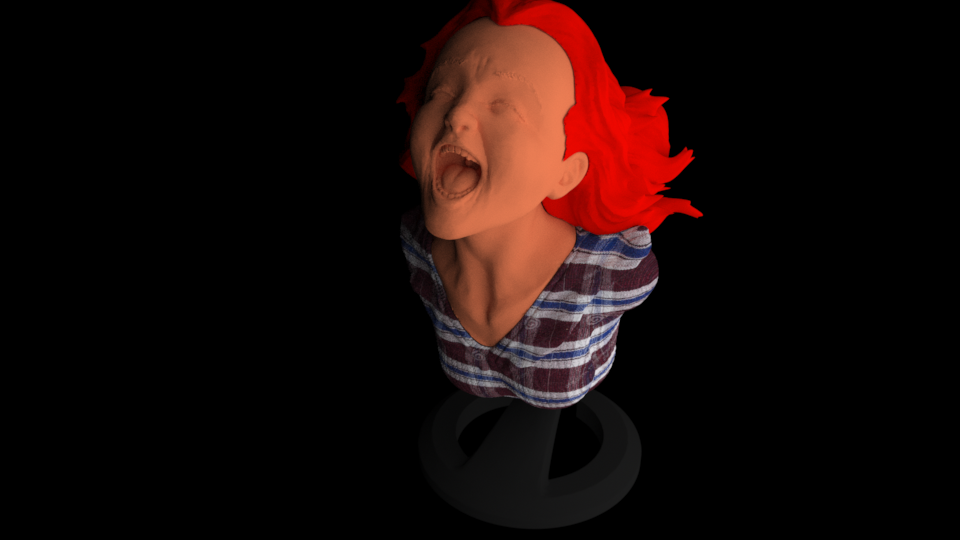} &
		\includegraphics[width = 0.28\linewidth, trim={8cm 2cm 8cm 0},clip]{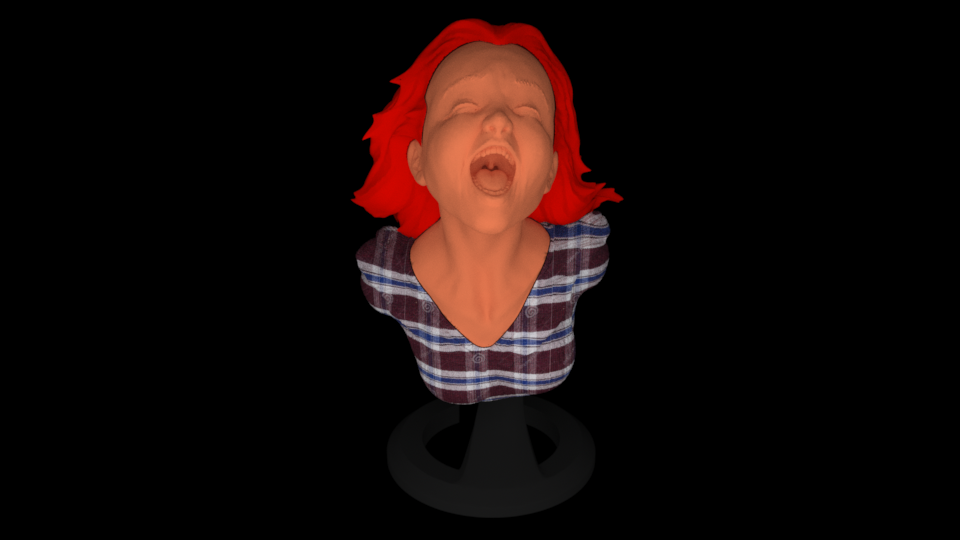} &
		\includegraphics[width = 0.28\linewidth, trim={10cm 2cm 6cm 0},clip]{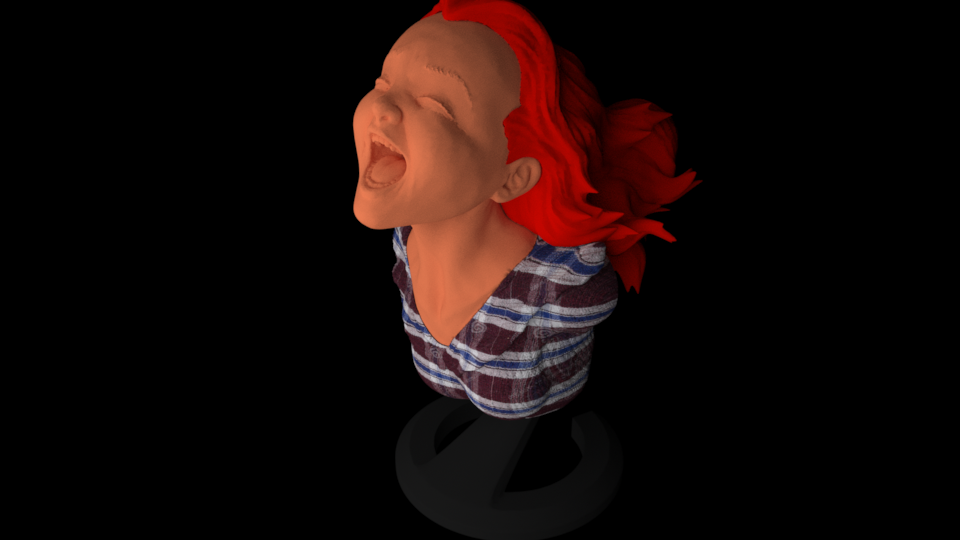} \\[0.2em]
		\includegraphics[width = 0.28\linewidth, trim={10cm 2cm 6cm 0},clip]{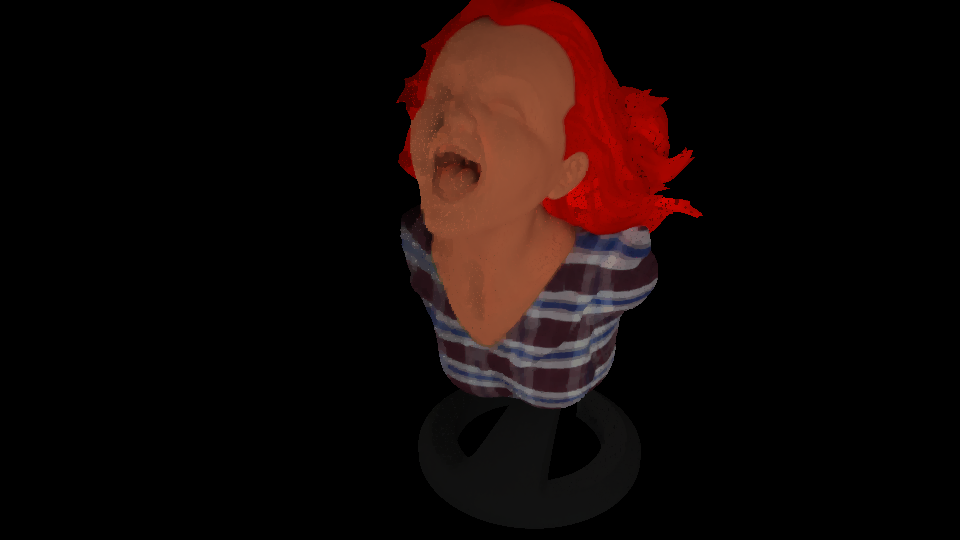} &
		\includegraphics[width = 0.28\linewidth, trim={8cm 2cm 8cm 0},clip]{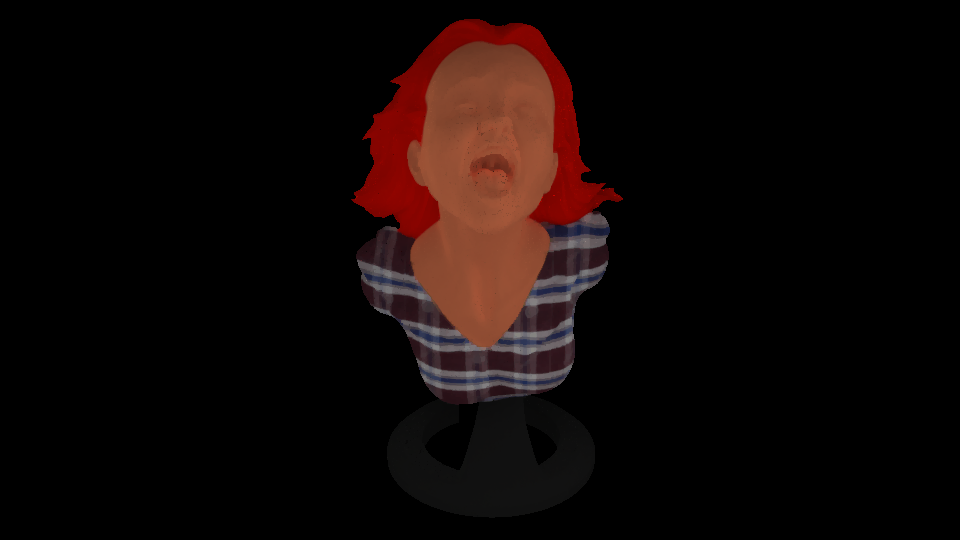} &
		\includegraphics[width = 0.28\linewidth, trim={10cm 2cm 6cm 0},clip]{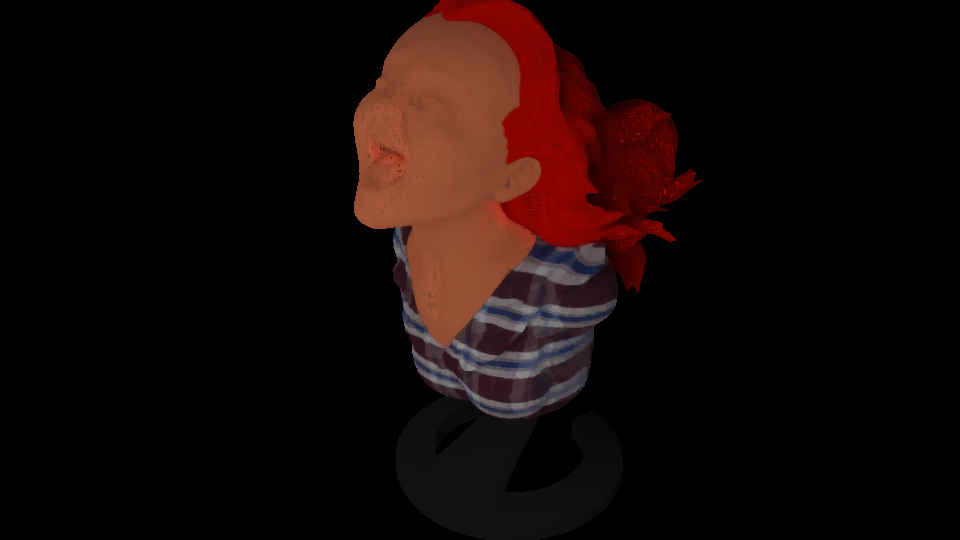}
	\end{tabular}
\caption{First row: three (out of $m=13$) synthetic images computed under varying lighting (which comes here from the right, from the front and from the left, respectively). Second row: estimated reflectance maps using the proposed approach (with $\lambda = 1$ and $\mu = 1000$). The thin structures of the shirt are preserved, while shading on the face is largely reduced. These results must be compared with those of the first row in Figure~\ref{fig:7}, obtained with the same value of $\lambda$ but under fixed lighting.}
\label{fig:8}
\end{figure*}

There is one situation where this tuning is much easier. It is when the lighting is not fixed, but strongly varying. As discussed in Section~\ref{sec:model}, the problem of jointly estimating reflectance and lighting is then over-deter\-mi\-ned, which theoretically makes the regularization unnecessary. In Figure~\ref{fig:8}, we show the results obtained in the case where each image is obtained under a different lighting. In that case, the thin structures of the shirt are preserved, while shading on the face is largely reduced, despite the choice of a very low regularization weight $\lambda = 1$. Note that we cannot use the limit case $\lambda =0$ because not all pixels have correspondences in all images: there may thus be a few pixels for which the problem remains under-determined, and for which diffusion is required. Overall, this experiment shows that, without any prior knowledge on the lighting, the only way to avoid introducing an empirical prior on the reflectance, and thus its tuning, is to actively control lighting during the acquisition process. This means, combining multi-view and photometric stereo.

It happens that this problem is actively being addressed by the computer vision community~\cite{Park2017}. Interestingly, in this research the focus is put on highly accurate geometry estimation, and not so much on reflectance estimation (no reflectance estimation result is shown). Therefore, it may be an interesting future research direction to incorporate our reflectance estimation framework in such multi-view, multi-lighting approaches. Both highly accurate geometry and reflectance could indeed be expected.

\section{Conclusion and Perspectives}
\label{sec:conclusion}

We have proposed a variational framework for estimating the reflectance of a scene from a series of multi-view images. We advocate a 2D-parameterization of reflectance, turning the problem into that of converting the input images into reflectance maps. Invoking a Bayesian rationale leads to a variational model comprising a $\ell^1$-norm-based photometric data term, a Potts regularizer and a multi-view consistency constraint. For simplicity, both the latter are relaxed into a total variation term and a $\ell^1$-norm term, respectively. Numerical solving is carried out using an alternating majo\-ri\-za\-tion-minimization algorithm. Empirical results on both synthetic and real-world datasets demonstrate the interest of considering multi-view images for reflectance estimation, as it allows to benefit from prior knowledge of the geometry, to improve robustness to specularities and to guarantee consistency of the reflectance estimates.

However, the critical analysis of our results also highlighted some limitations and possible future research directions. For instance, avoiding the relaxation of the non-smooth, non-convex regularization, seems to be necessary in order to really ensure that the estimated reflectance maps are piecewise-constant. In addition, the choice of parameterizing reflectance in the image (2D) domain is advocated for reasons of numerical simplicity, yet it seems somewhat more natural to work directly on the surface (this would avoid the multi-view consistency constraint). However, this would require turning our simple variational framework into a more arduous optimization problem over a manifold.

Finally, we could disambiguate the problem by measuring upstream the incoming light, using, for instance, environment maps. Without prior measurement, it seems that the only way to avoid resorting to an arbitrary prior for limiting the arising ambiguities consists in actively controlling the lighting (this would avoid resorting to spatial regularization). Therefore, another extension of our work consists in estimating reflectance from multi-view, multi-lighting data, in the spririt of multi-view photometric stereo techniques. However, this would require appropriately modifying the SfM/MVS pipeline, which relies on the constant brightness assumption.

\begin{acknowledgements}

Yvain \textsc{Qu\'eau} and Daniel \textsc{Cremers} were supported by the ERC Consolidator Grant ``3D Reloaded''.

\end{acknowledgements}

\bibliographystyle{splncs03} 
\bibliography{biblio}

\begin{thebibliography}{10}
\providecommand{\url}[1]{\texttt{#1}}
\providecommand{\urlprefix}{URL }

\bibitem{Adelson}
Adelson, E.H., Pentland, A.P.: Perception as Bayesian inference, chap. {The
  perception of shading and reflectance}, pp. 409--423. Cambridge University
  Press (1996)

\bibitem{RomeInADay}
Agarwal, S., Snavely, N., Simon, I., Seitz, S.M., Szeliski, R.: {Building Rome
  in a Day}. In: Proceedings of the IEEE International Conference on Computer
  Vision. pp. 72--79 (2009)

\bibitem{Aujol2006}
Aujol, J.F., Gilboa, G., Chan, T., Osher, S.: {Structure-Texture Image
  Decomposition -- Modeling, Algorithms, and Parameter Selection}.
  International Journal of Computer Vision  67(1),  111--136 (2006)

\bibitem{Barron}
Barron, J., Malik, J.: Shape, illumination, and reflectance from shading. IEEE
  Transactions on Pattern Analysis and Machine Intelligence  37(8),  1670--1687
  (2015)

\bibitem{Basri2007}
Basri, R., Jacobs, D., Kemelmacher, I.: {Photometric Stereo with General,
  Unknown Lighting}. International Journal of Computer Vision  72(3),  239--257
  (2007)

\bibitem{Basri}
Basri, R., Jacobs, D.P.: {Lambertian reflectances and linear subspaces}. IEEE
  Transactions on Pattern Analysis and Machine Intelligence  25(2),  218--233
  (2003)

\bibitem{bell14intrinsic}
Bell, S., Bala, K., Snavely, N.: Intrinsic images in the wild. ACM Transactions
  on Graphics  33(4),  159:1--159:12 (2014)

\bibitem{BPD09}
Bousseau, A., Paris, S., Durand, F.: User assisted intrinsic images. ACM
  Transactions on Graphics  28(5),  130:1--130:10 (2009)

\bibitem{Chen2013}
Chen, Q., Koltun, V.: A simple model for intrinsic image decomposition with
  depth cues. In: Proceedings of the IEEE International Conference on Computer
  Vision. pp. 241--248 (2013)

\bibitem{Cho2016}
Cho, D., Matsushita, Y., Tai, Y.W., Kweon, I.S.: Photometric stereo under
  non-uniform light intensities and exposures. In: Proceedings of the European
  Conference on Computer Vision. pp. 170--186 (2016)

\bibitem{Frolova}
Frolova, D., Simakov, D., Basri, R.: {Accuracy of spherical harmonic
  approximations for images of Lambertian objects under far and near lighting}.
  In: Proceedings of the European Conference on Computer Vision. pp. 574--587
  (2004)

\bibitem{MVS}
Furukawa, Y., Hern{\'a}ndez, C., et~al.: Multi-view stereo: A tutorial.
  Foundations and Trends in Computer Graphics and Vision  9(1-2),  1--148
  (2015)

\bibitem{GMLG12}
Garces, E., Munoz, A., Lopez-Moreno, J., Gutierrez, D.: {Intrinsic Images by
  Clustering}. Computer Graphics Forum  31(4),  1415--1424 (2012)

\bibitem{Gehler2011}
Gehler, P., Rother, C., Kiefel, M., Zhang, L., Sch\"olkopf, B.: {Recovering
  Intrinsic Images with a Global Sparsity Prior on Reflectance}. In: Advances
  in Neural Information Processing Systems. pp. 765--773 (2011)

\bibitem{GolubV4}
Golub, G.H., Van~Loan, C.F.: Matrix Computations (4th Ed.). Johns Hopkins
  University Press (2013)

\bibitem{Horn}
Horn, B.K.P.: {Shape From Shading: A Method for Obtaining the Shape of a Smooth
  Opaque Object From One View}. Ph.D. thesis, Department of Electrical
  Engineering and Computer Science, Massachusetts Institute of Technology
  (1970)

\bibitem{Jin-et-al-08}
Jin, H., Cremers, D., Wang, D., Yezzi, A., Prados, E., Soatto, S.: {3-D
  Reconstruction of Shaded Objects from Multiple Images Under Unknown
  Illumination}. International Journal of Computer Vision  76(3),  245--256
  (2008)

\bibitem{Kim}
Kim, K., Torii, A., Okutomi, M.: {Multi-view Inverse Rendering Under Arbitrary
  Illumination and Albedo}. In: Proceedings of the European Conference on
  Computer Vision. pp. 750--767 (2016)

\bibitem{Laffont2013}
Laffont, P.Y., Bousseau, A., Drettakis, G.: {Rich Intrinsic Image Decomposition
  of Outdoor Scenes from Multiple Views}. IEEE Transactions on Visualization
  and Computer Graphics  19(2),  210--224 (2013)

\bibitem{Laffont2012}
Laffont, P.Y., Bousseau, A., Paris, S., Durand, F., Drettakis, G.: Coherent
  intrinsic images from photo collections. ACM Transactions on Graphics  31,
  202:1--202:11 (2012)

\bibitem{Land1971}
Land, E.H., McCann, J.J.: {Lightness and retinex theory}. Journal of the
  Optical Society of America  61,  1--11 (1971)

\bibitem{Langguth}
Langguth, F., Sunkavalli, K., Hadap, S., Goesele, M.: {Shading-aware Multi-view
  Stereo}. In: Proceedings of the European Conference on Computer Vision. pp.
  469--485 (2016)

\bibitem{IpolCartoon}
Le~Guen, V.: {Cartoon + Texture Image Decomposition by the TV-L1 Model}. Image
  Processing On Line  4,  204--219 (2014)

\bibitem{Robert}
Maier, R., Kim, K., Cremers, D., Kautz, J., Nie{\ss}ner, M.: {Intrinsic3D:
  High-Quality 3D Reconstruction by Joint Appearance and Geometry Optimization
  with Spatially-Varying Lighting}. In: Proceedings of the IEEE International
  Conference on Computer Vision (2017)

\bibitem{Maurer}
Maurer, D., Ju, Y.C., Breu\ss, M., Bruhn, A.: {Combining Shape from Shading and
  Stereo: A Variational Approach for the Joint Estimation of Depth,
  Illumination and Albedo}. In: Proceedings of the British Machine Vision
  Conference (2016)

\bibitem{SSVM_2017_Jean}
M\'elou, J., Qu\'eau, Y., Durou, J.D., Castan, F., Cremers, D.: {Beyond
  Multi-view Stereo: Shading-Reflectance Decomposition}. In: Proceedings of the
  International Conference on Scale Space and Variational Methods in Computer
  Vision. pp. 694--705 (2017)

\bibitem{Moulon}
Moulon, P., Monasse, P., Marlet, R.: {openMVG: An open multiple view geometry
  library}. \url{https://github.com/openMVG/openMVG} (2014)

\bibitem{Mumford1994}
Mumford, D.: {Bayesian rationale for the variational formulation}. In:
  Geometry-Driven Diffusion in Computer Vision, pp. 135--146. Springer (1994)

\bibitem{Nadian-Ghomsheh2016}
Nadian-Ghomsheh, A., Hassanian, Y., Navi, K.: {Intrinsic Image Decomposition
  via Structure-Preserving Image Smoothing and Material Recognition}. PLoS ONE
  11(12),  1--22 (2016)

\bibitem{Park2017}
Park, J., Sinha, S.N., Matsushita, Y., Tai, Y.W., Kweon, I.S.: Robust multiview
  photometric stereo using planar mesh parameterization. IEEE Transactions on
  Pattern Analysis and Machine Intelligence  39(8),  1591--1604 (2017)

\bibitem{JMIV2017_LEDS}
Qu{\'e}au, Y., Durix, B., Wu, T., Cremers, D., Lauze, F., Durou, J.D.:
  {LED-based Photometric Stereo: Modeling, Calibration and Numerical Solution}.
  Journal of Mathematical Imaging and Vision  (2018), {(to appear)}

\bibitem{Queau2017bis}
Qu\'eau, Y., Pizenberg, M., Durou, J.D., Cremers, D.: Microgeometry capture and
  {RGB} albedo estimation by photometric stereo without demosaicing. In:
  Proceedings of the International Conference on Quality Control by Artificial
  Vision (2017)

\bibitem{Ramamoorthi}
Ramamoorthi, R., Hanrahan, P.: {An Efficient Representation for Irradiance
  Environment Maps}. In: Proceedings of the Annual Conference on Computer
  Graphics and Interactive Techniques. pp. 497--500 (2001)

\bibitem{ReinhardWardPattanaikDebevec}
Reinhard, E., Ward, G., Pattanaik, S., Debevec, P.: High Dynamic Range Imaging.
  Morgan Kaufmann (2005)

\bibitem{Seitz}
Seitz, S.M., Curless, B., Diebel, J., Scharstein, D., Szeliski, R.: {A
  Comparison and Evaluation of Multi-View Stereo Reconstruction Algorithms}.
  In: Proceedings of the IEEE Conference on Computer Vision and Pattern
  Recognition. pp. 519--528 (2006)

\bibitem{Shen2011}
Shen, L., Yeo, C.: Intrinsic images decomposition using a local and global
  sparse representation of reflectance. In: Proceedings of the IEEE Conference
  on Computer Vision and Pattern Recognition. pp. 697--704 (2011)

\bibitem{Song2017}
Song, J., Cho, H., Yoon, J., Yoon, S.M.: Structure adaptive total variation
  minimization-based image decomposition. IEEE Transactions on Circuits and
  Systems for Video Technology  (2017), (to appear)

\bibitem{Storath2014}
Storath, M., Weinmann, A.: Fast partitioning of vector-valued images. SIAM
  Journal on Imaging Sciences  7(3),  1826--1852 (2014)

\bibitem{UnityDelight}
{Unity-Technologies}: Delightingtool,
  \url{https://github.com/Unity-Technologies/DeLightingTool/}

\bibitem{Woodham1980a}
Woodham, R.J.: {Photometric Method for Determining Surface Orientation from
  Multiple Images}. Optical Engineering  19(1),  139--144 (1980)

\bibitem{Wu}
Wu, C., Wilburn, B., Matsushita, Y., Theobalt, C.: {High-quality shape from
  multiview stereo and shading under general illumination}. In: Proceedings of
  the IEEE Conference on Computer Vision and Pattern Recognition. pp. 969--976
  (2011)

\end{thebibliography}

\end{document}